%% file: ObsPartialOrderFinal.tex
\documentclass[twocolumn,nofootinbib]{article} 
\usepackage[utf8]{inputenc}
\usepackage{amsmath}
\usepackage{amsthm}
\usepackage[dvipsnames]{xcolor}
\usepackage{amssymb}
\usepackage{mathtools}
\usepackage{xcolor}
\usepackage[normalem]{ulem}
\usepackage{enumitem}
\newcommand{\stkout}[1]{\ifmmode\text{\sout{\ensuremath{#1}}}\else\sout{#1}\fi}

\usepackage[T1]{fontenc}
\usepackage[most]{tcolorbox}

\usepackage{caption}
\usepackage{subcaption}
\usepackage{amsfonts}
\usepackage{graphicx} 
\usepackage[hidelinks]{hyperref}
\usepackage{placeins}

\usepackage{pifont}

\usepackage{thmtools}
\usepackage{xspace}
\usepackage[font={small}]{caption}
\usepackage{appendix}
\newtheorem{definition}{Definition}
\newtheorem{lemma}{Lemma}
\newtheorem{theorem}{Theorem}
\newtheorem{proposition}{Proposition}

\newtheorem{conjecture}{Conjecture}
\usepackage{chngcntr}
\counterwithin{figure}{section}
\usepackage{tikz}
\usetikzlibrary{shapes,positioning,arrows}
\tikzset{>=stealth', c/.style={draw, rectangle, minimum height=2em, inner sep=0.5em}, q/.style={draw, circle, minimum height=2.4em, inner sep=0}, e/.style={<-}, ed/.style={dashed,<-}}
\newcommand{\dbot}{\mathbin{\text{$\bot\mkern-8mu\bot$}}}
\newcommand{\dsep}{\mathbin{\text{$\bot\mkern-8mu\bot$}_{\text{d}}}}
\newcommand{\esep}{\mathbin{\text{$\bot\mkern-8mu\bot$}_{\text{e}}}}
\newcommand{\vis}{\mathtt{Vnodes}}
\newcommand{\algebraic}{\text{algebraic}\xspace}
\newcommand{\nonalgebraic}{\text{nonalgebraic}\xspace}
\newcommand{\algebraicness}{\text{algebraicness}\xspace}
\newcommand{\nonalgebraicness}{\text{nonalgebraicness}\xspace}
\newcommand{\lat}{\mathtt{Lnodes}}
\newcommand{\edges}{\mathtt{edges}}
\newcommand{\nodes}{\mathtt{nodes}}
\newcommand{\pa}{\mathtt{pa}}
\newcommand{\Mtower}{\mathcal{M}{-}{\tt{profile}}}
\newcommand{\Stower}{\mathcal{S}{-}{\tt{profile}}}
\newcommand{\ch}{\mathtt{ch}}
\newcommand{\des}{\mathtt{des}}
\newcommand{\an}{\mathtt{an}}

\newcommand{\mdag}{\mathtt{LnodesToFaces}}
\newcommand{\can}{\mathtt{can}}

\newcommand{\param}{\mathtt{par}}
\usepackage{authblk}

\title{The Observational Partial Order of Causal Structures with Latent Variables}
\author{Marina Maciel Ansanelli}
\author{Elie Wolfe}
\author{Robert W. Spekkens}
\affil{Perimeter Institute for Theoretical Physics, 31 Caroline Street North, Waterloo, Ontario Canada N2L 2Y5 \\ Department of Physics and Astronomy, University of Waterloo, Waterloo, Ontario, Canada, N2L 3G1}
\date{}
\begin{document}

\maketitle

\begin{abstract}
	For two causal structures with the same set of visible variables, one is said to observationally dominate the other if the set of distributions over the visible variables realizable by the first contains the set of distributions over the visible variables realizable by the second.  
	Knowing such dominance relations is useful for adjudicating between these structures given observational data. 
	Here, we consider the problem of determining the partial order of equivalence classes of causal structures with latent variables relative to observational dominance. 
	We provide a complete characterization of the dominance order in the case of three visible variables, and a partial characterization in the case of 
	four visible variables.
	Our techniques also help to identify which observational equivalence classes have a set of realizable distributions that is characterized by nontrivial inequality constraints, analogous to Bell inequalities and instrumental inequalities.   
	We find evidence that as one increases the number of visible variables, the equivalence classes satisfying nontrivial inequality constraints become ubiquitous. (Because such classes are the ones for which there can be a difference in the distributions that are quantumly and classically realizable, this implies that the potential for quantum-classical gaps is also ubiquitous.) Furthermore, we find evidence that constraint-based causal discovery algorithms that rely solely  on conditional independence constraints have a significantly weaker distinguishing power among observational equivalence classes than algorithms that go beyond these (i.e., algorithms that also leverage nested Markov constraints and inequality constraints).
\end{abstract}

\vspace{5mm}



\tableofcontents

\section{Introduction}
\label{introductionsection}

One way to probe the causal influence of one variable on another is to intervene upon the first and see whether the distribution over the other changes.  Interventions are the basis of the randomized control trial. While such schemes yield strong causal conclusions, they can be prohibitively expensive (e.g., a drug trial), unethical (e.g., deciding whether smoking causes lung cancer), or simply impossible (e.g., in observational astronomy).  For this reason, it is useful to understand what causal conclusions can be drawn from purely observational data, that is, from the probability distribution over the visible variables.

The problem of determining which causal structures are viable explanations of a given distribution over visible variables is referred to as the problem of {\em causal discovery}. The set of probability distributions over the visible variables that can be causally explained by a given causal structure generally satisfy nontrivial constraints.  We refer to these constraints as {\em causal-compatibility constraints}.

An important class of causal inference problems concern causal structures wherein some of the variables may be latent (i.e., unobserved). 
In the special case where a causal structure does \emph{not} include latent variables, it will be called latent-confounder-free. While the causal-compatibility constraints of a latent-confounder-free causal structure always take the form of equalities, a causal structure that has latent variables can also exhibit causal-compatibility constraints in the form of inequalities. We will refer to these two types of constraints as \emph{equality constraints} and \emph{inequality constraints} respectively.  A complete solution of the causal discovery problem requires a characterization of all causal-compatibility constraints for each causal structure, which is a very difficult problem in general.   This is because the inequality constraints are hard to find.

A more straightforward project is to simply {\em classify} the causal structures. This task requires determining when a pair of such causal structures can realize the {\em same} set of distributions over the visible variables, in which case they are termed {\em observationally equivalent}. 
We can then classify these causal structures into  \emph{observational equivalence classes}.  More generally, one wishes to know, for each pair of causal structures, whether the set of distributions realizable by one contains the set of distributions realizable by the other, in which case the first is said to {\em observationally dominate} the second.  This dominance relation defines a partial order among the observational equivalence classes of  causal structures, which we term the \emph{observational partial order}. The determination of the observational partial order is the main aim of this article.
 
Of course,  a determination of {\em all} of the causal-compatibility constraints for each causal structure would be {\em sufficient} for determining the observational partial order, but it is not {\em necessary}, and this is why the project of identifying the observational partial order is easier than the project of achieving a full characterization of the causal-compatibility constraints for each causal structure. The project of determining the observational partial order, however, has applications in causal discovery: given knowledge of when two causal structures are observationally equivalent,  if we find a causal-compatibility constraint for one of them, we know it applies also to the other. In particular, a classification into observational equivalence classes is helpful for attesting the absence or presence of inequality constraints: if a causal structure is known to be observationally equivalent to another one that is latent-confounder-free, then we know that it cannot present inequality constraints.

Characterizing the observational partial order is also significant for identifying the best causal explanations of observational data.
  It is standard to argue that if two causal structures can both realize a probability distribution, one should prefer the structure that  has less explanatory power, that is, the one that   is lower in the observational dominance partial order.  (This is sometimes motivated by the fact that this structure is more falsifiable.). Identifying the set of causal structures that provide the best explanation of a given distribution, therefore, requires identifying the lowest elements in the observational dominance order  that can realize the distribution.   
      In this way, a knowledge of the observational partial order permits a narrowing-down of the scope of possibilities for a good causal explanation of the data.    As we will see in Section~\ref{significanceforcausaldiscovery}, the observational partial order is particularly significant when one seeks to take into account finite-sample effects in causal discovery.

This work focuses on rules for proving that two causal structures are observationally equivalent, inequivalent, or that one dominates the other.  We gather the known such rules from the literature, determine logical relations among these, and present some new ones 
 (Propositions~\ref{prop_moderateFS} and~\ref{Simultaneous_Splitting_Prop}).  With these rules, we aim to determine the resulting observational partial order. Sec.~\ref{sec_proven_partitions} will explain the precise procedure that we follow to do so.

\begin{figure}[htbp]
	\centering
	\includegraphics[width=0.4\textwidth]{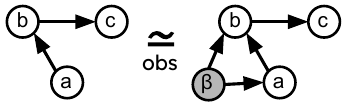}
	\caption{Two observationally equivalent causal structures.}
	\label{fig_example_constraints}
\end{figure}

We will use the convention adopted in the standard framework of causal inference, where a causal structure is represented by a Directed Acyclic Graph (DAG) whose nodes are associated to random variables and whose edges indicate direct causal influences. Here, we will refer to a DAG whose nodes are partitioned into two sets, one associated to latent variables and the other associated to visible variables, as a \emph{partitioned DAG} (pDAG). 
An example of observational equivalence is given by the pair of pDAGs depicted in Fig.~\ref{fig_example_constraints}, where the nodes with white background ($a$, $b$ and $c$) are associated with visible variables, and the node with grey background ($\beta$) is associated with a latent variable.  For both of these causal structures, the only constraint on the realizable distributions is that the variable associated with node $a$, which we will call $X_a$, is independent of the variable associated with the node $c$, which we will call $X_c$, upon conditioning on the variable associated with the node $b$, which we will call $X_b$. In other words, in both cases, $X_a$ and $X_c$ are conditionally independent given $X_b$. 

For reasons that are presented in Section~\ref{sec_temporal_order}, we will pursue this classification project   for a set of causal structures that are consistent with a fixed ordering of the visible nodes, i.e., a fixed  \emph{nodal ordering}. For example, in our classification of 2-node mDAGs, the set of mDAGs we consider will not include both $a\rightarrow b$ and  $a\leftarrow b$.  This is because the former is only consistent with the nodal ordering $(a,b)$, while the latter is only consistent with the nodal ordering $(b,a)$.

As mentioned, meeting the challenge of identifying the
 observational equivalence classes also contributes towards identifying pDAGs that present \emph{inequality} constraints among their causal-compatibility constraints. Such pDAGs will be called \emph{\nonalgebraic}, for reasons that we explain in Section~\ref{sec_causal_compatibility}. The study of inequality constraints is of particular importance for quantum physicists who are interested in causal inference,  because such constraints are analogues of Bell inequalities~\cite{Bell_1964}, and it is only through their violation that one can witness the need for quantumness of a causal model~\cite{Fritz_2012, VanHimbeeck2019quantumviolationsin, Lauand_evans, Alanon_Shadow}. Motivated by this, there has been work on the problem of identifying when a causal structure will present inequality constraints~\cite{henson_theory-independent_2014, Pienaar_2017, Khanna_2023}. 

In the realm of classical causal inference, the most common constraint-based causal discovery algorithms (such as IC and PC)~\cite{Verma_Pearl_equivalence1990, Spirtes2000} leverage only conditional independence relations (which are \emph{equality} constraints) in trying to adjudicate between the different observational equivalence classes. 
 These algorithms can be supplemented by techniques for identifying the full set of equality constraints, including nested Markov constraints~\cite{behaviormetrika}. However, any algorithm that is restricted to using equality constraints necessarily lacks 
 discriminatory power relative to algorithms that appeal to inequality constraints.

An important question is: how significant are causal discovery techniques that go beyond conditional independence constraints?  For every causal structure whose causal-compatibility constraints are not simply conditional independence relations, it is known~\cite{Evans2023} that there are nontrivial inequality constraints.  Consequently, the first question one might ask is how \emph{rare} inequality constraints are; if they are too rare, one might conclude that not much is lost by leveraging only conditional independence relations in one's causal discovery algorithm.

When we look at data obtained from passive observation of the visible nodes, the only opportunity for distinguishing two causal structures is if the set of probability distributions realizable by one is different from the set realizable by the other.  For example, there is no opportunity for observationally distinguishing two distinct causal structures that are both in the saturating class, since the realizable set of distributions is the full set of possible distributions over the visible variables in both cases.
In particular, given data obtained from passive observation of the visible nodes, one does not have the ability to make any discrimination that is more fine-grained than the observational equivalence classes.\footnote{The discriminations one can make may generally also be more {\em coarse-grained} than the observational equivalence classes.  Consider again data obtained from passive observation of the visible nodes.  Without making further assumptions (such as appealing to the principle of faithfulness), this data will not uniquely pick out an observational equivalence class.  For one, there might be two or more observational equivalence classes that are {\em incomparable} within the partial order but for which the corresponding sets of realizable distributions have a nontrivial intersection, and the distribution of interest might lie in this intersection.  For another, for every class that can realize the distribution, the set of classes in its upward closure within the observational partial order can {\em also} realize the distribution.}
It follows that in order to get a sense of the significance of inequality constraints in making discriminations among causal structures, one can count the fraction of observational equivalence classes that imply inequality constraints. 

 It is worth noting that we are {\em not} seeking to determine whether \nonalgebraic causal structures are  {\em physically} ubiquitous in a given context, in the sense of arising with greater frequency.  To do so, one would require a prior probability distribution over the set of possible causal structures, which will be context-dependent and generally difficult to estimate.  Rather, we are here focussed on the question: what is everything that can in principle be inferred about the causal structure from passive observation of the visible variables?  Such inferences proceed by ruling out every causal structure for which there  is a causal-compatibility constraint that is violated by the observed distribution.  All of the causal structures in an observational equivalence class are ruled out (or ruled in) together.  Consequently, when seeking to determine the significance of inequality constraints in such inferences, it is natural to determine how prevalent they are {\em among the observational equivalence classes}.

Here, we consider the question of  what proportion of { observational equivalence classes} are \nonalgebraic for sets of causal structures consistent with a fixed nodal ordering that have two, three and four visible nodes.

For causal structures consistent with a fixed nodal ordering that have two nodes, there are only two observational equivalence classes, and both are \algebraic. For causal structures consistent with a fixed nodal ordering that have three nodes,  we find that there are fifteen observational equivalence classes, of which five are \nonalgebraic. Thus, the fraction of \nonalgebraic observational equivalence classes is $1/3$.
For causal structures consistent with a fixed nodal ordering that have four nodes, we find that this fraction is at least $0.852$  (the uncertainty about this number comes from the fact that we do not yet have a full classification of pDAGs with four visible variables into observational equivalence classes).

The trend that is emerging is that the fraction of \nonalgebraic observational equivalence classes 
 increases with the number of visible variables\footnote{Recent work by Evans~\cite{Evans2023} shows that pDAGs that have nontrivial Nested Markov constraints always have nontrivial inequality constraints as well. That finding, however, is insufficient to establish the claim that the proportion of observational equivalence classes that have inequalities increases with the number of visible nodes.}. This suggests that as one increases the number of visible variables in the causal structures being considered, 
  inequality constraints become generic among the observational equivalence classes.


Our result about the abundance of \nonalgebraic observational equivalence classes has particular relevance for the study of quantum causal models: 
  it indicates that the majority of observational equivalence classes of causal structures have the potential to show  a quantum-classical gap.

  Ref.~\cite{Khanna_2023} already identified which causal structures (i.e., which mDAGs) with 2, 3 or 4 visible nodes are \nonalgebraic.  However, only if one sorts  these causal structures into observational equivalence classes can one determine what fraction of the observational equivalence classes are \nonalgebraic.  
 In short,  our results show that causal discovery algorithms that leverage only conditional independence constraints have a poor discriminatory power relative to algorithms that go beyond these for the case of four visible nodes, and this is likely to also be true for higher numbers of visible nodes.

The paper is structured as follows: after an exposition of the relevant background concepts in Section~\ref{sec_preliminary_notions}, we will present our objectives and our results for the case of three visible variables in Section~\ref{sec_objectives}. In particular, Section~\ref{sec_3_observed_nodes} presents and discusses the complete observational partial order of causal structures of three visible nodes consistent with a fixed nodal ordering. We will then present the known rules to show observational dominance and nondominance in Sections \ref{sec_show_equivalence} and \ref{sec_show_inequivalence} (including two new dominance-proving rules in Sections \ref{sec_face_splitting} and \ref{sec_simultaneous_splitting}). In light of that, Section~\ref{sec_back_to_3visible} will justify the results for causal structures with three visible nodes that were presented without justification in  Section~\ref{sec_3_observed_nodes}. Section~\ref{sec_4} then presents some partial results about the case of causal structures of four visible nodes consistent with a fixed nodal ordering, and we end the paper with concluding remarks in Section~\ref{sec_conclusion}.

\section{Preliminaries}
\label{sec_preliminary_notions}

In this work, causal structures are represented by Directed Acyclic Graphs (DAGs), directed graphs that have no cycles.  DAGs are composed of nodes and directed edges, specified as ordered pairs of nodes. Here, we will denote the set of nodes of a DAG $\mathcal{G}$ as $\nodes(\mathcal{G})$. The set of edges of a DAG $\mathcal{G}$ will be denoted $\edges(\mathcal{G})$.

Below, we define some terminology about DAGs.

\begin{definition}[Children, Parents, Descendants, Ancestors]
	
	Let $\mathcal{G}$ be a DAG, and let $u,v,w\in\nodes(\mathcal{G})$.
	
	If $\mathcal{G}$ includes a directed edge $v\rightarrow w$, then $w$ is called a \underline{child} of $v$. Conversely, a node $u$ that has $v$ as a child is called a \underline{parent} of $v$. The set of all children of $v$ is denoted $\ch_{\mathcal{G}}(v)$, while the set of all parents of $v$ is denoted $\pa_{\mathcal{G}}(v)$.
	
	A node $w\in \nodes(\mathcal{G})$ is a \underline{descendant} of $v$ if it can be reached from $v$ by a sequence of nodes through directed edges pointing away from $v$. Conversely, a node $u\in\nodes(\mathcal{G})$ that has $v$ as a descendant is called an \underline{ancestor} of $v$. The set of all descendants of $v$ is denoted $\des_{\mathcal{G}}(v)$, and the set of all ancestors of $v$ is denoted $\an_{\mathcal{G}}(v)$.
\end{definition}

\begin{definition}[Subgraph]
	\label{def_subgraph}
	Let $\mathcal{G}$ and $\mathcal{G}'$ be DAGs. If $\nodes(\mathcal{G}')\subseteq \nodes(\mathcal{G})$ and $\edges(\mathcal{G}')\subseteq\edges(\mathcal{G})$, we say that $\mathcal{G'}$ is a subgraph of $\mathcal{G}$. 
	
	In particular, if we start from $\cal G$ and delete all of the nodes that are \emph{not} in a particular subset $S\in\nodes(\cal G)$ (hence also deleting their incoming and outgoing edges), the resulting induced subgraph will be denoted by $\mathcal{G}_S$. 
\end{definition}

As already noted in the introduction, the causal structures that we consider in this work have two types of nodes: visible nodes, associated with variables that are observed, and latent nodes, associated with variables that are not observed. In order to save the term ``DAG'' for the mathematical object composed of nodes and edges, we define a new object to represent a DAG with a partitioned set of nodes:

\begin{definition}[Partitioned DAG]
	A \emph{partitioned DAG (pDAG)} $\mathcal{G}$ is a DAG together with a partition of its nodes into two subsets. The first subset is referred to as the set of
	\emph{visible} nodes, denoted $\vis(\cal G)\subseteq \nodes(\cal G)$, and the second subset is referred to as the set of
	\emph{latent} nodes, denoted $\lat(\cal G)\subseteq \nodes(\cal G)$.
\end{definition}

All of the notions defined above for DAGs also apply for pDAGs. In this paper, we denote visible nodes with white backgrounds and latent nodes with grey backgrounds (see, e.g.,  Fig.~\ref{fig_example_constraints}).  A pDAG that is composed only of visible nodes will be called a \emph{latent-confounder-free} pDAG. Whenever two visible nodes of a pDAG are children of the same latent node, we will refer to the latent node as a \emph{ latent  confounder} of the two children.

We now define one last preliminary notion, the notion of a set of pDAGs consistent with a fixed ordering of the visible nodes:

\begin{definition}[Set of pDAGs consistent with a fixed nodal Ordering]
	\label{def_nodal_ordering}
	
	Consider a set $S$ of pDAGs, all of which share the same set of visible nodes $\{v_1,...,v_n\}$. We say that $S$ is a \emph{set of pDAGs consistent with a fixed nodal ordering} if there is an ordering of the nodes, $(v_{\Pi(1)},...,v_{\Pi(n)})$ for some permutation $\Pi$, such that every pDAG of $S$ is such that  $v_{\Pi(j)}$ is not an ancestor of $v_{\Pi(i)}$ for all $i,j$ such that $\Pi(i)<\Pi(j)$.
\end{definition}

\subsection{Explaining Observed Data with a Causal Structure}
\label{sec_causal_compatibility}

A natural experiment is one that passively observes a set of variables over many runs. Such an experiment yields a statistical estimate of the probability distribution over the possible values of these variables. Determining which causal structures could have realized this distribution is the causal discovery problem.

To do so, one has to specify the \emph{semantics} of this causal model\footnote{A causal structure is analogous to syntax in linguistics, which is the structural form of a sentence, while the semantics of a causal model is the analogue of semantics in linguistics.  Semantics refers to a specific choice of objects that follow the structure defined by the syntax. In linguistics, semantics provides the meanings of the terms in a sentence.  In causal modelling, it provides the parameters of the model. 
	We thank Nick Ormrod for the suggestion of this terminology.}, a description of what type of entities are associated to the nodes, and what type of mechanism determines how each node depends causally on its parents in the DAG, and of how to parametrize statistical uncertainty.

In this work, we will only consider causal model semantics of the  \emph{classical} type, wherein the  entities that are associated with nodes are classical random variables. An alternative possibility, however, is the semantics used in quantum causal modelling. There, the entities that can be associated with nodes are quantum systems, each of which is described by a Hilbert space\footnote{The analogue of the cardinality of a random variable is the dimension of this Hilbert space.  The analogue of an error variable of a given cardinality is an ancillary quantum system of a given dimension, and the analogue of a distribution over the error variable is a quantum state of the ancillary system. Finally, the analogue of a functional dependence between classical random variables is a unitary channel between quantum systems.}. See Refs.~\cite{chaves_informationtheoretic_2015, elie_quantum_inflation, Smith_fullyquantuminf} for works on the causal discovery problem in causal models with quantum semantics.

The variables associated with visible nodes will be called \emph{visible variables}, and the variables associated with latent nodes will be called \emph{latent variables}. For each $a\in \nodes(\mathcal{G})$, we denote the associated variable by $X_a$, and for a set $S$ of nodes, we denote the associated joint variable by $X_S$.  The set of all possible values of $X_a$, termed its \emph{state space}, is denoted $\mathcal{X}_a$. In this work, we will assume that each variable has a discrete and finite number of possible outcomes.\footnote{As proven for example in Ref.~\cite{Fraser_Combinatorial_Solution}, when the visible variables have finite cardinality, we can assume that the latent variables also have finite cardinality without loss of generality. }
Under this assumption, 
the cardinality of each $\mathcal{X}_a$ is a natural number, so that the set of such cardinalities can be specified by a vector $\vec c\in \mathbb{N}^{|\nodes(\mathcal{G})|}$, where $|S|$ denotes the cardinality of a set $S$. The semantics of a causal model includes a specification of $\vec c$, the cardinality of the variables.  

 We also associate to each $a\in \nodes(\mathcal{G})$ a random variable $E_a$, called an error variable, whose state space is denoted by $\mathcal{E}_a$.  The variable $X_a$ is determined by its associated error variable $E_a$ and its causal parents $X_{\pa_{\mathcal{G}}(a)}$ via a function $f_a:\mathcal{X}_{\pa_{\mathcal{G}}(a)}\times \mathcal{E}_a\rightarrow \mathcal{X}_a$, that is, 
\begin{equation}
	X_a=f_a(X_{\pa_{\mathcal{G}}(a)},E_a) 
\end{equation}
Finally, to each error variable $E_a$, one associates a probability distribution $P(E_a)$ that stipulates how $E_a$  is sampled. Thus, the inclusion of the error variable $E_a$ allows for $X_a$ to respond probabilistically to $X_{\pa_{\mathcal{G}}(a)}$, even though $f_a$ is a deterministic function.

The cardinality of 
the visible, latent and error variables, the functions, and the error-variable distributions are jointly termed the {\em parameters} of the causal model:
\begin{equation}
	\param =  \{ (|\mathcal{X}_a|,|\mathcal{E}_a|,f_a , P(E_a)) : a\in\nodes (\mathcal{G})\} 
\end{equation}

The semantics of the causal model stipulates the scope of possibilities for $\param$.  For instance, the functional dependences could all be linear and the error-variable distributions could all be uniform. 
It could also be the case that the scope of possibilities for $\param$ is unrestricted. When this is the case,
we will say that the semantics is {\em classical unrestricted}.   Note that there are still different options of classical unrestricted semantics, corresponding to different choices of cardinalities of the visible, latent and error variables.

In all of this work, we will be dealing only with classical unrestricted semantics. In general, one might be interested in the distributions over visible variables realizable by a pDAG for different types of classical unrestricted semantics; for example, when the latent nodes are associated with classical random variables that have specific fixed cardinalities. In the causal inference literature, the case of most common interest is when the latent nodes are allowed to take arbitrary cardinalities. We now turn to the notion of realizability of observational data under this type of model.

\begin{definition}[Set of probability distributions with cardinality vector $\vec c_\text{vis}$ realizable by a pDAG by a causal model with classical unrestricted semantics under arbitrary cardinality of the latent variables]
	\label{SEP_definition}

	Let $\mathcal{G}$ be a pDAG. Associate it to a causal model with classical unrestricted semantics where the visible variables $\{X_v:v\in \vis(\mathcal{G})\}$ have cardinalities given by the vector  $\vec c_\text{vis}\in \mathbb{N}^{|\vis(\mathcal{G})|}$, and the latent variables are allowed to take arbitrary cardinalities. 
	
	The set of  probability distributions on the visible variables that are realizable under passive observation by $\cal G$ for this causal semantics, denoted $\mathcal{M(G},\vec c_\text{vis})$, is the set of all probability distributions $P$ over $X_{\vis(\cal G)}$ that can be obtained as
	\begin{align}
		&P(X_{\vis(\mathcal{G})}) \nonumber\\
		&=\sum_{X_{\lat(\mathcal{G})}}\sum_{E_{\nodes(\mathcal{G})}}  
		\prod_{a\in \nodes(\mathcal{G})}  \delta_{X_a, f_a(X_{\pa_{\mathcal{G}}(a)},E_a) } P(E_a)
		\label{eq_realizability}
	\end{align}
	for some choice of parameters, i.e., for some choice of the set $\{(|\mathcal{X}_a|,|\mathcal{E}_a|, f_a, P(E_a)) : a\in \nodes(\mathcal{G})\}$.
\end{definition}

We will refer to a distribution that is realizable by a pDAG  under arbitrary cardinality of the latent variables simply as a distribution that is realizable by that pDAG. 

If a given probability distribution over visible variables is not realizable in 
a certain causal structure, then this causal structure is \emph{not} a valid causal explanation for the statistical data on these variables. Causal structures that can realize \emph{all} the probability distributions over the visible variables are called \emph{saturated}.

The set $\mathcal{M}(\mathcal{G},\vec c_{\text{vis}})$ for a pDAG $\cal G$ is defined with respect to a specific assignment of the cardinalities of the visible variables. It is useful to consider the tuple of possibilities for this assignment, and the corresponding tuple of probability distributions that are realizable for each possibility.  
Since this tuple contains all the information that can be extracted from the causal structure by passive observations, we call it the \emph{observational profile} of that structure:

\begin{definition}[Observational profile]
	Let $\mathcal{G}$ be a pDAG. The \emph{observational profile} of $\cal G$, denoted $\Mtower(\mathcal{G})$, is the tuple
	\begin{equation}
		\Mtower{(\mathcal{G})} \coloneqq \left( \mathcal{M}(\mathcal{G},\vec{c}_\text{vis} ): \vec c_\text{vis} \in\mathbb{N}^{|\vis(\mathcal{G})|}\right), 
	\end{equation}
	that describes the set of realizable distributions over the visible variables of $\mathcal{G}$ for each choice $\vec c_\text{vis}$ of their cardinalities.

\end{definition}

We refer here to a \emph{tuple} rather than a set (and use parentheses rather than curly brackets) to
 highlight that the order in which the terms appear inside the observational profile is important. In other words, in the observational profile every set  $\mathcal{M}(\mathcal{G},\vec c_{\text{vis}})$ is indexed by the vector $\vec c_\text{vis}$ of cardinalities of the visible variables.  This allows us to define an inclusion relation among  observational profiles:
\begin{definition}[{Element-wise} inclusion of observational profiles]
	\label{def_elementwise}
	Let $\mathcal{G}$ and $\mathcal{G}'$ be pDAGs such that $\vis(\mathcal{G})=\vis(\mathcal{G}')$. We say that the observational profile of $\cal G'$ is \emph{ {element-wise} included} in the observational profile of $\cal G$, denoted by $\Mtower(\cal G')\subseteq_\text{ew}\Mtower(\cal G)$, if 
		\begin{equation}
		\mathcal{M}(\mathcal{G}',\vec c_\text{vis}) \subseteq   \mathcal{M}(\mathcal{G},\vec c_\text{vis}) \quad \forall \vec c_\text{vis} \in \mathbb{N}^{|\vis(\mathcal{G})|}.
	\end{equation}
\end{definition}

The mathematical constraints on the set of realizable probability distributions of a given causal structure can be sorted into trivial and nontrivial types. The trivial type merely follows from nonnegativity of probabilities and normalization.
The nontrivial types of constraint are those that are implied by the causal structure, and are termed \emph{causal-compatibility constraints}. Using the observational profile, as we will see now, it is possible to classify pDAGs based on the types of causal-compatibility constraints that they present.

The set $\mathcal{M}(\mathcal{G},\vec c_{\text{vis}})$ is subject to both trivial and nontrivial constraints. To focus on the latter, we define $\tilde{\mathcal{M}}(\mathcal{G},\vec c_\text{vis})$ as the set obeying \emph{only} the nontrivial constraints. Note  that, geometrically, $\mathcal{M}(\mathcal{G},\vec c_\text{vis})$ can be recovered from $\tilde{\mathcal{M}}(\mathcal{G},\vec c_\text{vis})$ simply by restricting the latter to the simplex defined by the nonnegativity and normalization constraints. 
Out of the trivial constraints, normalization is a linear equality and nonnegativity is a linear inequality.  In Ref.~\cite{Rosset_Bound}, it is proven that for classical causal models, all nontrivial causal-compatibility constraints also take the form of polynomial equalities or inequalities.  If all such constraints are equalities, then the set $\tilde{\mathcal{M}}(\mathcal{G},\vec c_\text{vis})$ is an {\em algebraic set} (also termed an algebraic variety),  \footnote{Note that the set $\mathcal{M}(\mathcal{G},\vec c)$, unlike $\tilde{\mathcal{M}}(\mathcal{G},\vec c)$, is  {\em never} an algebraic set  simply because the nonnegativity constraints are inequality constraints. } 
It is convenient to use  the mathematical notion of algebraicness of sets   
 to define a dichotomy among pDAGs,  for which we will re-use the ``algebraic'' terminology.

\begin{definition}[Algebraicness]
	A pDAG $\mathcal{G}$  will be said to be \algebraic if, for every element $\mathcal{M}(\mathcal{G},\vec c_\text{vis})$ of $\Mtower(\mathcal{G})$, all nontrivial causal-compatibility constraints are equality constraints, or, equivalently, if the set $\tilde{\mathcal{M}}(\mathcal{G},\vec c_\text{vis})$ (wherein one has dropped the trivial constraints of nonnegativity and normalization relative to  $\mathcal{M}(\mathcal{G},\vec c_\text{vis})$) is an algebraic set.  Otherwise, the pDAG $\mathcal{G}$ is said to be \nonalgebraic.
\end{definition}

As established by Ref.~\cite{dsep_complete}, the only causal-compatibility constraints that are implied by a {\em latent-confounder-free} pDAG $\mathcal{G}$ are conditional independence relations, which are equality constraints that are valid regardless of the choice of cardinalities. Hence, all latent-confounder-free pDAGs are \algebraic. 
One example of a \nonalgebraic pDAG is the causal structure appearing
in Bell's theorem~\cite{Bell_1964}, because this structure implies causal-compatibility constraints that take the form of inequalities, namely, Bell inequalities~\cite{Wood_and_Spekkens}. As we will see later on, the classification of pDAGs under observational equivalence and dominance is highly pertinent to the question of which pDAGs are \algebraic and which are \nonalgebraic. 

The terminology of \algebraic/\nonalgebraic causal structures was introduced in Ref.~\cite{Khanna_2023}. Note, however, that we have here been more explicit than in Ref.~\cite{Khanna_2023} about the fact that this distinction relies on the observational profile of a pDAG, that is, the tuple obtained by looking at all possible cardinalities of the visible variables. 
This emphasis on cardinality is important because the presence of certain inequality constraints might depend on the choice of cardinalities of the visible variables. For example, the Bonet inequalities~\cite{Bonet_Instrumental} for the instrumental scenario (first pDAG of Fig.~\ref{fig_instrumental_pDAG}) are constraints on distributions where the instrumental variable (associated to node $a$ in Fig.~\ref{fig_instrumental_pDAG}) has cardinality at least $3$. On the other hand, Pearl's instrumental inequalities~\cite{Pearl_Instrumental} for the same pDAG have to be satisfied even when all variables have binary cardinalities. 

Our definition of \algebraicness is such that a pDAG only needs to present inequality constraints for \emph{one} choice of cardinalities of the visible variables to be considered \nonalgebraic.  This captures the {potential} that a causal structure has of presenting inequality constraints in \emph{some} scenario, instead of being a definition that is only applicable to a specific causal discovery  scenario (in which case the cardinality of the visible variables $\vec c_\text{vis}$ would be known, and we would not need to discuss the entire observational profile). 
As of now, however, we do not  know of any  
 example of a pDAG that is   \nonalgebraic while also being free of inequality constraints  when all visible variables are binary, i.e., have cardinality $2$.  To highlight this open problem, we conjecture that such examples do not exist. 

\begin{conjecture}
	\label{conj_nonalgebraic}
	 If a pDAG is \nonalgebraic, this implies
	 inequality constraints even when the   cardinalities of all the visible variables have their smallest possible nontrivial values, i.e., $\vec c_{\text{vis}} = (2,2,\dots)$.
\end{conjecture}

In Section~\ref{sec_every_support}, we will present possible candidates for  counterexamples to Conjecture~\ref{conj_nonalgebraic}: the causal structures of Fig.~\ref{fig_compatible_any_support}(a)-(d) are known to present inequality constraints at cardinalities $\vec c_\text{vis}=(3,2,2,2)$, but we still do not know if they present inequality constraints at cardinalities $\vec c_\text{vis}=(2,2,2,2)$.

Unlike inequality constraints, the equality constraints implied by a causal structure do not depend on  the cardinality of the visible variables. Furthermore, it is much easier to obtain equality constraints than inequality constraints.  The simplest type of equality constraints, namely the conditional independence relations, can be obtained from a graphical criterion called the \emph{d-separation criterion}~\cite{verma_pearl, Geiger1988}.  Given a pDAG $\mathcal{G}$ and three disjoint sets of visible nodes $A,B,C\subseteq \nodes(\mathcal{G})$, the criterion returns whether $A$ and $B$ are ``d-separated'' by $C$ or not. The affirmative case is denoted by $A\dsep B|C$. As shown in Ref.~\cite{verma_pearl} and reproduced in Theorem \ref{th:d-sep} below, a d-separation relation of a pDAG implies that the realizable distributions have to satisfy an associated conditonal independence relation. In a certain distribution $P$, the variable $X_A$ is said to be independent of the variable $X_B$ after conditioning on the variable $X_C$ if: 
\begin{equation}
	P(X_A X_B|X_C)=P(X_A|X_C)P(X_B|X_C).
\end{equation}
Such a conditional independence relation is denoted by $X_A \dbot X_B |X_C$. Note that it has to be satisfied regardless of the cardinality of $X_A$, $X_B$ and $X_C$. 

\begin{theorem}[d-separation and conditional independence~\cite{verma_pearl}]
	\label{th:d-sep}
	Let $\mathcal{G}$ be a pDAG, and associate it to a classical unrestricted semantics under arbitrary cardinalities of the latent variables. Let $A$, $B$ and $C$ be three disjoint subsets of the set of visible nodes, $\vis(\cal G)$. Then:
	
	\begin{enumerate}
		\item If $P$ is a probability distribution over $X_{\vis(\mathcal{G})}$ realizable by $\mathcal{G}$, then $A\dsep B|C$ in $\mathcal{G}$ implies that $X_A \dbot X_B |X_C$ in $P$.
		\item If $X_A \dbot X_B |X_C$ for \underline{every} probability distribution over $X_{\vis(\mathcal{G})}$ that is realizable by $\mathcal{G}$, then $A\dsep B|C$.
	\end{enumerate}
	
\end{theorem}

As an example, both pDAGs of Fig.~\ref{fig_example_constraints} present the d-separation relation $a\dsep c|b$, and thus Theorem \ref{th:d-sep} implies that any distribution realizable by these pDAGs must satisfy the conditional independence constraint $X_a \dbot X_b |X_c$.

\subsection{Observational Equivalence and Dominance}

The central objective of this work is to determine the observational equivalence and dominance relations that hold between pDAGs, and so we turn now to defining these formally.

\begin{definition}[Observational dominance and equivalence of pDAGs]
	\label{def_obs_equivalence}
	Let $\mathcal{G}$ and $\mathcal{G}'$ be pDAGs such that $\vis(\mathcal{G})=\vis(\mathcal{G}')$.
	
	We say that $\mathcal{G}$ \emph{observationally dominates} $\mathcal{G}'$ when the set of realizable distributions of  $\mathcal{G}$ includes the set of realizable distributions of  $\mathcal{G}'$, regardless of the  cardinalities of the visible variables. That is, when  $\Mtower(\cal G')\subseteq_\text{ew}\Mtower(\cal G)$. 
	The observational dominance relation is denoted by $\mathcal{G}\succeq\mathcal{G}'$.
	
	We say that $\mathcal{G}$ is \emph{observationally equivalent} to $\mathcal{G}'$ when  $\Mtower(\cal G')=\Mtower(\cal G)$.

 The observational equivalence relation is denoted by $\mathcal{G}\cong\mathcal{G}'$. 
	
	If $\mathcal{G}\succeq\mathcal{G}'$ but $\mathcal{G}\not\cong\mathcal{G}'$,
	we say that $\mathcal{G}$ \emph{strictly dominates} $\mathcal{G}'$ and denote this relation as $\mathcal{G}\succ\mathcal{G}'$.
	If $\mathcal{G}\not\succeq\mathcal{G}'$ and $\mathcal{G}'\not\succeq\mathcal{G}$, we say that $\mathcal{G}$ and $\mathcal{G}'$ are \emph{incomparable}.
\end{definition}

\begin{figure}[h!]
	\centering
	\includegraphics[width=0.5\textwidth]{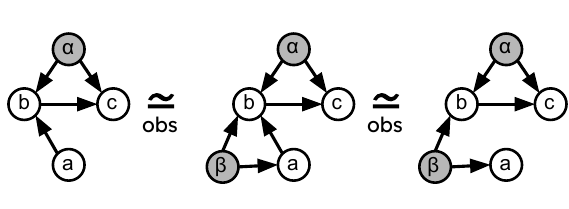}
	\caption{An example of \nonalgebraic pDAGs that are observationally equivalent. The first of this triple is generally known as the \emph{instrumental scenario}.} 
	\label{fig_instrumental_pDAG} 
\end{figure}

Note that to prove observational equivalence,  one must demonstrate  that $\Mtower(\cal G')=\Mtower(\cal G)$, which requires demonstrating equality for all possible cardinalities of the visible nodes.
To prove observational {\em in}equivalence, on the other hand, it suffices to show that $\exists \vec c_\text{vis}\in \mathbb{N}^{\vis(\mathcal{G})} :\mathcal{M}(\mathcal{G},\vec c_\text{vis})\ne \mathcal{M}(\mathcal{G}',\vec c_\text{vis})$, that is, it suffices to demonstrate inequality for just one particular assignment of cardinalities. 
An example of two pDAGs
that are observationally equivalent to each other was given in Fig.~\ref{fig_example_constraints}.  Another example, involving \nonalgebraic pDAGs, is given in Fig.~\ref{fig_instrumental_pDAG}; the observational equivalence of the three pDAGs there presented will be formally justified in Section~\ref{sec_back_to_3visible}.

Recall that if two pDAGs are observationally equivalent, they impose the same set of causal-compatibility constraints on the  
probability distributions over the visible variables. Therefore, it is clear that all of the pDAGs that are observationally equivalent to an \algebraic pDAG are also \algebraic, and all of the pDAGs that are observationally equivalent to a \nonalgebraic pDAG are also \nonalgebraic.

Note  that observational equivalence is \emph{not} the same as Markov equivalence~\cite{Verma_Pearl_equivalence1990,Ayesha_Markov}. Two pDAGs are Markov equivalent if their sets of conditional independence constraints are the same. For two pDAGs to be observationally equivalent, on the other hand, their entire set of causal-compatibility constraints (including conditional independence constraints, nested Markov constraints and inequality constraints) need to coincide.

\subsection{mDAGs}

In this section, we recall the notion of an mDAG, defined in Ref.~\cite{evans_graphs_2016}. For each mDAG, we can associate a set of pDAGs. The pDAGs in such a set are not only observationally equivalent, but equivalent under any interventional probing scheme~\cite{ansanelli2024everything}.
That is, if two pDAGs are associated with the same mDAG, then they cannot be distinguished even when there is access to data from observations and arbitrary interventions. Therefore, the mDAG is a fundamental object in the causal discovery problem with latent variables. Note, however, that observational equivalence does not imply equivalence of the mDAG structure: there are different mDAGs which are observationally equivalent. 

The definition of an mDAG is motivated by two graphical operations defined in Ref.~\cite{evans_graphs_2016}, here referred to as \emph{exogenization of latent nodes} and \emph{removal of redundant latent nodes}. Ref.~\cite{evans_graphs_2016} showed that these two graphical operations preserve the observational profile. That is, any two pDAGs related by such an operation are observationally equivalent. These operations are discussed separately from the rest of the rules for showing observational equivalence (which will be discussed in Section~\ref{sec_show_equivalence}) because they are more fundamental: as Ref.~\cite{ansanelli2024everything} showed, two pDAGs related by these operations are   not only indistiguishable when we are restricted to passive observations, but also when we have access to interventions. In fact, Ref.~\cite{ansanelli2024everything} showed that such pDAGs are indistinguishable even under access to the \emph{strongest possible} interventional probing schemes on the visible variables, which are therein called \emph{informationally complete}.   In this paper, we are only concerned with observational data, and will not further discuss interventions.

\begin{figure*}[h!]
	\centering
	\includegraphics[width=0.7\textwidth]{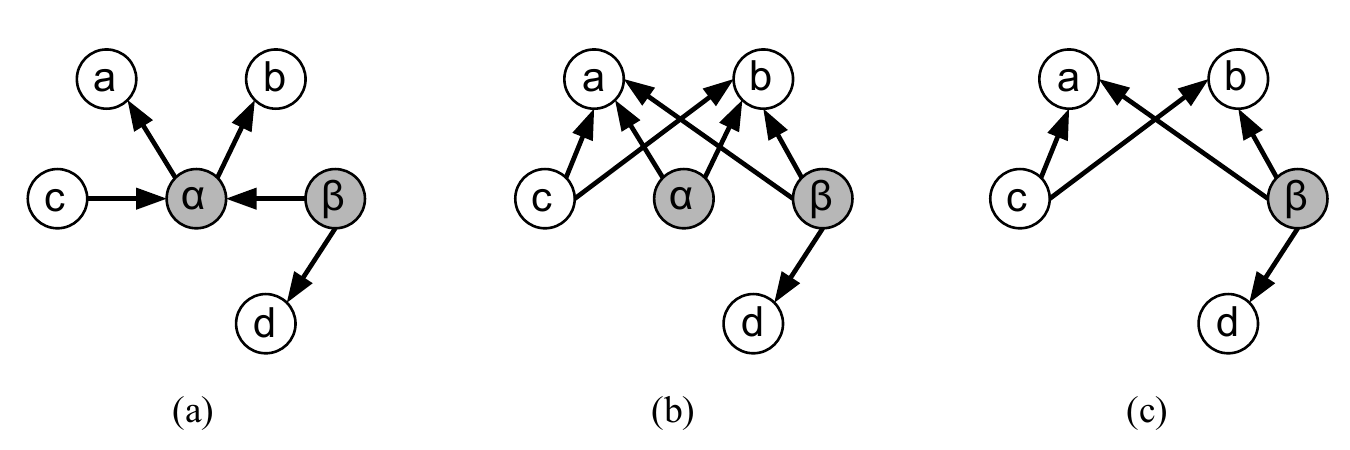}
	\caption{(a) A pDAG. (b) The pDAG obtained from (a) by exogenizing the latent node $\alpha$.
		(c) The pDAG obtained from that of (b) by removing the latent node $\alpha$, which is redundant to $\beta$ since the children of $\alpha$ are a subset of the children of $\beta$.
		By Lemmas \ref{lemma_exogenize_latents} and \ref{lemma_remove_redundant_latents}, these three pDAGs are observationally equivalent. The pDAG in (c) is {\tt RE}-reduced.}
\label{fig_example_lemmas} 
\end{figure*}

Before defining these two operations in full generality, we can get a sense of their content by analyzing Fig.~\ref{fig_example_lemmas}. In Fig.~\ref{fig_example_lemmas}(a), we have a pDAG where the latent node $\alpha$ is \emph{endogenous}, since it has parents $\beta$ and $c$ in the pDAG.  It passes information from $\beta$ and $c$ to its children $a$ and $b$. In Fig.~\ref{fig_example_lemmas}(b), the latent node $\alpha$ does not have parents anymore, but its children $a$ and $b$ now have direct access to the information about $\beta$ and $c$. The operation of transitioning from the pDAG of Fig.~\ref{fig_example_lemmas}(a) to that of Fig.~\ref{fig_example_lemmas}(b) is an instance of what we term ``exogenization of latent nodes''. It consists in making the latent variables of a pDAG \emph{exogenous} (parentless) while connecting their parents directly to their children.

In Fig.~\ref{fig_example_lemmas}(b), the latent node $\alpha$ establishes a common cause between $a$ and $b$. On the other hand, the latent node $\beta$ establishes a common cause between $a$, $b$ and $d$. Since we allow the cardinalities of latent variables to be arbitrary (Def.~\ref{SEP_definition}), the latent node $\alpha$ can be absorbed into $\beta$, thus leading to Fig.~\ref{fig_example_lemmas}(c). The operation of transitioning from the pDAG of Fig.~\ref{fig_example_lemmas}(b) to that of Fig.~\ref{fig_example_lemmas}(c) is an instance of what we term ``removal of redundant latent nodes''. It consists in deleting latent nodes such as $\alpha$, whose set of children is a subset of the set of children of another latent node. The fact that these nodes only establish a  latent  confounder between nodes that already had a  latent  confounder is what justifies referring to them as redundant.

The proof that these two operations always preserve  observational equivalence was given in Lemmas 3.7 and 3.8 of Ref.~\cite{evans_graphs_2016}, reproduced below:

\begin{lemma}[Exogenize Latent Nodes~\protect{\cite[Lemma 3.7]{evans_graphs_2016}}]
\label{lemma_exogenize_latents}
Let $\mathcal{G}$ be a pDAG, and let ${\tt EndoLNodes}(\mathcal{G})\subseteq\lat(\mathcal{G})$ be the set of latent nodes of $\mathcal{G}$ that are endogenous (i.e., they have one or more parents in $\cal G$).  Construct the pDAG ${\tt Exog}(\mathcal{G})$ as follows. For every $u\in {\tt EndoLNodes}(\mathcal{G})$, start from $\mathcal{G}$ and: (i) delete  all directed edges leading into $u$, (ii) add a directed edge from every parent of $u$ to every child of $u$. The pDAG ${\tt Exog}(\mathcal{G})$ so constructed is observationally equivalent to $\mathcal{G}$,  ${\tt Exog}(\mathcal{G})\cong  \mathcal{G}$.
\end{lemma}

\begin{lemma}[Remove Redundant Latent Nodes~\protect{\cite[Lemma 3.8]{evans_graphs_2016}}]
\label{lemma_remove_redundant_latents}
Let $\mathcal{G}$ be a pDAG where all latent nodes are exogenous.
Let ${\tt RedundLnodes}(\mathcal{G})\subset\lat(\mathcal{G})$ be a maximal subset of latent nodes such that  their set of children is already common-cause-connected by another latent node. 
(Note that there is ambiguity in the choice, since when two latent nodes have the same set of children, one can take either to be the redundant one.) Formally, we define such a subset of latent nodes of $\cal G$, denoted ${\tt RedundLnodes}(\mathcal{G})$, by the following two conditions:
(i) For every $u \in {\tt RedundLnodes}(\mathcal{G})$ there exists some distinct latent node  $v \in \lat(\mathcal{G})\setminus {\tt RedundLnodes}(\mathcal{G})$  such that $\ch(u) \subseteq \ch(v)$. (ii) ${\tt RedundLnodes}(\mathcal{G})$ is not a strict subset of any other set of latent nodes that obeys (i).  Let ${\tt RemoveRedund}(\mathcal{G})$ be the pDAG constructed by removing from $\mathcal{G}$ every node in ${\tt RedundLnodes}(\mathcal{G})$. The pDAG ${\tt RemoveRedund}(\mathcal{G})$ so constructed is observationally equivalent to $\cal G$, i.e.,  
${\tt RemoveRedund}(\mathcal{G}) \cong \mathcal{G}$. 
\end{lemma}

For any pDAG $\mathcal{G}$, one can find another pDAG whose latent variables are all exogenous and non-redundant that is observationally equivalent to $\mathcal{G}$. Namely, it is the pDAG ${\tt RemoveRedund}\circ {\tt Exog}(\mathcal{G})$, which we will call the \emph{{\tt RE}-reduction} of $\cal G$. Accordingly, the map {\tt RE{-}reduce} is defined as:
\begin{align}\label{REreductionmap}
	\text{{\tt RE{-}reduce}} := {\tt RemoveRedund}\circ {\tt Exog}
\end{align}
It is easy to see that many different pDAGs have the same {\tt RE}-reduction. A pDAG which is invariant under the {\tt RE{-}reduce} map, i.e., such that $\mathcal{G}=\text{{\tt RE{-}reduce}}(\cal G)$, will be called a \emph{{\tt RE}-reduced} pDAG.

As noted in~\cite{evans_graphs_2016}, this justifies the introduction of a new structure for studying realizable distributions for pDAGs: the \emph{marginalized DAG} (mDAG). The notion of an mDAG makes use of the concept of a \emph{simplicial complex}: 
\begin{definition}[Simplicial complex]
A simplicial complex over a finite set $V$ is a set $\mathcal{B}$ of subsets of $V$ such that
\begin{itemize}
	\item $\{v\}\in \mathcal{B}$ for all $v\in V$;
	\item If $A\subseteq B\subseteq V$ and $B\in \mathcal{B}$, then $A\in \mathcal{B}$.
\end{itemize}
The elements of $\mathcal{B}$ are called faces. The inclusion-maximal elements of a simplicial complex (the faces that are maximal in the order over faces induced by subset inclusion) are called facets.
\end{definition}

An mDAG is a pair $\mathfrak{G}=(\cal D,B)$, where $\cal D$ is a DAG and $\cal B$ is a simplicial complex over the set of nodes of $\cal D$. The DAG $\mathcal{D}$ is called the \emph{directed structure} of $\mathfrak{G}$. The nodes of $\cal D$ will also be referred to as the nodes of $\mathfrak{G}$. The edges of $\cal D$ will be referred to as the directed edges of $\mathfrak{G}$ and denoted ${\tt DirectedEdges}(\mathfrak{G})$.  Here, we will use the font $\mathfrak{G}$ to denote mDAGs, while the font $\cal G$ will continue to be used to denote DAGs and pDAGs. 

With this notion in hand, we can introduce the map $\mdag$ that associates an mDAG to each pDAG:

\begin{definition}[Map $\mdag$ from pDAGs to mDAGs]
Let $\mathcal{G}$ be a pDAG, and let  $\tilde{\mathcal{G}}$ be the {\tt RE}-reduction of $\mathcal{G}$, i.e., 
$\tilde{\mathcal{G}} = \text{{\tt RE{-}reduce}}(\mathcal{G})$. Recall that $\vis(\tilde{\mathcal{G}})=\vis({\mathcal{G}})$.
The \emph{mDAG associated with $\mathcal{G}$}, denoted by $\mdag(\mathcal{G})$, is the pair $\mathcal{(D,B)}$ where 
\begin{itemize}
	\item $\mathcal{D}$ is a DAG such that $\nodes(\mathcal{D})=\vis({\mathcal{G}})$ and whose edges correspond to the edges between the visible nodes of $\tilde{\mathcal{G}}$.
	\item $\mathcal{B}$ is a simplicial complex over $\vis({\mathcal{G}})$.        The facets of $\mathcal{B}$ are the maximal subsets of $\vis({\mathcal{G}})$ that 
	are children of the same latent node in $\tilde{\mathcal{G}}$, i.e., for each $u\in \lat(\tilde{\mathcal{G}})$, the set $\ch_{\tilde{\mathcal{G}}}(u)$ 
	is a facet of $\mathcal{B}$. 
\end{itemize}

\end{definition}

\begin{figure}[htbp]
\centering
\includegraphics[width=0.48\textwidth]{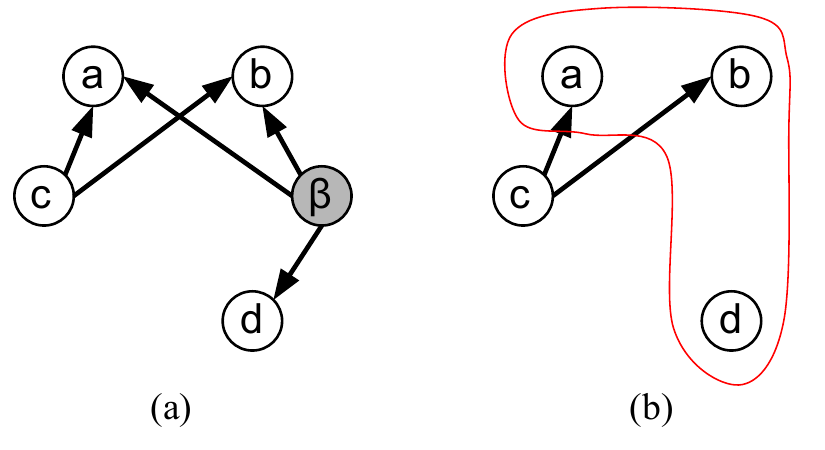}
\caption{(a) A {\tt RE}-reduced pDAG. (b) The mDAG associated with (a). This mDAG has simplicial complex
	$\mathcal{B}=\{\{a\},\{b\},\{d\},\{a,b\},\{a,d\},\{b,d\},\{a,b,d\}\}$. This is the mDAG associated with all of the pDAGs of Fig.~\ref{fig_example_lemmas}. The inclusion-maximal element of $\mathcal{B}$ are indicated by the red loop on the mDAG.} 
\label{fig_mDAG}
\end{figure}

\begin{figure}[htbp]
	\centering
	\includegraphics[width=0.48\textwidth]{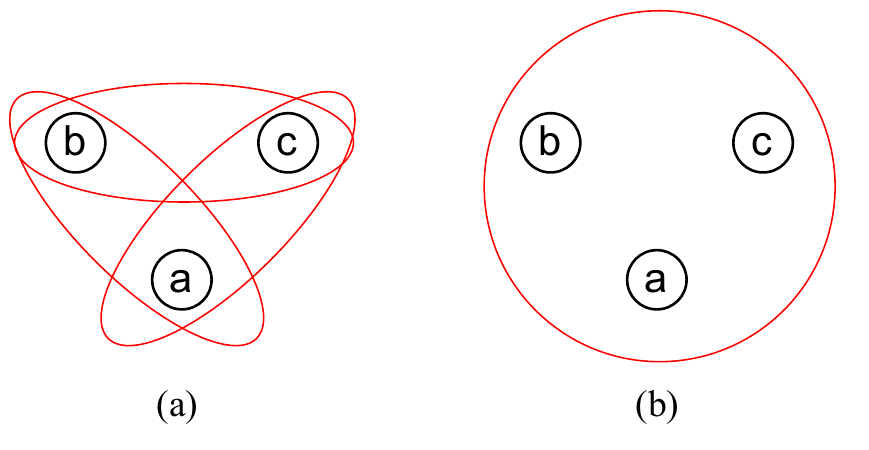}
	\caption{(a) mDAG with trivial directed structure and with simplicial complex $\mathcal{B}=\{\{a\},\{b\},\{c\},\{a,b\},\{a,c\},\{b,c\}\}$. (b)  mDAG with trivial directed structure and with simplicial complex $\mathcal{B'}=\{\{a\},\{b\},\{c\},\{a,b\},\{a,c\},\{b,c\},\{a,b,c\}\}$.The facets (inclusion-maximal elements of $\mathcal{B}$ and $\mathcal{B'}$) are indicated by red loops.} 
	\label{fig_triangle}
\end{figure}

The correspondence between a pDAG  and its associated mDAG is exemplified in  Fig.~\ref{fig_mDAG}: each latent node $u$ in the pDAG corresponds to a facet in the mDAG whose elements are the children of $u$ in the pDAG. 
One can conceptualize the mDAG as a hypergraph with two types of edges, namely, the directed edges and a set of undirected hyperedges, where the latter represent the facets of the simplicial complex and are depicted by red loops.\footnote{ Note that mDAGs are different from other hypergraph constructions, such as acyclic directed mixed graphs (ADMGs)~\cite{ADMGs}, insofar as mDAGs capture information about \emph{all} of the causal-compatibility constraints of the original pDAG, and not only its conditional independence constraints. For example, the two mDAGs of Fig.~\ref{fig_triangle} are associated with the same ADMG, because they have the same set of conditional independence constraints (in this case, the empty set). Graphically, while the hyperedges of an ADMG include only two nodes each, the hyperedges of the mDAG (i.e., facets of the simplicial complex) can include any number of nodes.  } 

We now define the notion of \emph{structural dominance} of one mDAG over another.

\begin{definition}[Structural Dominance of mDAGs]
\label{def_structural_dominance}
Let $\mathfrak{G}$ and $\mathfrak{G}'$ be two mDAGs with the same sets of nodes. $\mathfrak{G}$ is said to \emph{structurally dominate} $\mathfrak{G}'$ if the following pair of conditions hold: (i) the directed structure of $\mathfrak{G}'$ can be obtained from the directed structure of $\mathfrak{G}$ by dropping edges, ${\tt DirectedEdges}(\mathfrak{G}') \subseteq {\tt DirectedEdges}(\mathfrak{G})$, and  (ii) the simplicial complex of $\mathfrak{G}'$ can be obtained from the simplicial complex of $\mathfrak{G}$ by dropping faces, $\mathcal{B}' \subseteq \mathcal{B}$.  
\end{definition}

Note that the definition of structural dominance requires a subset inclusion relation for the {\em faces}, not the {\em facets} of the simplicial complexes. 
For example, in Fig.~\ref{fig_triangle}, the set of facets 
of the simplicial complex of the mDAG \ref{fig_triangle}(a) 
is not a subset of the set of facets of the simplicial complex  of \ref{fig_triangle}(b), but the set of {\em faces} do stand in a relation of subset inclusion to one another, so that mDAG \ref{fig_triangle}(b) structurally dominates mDAG \ref{fig_triangle}(a). 

It is convenient to define a map that takes an mDAG to the {\tt RE}-reduced pDAG that yields it. We will call this the \emph{canonical pDAG} associated with the mDAG:

\begin{definition}[Canonical pDAG]
Let $\mathfrak{G}=(\mathcal{D},\mathcal{B})$ be an mDAG. The map $\can$ is given by the following procedure: starting from $\cal D$, add one latent node $l$ for each \emph{facet} $A\in \cal B$ and add edges from $l$ to each element of $A$ so that $\ch(l)=A$. The resulting pDAG, $\can(\mathfrak{G})$, is the canonical pDAG associated with $\mathfrak{G}$.   
\end{definition}

\begin{figure}[htbp]
	\centering
	\includegraphics[width=0.48\textwidth]{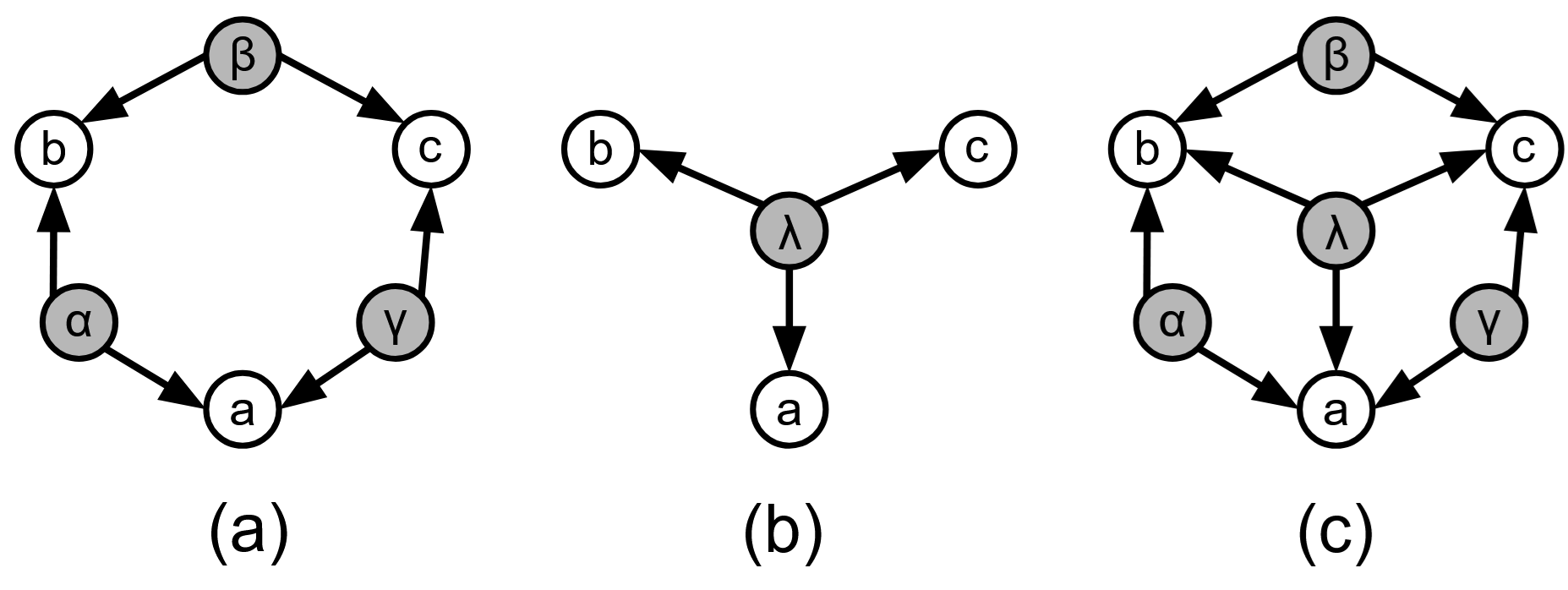}
	\caption{(a) Canonical pDAG of Fig.~\ref{fig_triangle}(a). (b)  Canonical pDAG of Fig.~\ref{fig_triangle}(b). (c) pDAG obtained by adding the latent nodes of (a) to (b). This is not a canonical pDAG. By Lemma~\ref{lemma_remove_redundant_latents}, (b) and (c) are observationally equivalent.} 
	\label{fig_triangle_pDAGs}
\end{figure}

Consider the canonical pDAG associated with the mDAG of Fig.~\ref{fig_triangle}(a), which is represented in Fig.~\ref{fig_triangle_pDAGs}(a), and the canonical pDAG associated with the mDAG of Fig.~\ref{fig_triangle}(b), which is represented in Fig.~\ref{fig_triangle_pDAGs}(b). Now, construct the pDAG of Fig.~\ref{fig_triangle_pDAGs}(c) by starting from Fig.~\ref{fig_triangle_pDAGs}(b) and adding to it all of the latent nodes of Fig.~\ref{fig_triangle_pDAGs}(a), as well as their connections to visible nodes. Clearly, Fig.~\ref{fig_triangle_pDAGs}(c) and Fig.~\ref{fig_triangle_pDAGs}(b) are observationally equivalent by Lemma~\ref{lemma_remove_redundant_latents}, because all of the added latent nodes are redundant. The pDAG of Fig.~\ref{fig_triangle_pDAGs}(a) is a subgraph (in the sense of Definition \ref{def_subgraph}) of  Fig.~\ref{fig_triangle_pDAGs}(c).

This observation is generic: Let $\mathfrak{G}$ and $\mathfrak{G}'$ be mDAGs such that $\mathfrak{G}'$ is structurally dominated by  $\mathfrak{G}$.  Construct the pDAG $\bar{\mathcal{G}}$ by starting from $\can(\mathfrak{G})$ and adding to it all of the latent nodes of $\can(\mathfrak{G}')$, as well as the corresponding edges connecting them to visible nodes.
Then, $\can(\mathfrak{G}')$
is a subgraph of  $\bar{\mathcal{G}}$.

Therefore, Lemma~\ref{lemma_remove_redundant_latents} together with the well-known fact that dropping edges from a pDAG results in an observationally dominated pDAG (see, for example, Proposition 3.3(b) in Ref.~\cite{evans_graphs_2016} or Theorem 26.1 in Ref.~\cite{henson_theory-independent_2014}) imply that:
\begin{lemma}[Structural dominance $\Rightarrow$ Observational dominance~\cite{evans_graphs_2016,henson_theory-independent_2014}]
\label{prop_edge_dropping}
Let $\mathfrak{G}$ and $\mathfrak{G}'$ be two mDAGs with the same sets of nodes. If $\mathfrak{G}$ structurally dominates $\mathfrak{G}'$, then it also observationally dominates it, i.e.,  $\mathfrak{G}\succeq\mathfrak{G}'$.
\end{lemma}

The results and definitions that we presented for pDAGs are extended to mDAGs through their associated canonical pDAG. For example, if the canonical pDAG of a certain mDAG is \algebraic, we say that the mDAG itself is \algebraic. 
When there is a  latent  confounder between nodes $a$ and $b$ in the canonical pDAG, i.e., when the simplicial complex of the mDAG has at least one facet that includes nodes $a$ and $b$, then we say that there is a \emph{latent confounder} between $a$ and $b$ in the mDAG.  When the directed structure of $\mathfrak{G}$ is trivial in the sense of containing no edges, then $\mathfrak{G}$ is said to be \emph{directed-edge-free}. The mDAGs of Fig.~\ref{fig_triangle} are examples of directed-edge-free mDAGs.
Similarly, when the simplicial complex of $\mathfrak{G}$ is trivial in the sense of all its facets being singleton sets,
then $\mathfrak{G}$ is said to be \emph{latent-confounder-free}. Clearly, if the pDAG $\mathcal{G}$ is latent-confounder-free, then the mDAG $\mdag(\cal G)$ is latent-confounder-free. Finally,  a distribution is realizable by an mDAG if it is realizable by its canonical pDAG, an mDAG observationally dominates another if the observational dominance relation holds for their canonical pDAGs, and so on.  A final definition concerning mDAGs that will be useful is presented below:

\begin{definition}[Subgraph of an mDAG over a subset of nodes]
	Let $\mathfrak{G}$ be an mDAG, and let $S\subseteq \nodes(\mathfrak{G})$ be a subset of its nodes. The subgraph of $\mathfrak{G}$ over $S$, denoted $\mathfrak{G}_S$, is the mDAG obtained by starting from $\mathfrak{G}$ and deleting all of the nodes in $ \nodes(\mathfrak{G})\setminus S$. 
\end{definition}

Note that $\mathfrak{G}_S$ is the same as the mDAG associated with the pDAG $\can(\mathfrak{G})_{\lat(\can(\mathfrak{G}))\cup S}$, that is, the  subgraph of $\can(\mathfrak{G})$ obtained by deleting all of the \emph{visible} nodes that are not in $S$.

Lemmas \ref{lemma_exogenize_latents} and \ref{lemma_remove_redundant_latents} imply that whenever two pDAGs $\mathcal{G}$ and $\mathcal{G}'$ are associated with the same mDAG, they are observationally equivalent.  
Note, however, that if two pDAGs $\mathcal{G}$ and $\mathcal{G}'$ are associated to {\em distinct} mDAGs, they may or may not be observationally equivalent.  That is, a difference in the associated mDAGs is necessary but not sufficient for observational inequivalence of pDAGs.

The set of mDAGs that have a fixed number of nodes can, therefore, be organized into equivalence classes under the observational equivalence relation. These classes are called \emph{observational equivalence classes}. In turn, the observational dominance relation defines a partial order among the observational equivalence classes, which we call the \emph{observational partial order}. Our goal in this work is to  characterize this partial order. 

\begin{figure}[h!]
\centering
\includegraphics[width=0.45\textwidth]{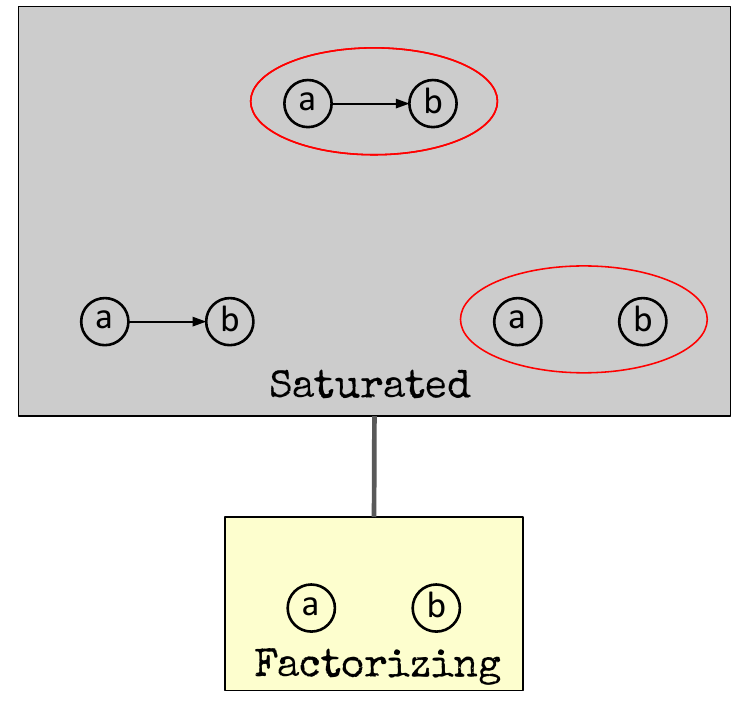}
\caption{Observational partial order for the set of two-node mDAGs that are consistent with a fixed nodal ordering. The line between the two classes indicates that the mDAGs of the {\tt Saturated} class observationally dominate the mDAG of the {\tt Factorizing} class.}
\label{fig_2_node_classes}
\end{figure}

As will be justified in Section~\ref{sec_temporal_order}, here we will only discuss the observational equivalence classes and the observational partial order that hold for a set of mDAGs that are consistent with a fixed nodal ordering, in the sense of Def.~\ref{def_nodal_ordering}. The simplest example of an observational partial order of this type is the one for mDAGs with two nodes. There are four mDAGs of two nodes that are consistent with a fixed nodal ordering. Three of these are saturating, i.e, the set of probability distributions over the visible variables that are realizable by them
is the set of \emph{all} such distributions. 
The remaining mDAG is the one where the two nodes are disconnected. 
For it, the set of realizable 
distributions is the set of all \emph{factorizing} distributions.
Therefore, the partial order is the one depicted in Fig.~\ref{fig_2_node_classes}: the three mDAGs with connected nodes make up one observational equivalence class, 
termed the {\tt Saturated} class, while the mDAG with disconnected nodes defines the {\tt Factorizing} class, and the former observationally dominates the latter.

\section{Objectives and Methods}
\label{sec_objectives}

\subsection{Restriction of scope to sets of mDAGs consistent with a fixed nodal ordering}
\label{sec_temporal_order}

In this work, we will take the nodal ordering of the visible variables to be fixed, and we will only compare candidate causal structures for which the visible nodes are consistent with this nodal ordering. There are several reasons why we restrict the set of causal structures appearing in our classification in this way. As we argued in Ref.~\cite{ansanelli2024everything}:

\begin{quote}
	Clearly, if the variables of interest are temporally localized (and we do not need to take into account relativity theory, such as the possibility of space-like separation), then they will necessarily be temporally ordered.  The reason typically given for why fixed nodal ordering should {\em not} be presumed~\cite[Section 7.5.1]{causality_pearl} is that the variables of interest may fail to be temporally localized, referring instead to properties that persist over time. An individual's health and their exercise regime provides a good example. However, whenever the variables of interest are of this type, they can have a mutual influence on one another, and consequently one cannot restrict attention to directed graphs that are {\em acyclic} when considering the possibilities for the causal structure holding between them. Thus, if one is contemplating a causal discovery algorithm that returns only acyclic graphs (as we are doing here), then one should not apply it to variables that fail to be temporally localized. 
\end{quote}

One could still argue that, although the visible variables are temporally localized, for some reason we do not know the temporal order in which they were probed. However, one needs to have some knowledge about the probing scheme to be confident that it consists purely of passive observations. Situations where this is the case but one remains ignorant of when each variable was probed are likely to be the exception rather than the rule.

\subsection{Special Cases: Latent-confounder-free and Directed-edge-free mDAGs}
\label{sec_confounder_free_directededge_free}

In this section, we discuss two special cases whose observational partial order is completely solved for any number of visible nodes: the case of latent-confounder-free mDAGs and the case of directed-edge-free mDAGs.

Most of the causal inference literature studying observational equivalence discusses only latent-confounder-free mDAGs, or, equivalently, only latent-confounder-free pDAGs~\cite{Verma_Pearl_equivalence1990,Andersson, Chickering_greedy}. That literature, however, classifies the set of such mDAGs for \emph{all} nodal orderings of the visible nodes rather than a fixed one. When we restrict our attention to one nodal ordering of the visible nodes, the organization of latent-confounder-free mDAGs into observational equivalence classes is trivial: for a given nodal ordering of the nodes, if two latent-confounder-free mDAGs are different then they are observationally inequivalent. This fact seems to be well-known; it is noted, for instance, below Theorem 1.2.8 in Ref.~\cite{causality_pearl}. We also provided a proof of this fact in the Appendix C of our previous work~\cite{ansanelli2024everything}. Here, we adapt this proof to show something more, that finding the observational \emph{partial order} is also simple when one is restricted to the latent-confounder-free mDAGs for a fixed nodal ordering of the nodes:
\begin{proposition}[Consequence of Ref.~\protect{\cite[Appendix C]{ansanelli2024everything}}]
	Let $\mathfrak{G}$ and $\mathfrak{G}'$ be two latent-confounder-free mDAGs that are consistent with the same nodal ordering. Then, $\mathfrak{G}$ observationally dominates $\mathfrak{G}'$ if and only if it structurally dominates it.
\end{proposition}
\begin{proof}
	The ``if'' side follows from Lemma~\ref{prop_edge_dropping}. To prove the ``only if'' side, assume that $\mathfrak{G}$ \emph{does not} structurally dominate $\mathfrak{G}'$. Since they are latent-confounder-free, this implies that there is an arrow between two nodes that is present in $\mathfrak{G}'$ but not in $\mathfrak{G}$. In Appendix C of  Ref.~\cite{ansanelli2024everything}, we showed that this implies that there is a d-separation relation that is implied by  $\mathfrak{G}$ but not by  $\mathfrak{G}'$.  By point 2 of Theorem~\ref{th:d-sep}, this implies that there is a probability distribution $P$ realizable by  $\mathfrak{G}'$ that violates the conditional independence relation that is implied by this d-separation relation.  By point 1  of Theorem~\ref{th:d-sep}, $P$ is not realizable by $\mathfrak{G}$.	
	Therefore,   $\mathfrak{G}$  \emph{does not}  observationally dominate $\mathfrak{G}'$. 
\end{proof}

Therefore, the observational partial order of the set of latent-confounder-free mDAGs with a fixed nodal ordering is simply the partial order defined by structural dominance. The case of for 3-node and 4-node mDAGs of this type are shown in Figures 7.3(a) and 7.5 of Ref.~\cite{ansanelli2024everything}.

For the case of directed-edge-free mDAGs, the concept of nodal ordering is not relevant, since none of the visible nodes is an ancestor of any other visible node. In Proposition 6.8 of Ref.~\cite{evans_graphs_2016}, it was shown that two different directed-edge-free mDAGs are always observationally inequivalent. Here, we extend this idea to show that observational dominance for directed-edge-free mDAGs is  determined by structural dominance:
\begin{proposition}[Consequence of Ref.~\protect{\cite[Example 2]{steudel_ay_2015}}]
		Let $\mathfrak{G}$ and $\mathfrak{G}'$ be two directed-edge-free mDAGs. Then, $\mathfrak{G}$ observationally dominates $\mathfrak{G}'$ if and only if it structurally dominates it.
	\label{directed_edge_free_dominance}
\end{proposition}
\begin{proof}
	The ``if'' side follows from Lemma~\ref{prop_edge_dropping}. To prove the ``only if'' side, assume that $\mathfrak{G}$ \emph{does not} structurally dominate $\mathfrak{G}'$. Since they are directed-edge-free, this implies that there is a face that is present in the simplicial complex of $\mathfrak{G}'$ but not in the simplicial complex of $\mathfrak{G}$. That is, there is a set $S$ of visible nodes in $\can(\mathfrak{G}')$ that shares a latent common cause, while it does not share a latent common cause in $\can(\mathfrak{G})$. Since there are no arrows between visible nodes, this implies that the nodes in $S$ share a common ancestor in  $\can(\mathfrak{G}')$ but not in  $\can(\mathfrak{G})$.
	
	In Example 2 of Ref.~\cite{steudel_ay_2015}, it was shown that a distribution where all the variables $\{X_1,...,X_n\}$ are perfectly correlated, i.e., where the only points with nonzero weight are those where $X_1=X_2=...=X_n$, is only realizable by causal structures where all the nodes corresponding to such variables share a common ancestor.\footnote{ Ref.~\cite{steudel_ay_2015} does not consider causal structures with latent variables. However, this result is easily generalizable to that case, because a pDAG $\cal G$ with latent variables can realize perfect correlation between the visible variables $\{X_1,...,X_n\}$ if and only if the latent-confounder-free pDAG obtained by transforming all of the latent nodes of $\cal G$ into visible nodes can also realize it. } Therefore,  $\can(\mathfrak{G}')$ can explain perfect correlation between the variables of $X_S$, while  $\can(\mathfrak{G})$ cannot. Thus, $\mathfrak{G}$  \emph{does not}  observationally dominate $\mathfrak{G}'$. 
\end{proof}

Therefore, the observational partial order of directed-edge-free mDAGs is the partial order defined by structural dominance. The case of for 3-node and 4-node mDAGs of this type are shown in Figures 7.3(b) and 7.6 of Ref.~\cite{ansanelli2024everything}.

\subsection{Full Solution of Observational Partial Order for 3-node mDAGs}
\label{sec_3_observed_nodes}

Now, let us turn to the problem of classifying the entire set of mDAGs with a given number of nodes and a fixed nodal ordering. As we will see, this problem is still open for a general number of nodes: we do not yet know the necessary and sufficient conditions for two mDAGs to be observationally equivalent or to observationally dominate each other in the case of an arbitrary number of nodes.

For mDAGs with two nodes, of course, this problem is completely solved but the solution is  rather trivial, as we saw in Fig.~\ref{fig_2_node_classes}.  For mDAGs with three nodes, the problem is also completely solved, but the solution is more interesting in this case. 
In this section, we will describe this order; the proofs of the ordering relations are postponed to Section~\ref{sec_back_to_3visible}.

\begin{figure*}[h!]
\centering
\includegraphics[width=0.9\textwidth]{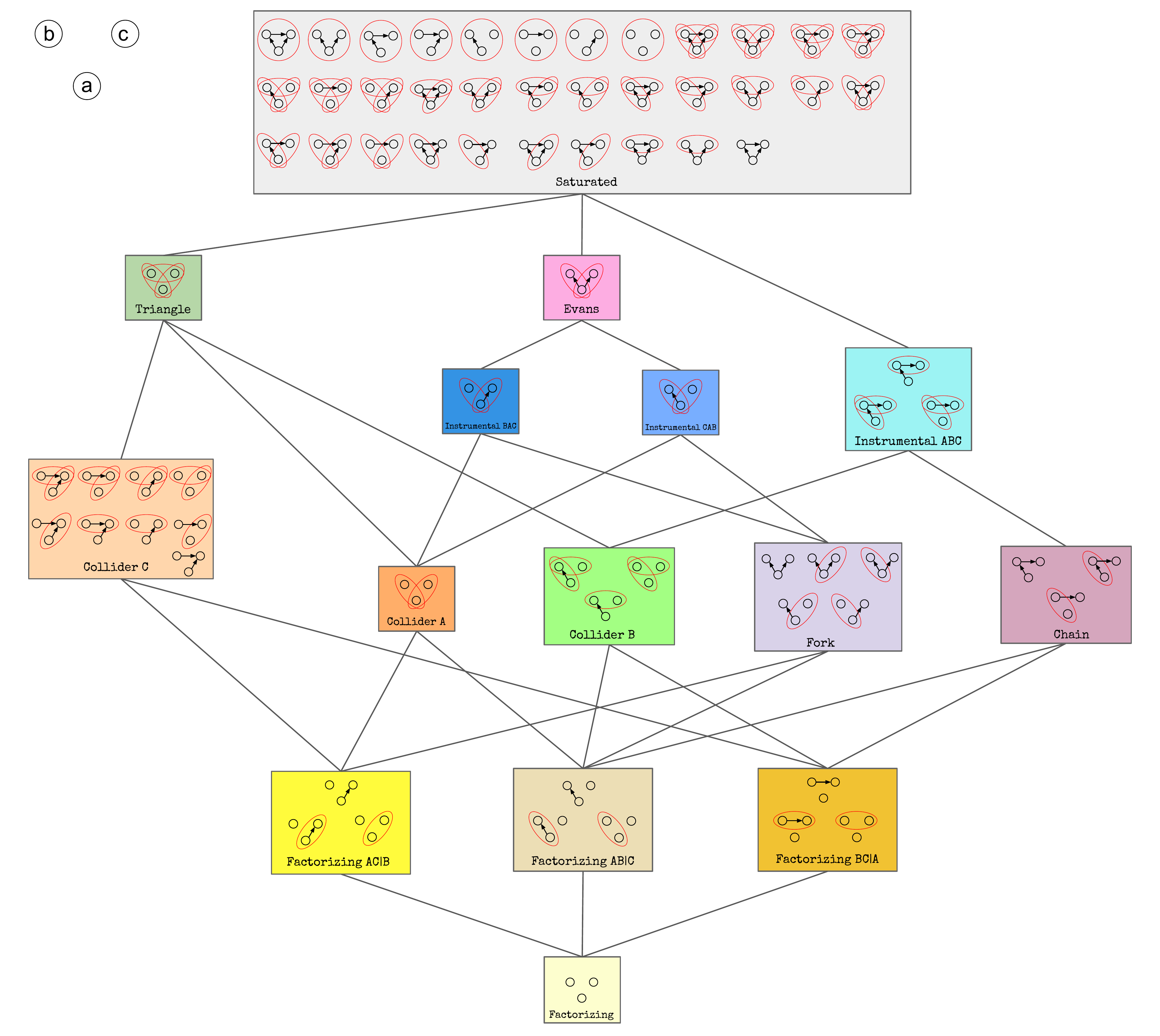}
\caption{The 72 mDAGs with three visible nodes with nodal ordering $(a,b,c)$ organized into observational equivalence classes and the observational dominance relations that hold between these classes. In each mDAG of this figure, the bottom node is $a$, the top-left node is $b$ and the top-right node is $c$.}
\label{fig_temporally_ordered_3_nodes}
\end{figure*}

There are 72 3-node mDAGs consistent with a fixed nodal ordering.
In Fig.~\ref{fig_temporally_ordered_3_nodes}, we group these into observational equivalence classes, and present the observational dominance relations that hold among the classes. In order to avoid having to explicitly label the nodes throughout the figure, we adopt a simplifying convention wherein the position of the node in a certain mDAG indicates its label: in each mDAG the bottom node is $a$, the top-left node is $b$ and the top-right node is $c$. All of the mDAGs of this figure follow the nodal ordering $(a,b,c)$. In this figure, the highest elements in the observational partial order appear at the top of the diagram and the lowest elements appear at the bottom, and if one class dominates another with no classes falling between them in the observational partial order, we draw a line from the former to the latter. 

\subsection{Significance of observational partial order for causal discovery}\label{significanceforcausaldiscovery}

We recall the problem of causal discovery:
the input data is a sample from a probability distribution over visible variables obtained from passive observations, and the objective is to  infer which causal structures can provide an  
explanation of this data.  In general, it is not possible to single out the pDAG or even the mDAG that generated the data, because many different causal structures might be able to realize it.  Nonetheless, one can seek to answer the question: 
what is \emph{everything} that can be learned about the underlying causal structure of a given phenomenon when we only have access to passive observations?   We here explain how determining the observational partial order is useful for this problem. 

We begin with a limiting case of the causal discovery problem, namely, one wherein the  sample size is infinite rather than finite.  
  Only in this case 
do the relative frequencies computed from the sample reflect the true probability distribution, whereas in the case of a  finite 
 sample, the relative frequencies merely yield an {\em estimate} of the true probability distribution.  In other words, we begin by considering the version of the causal discovery problem wherein the input is the true probability distribution.

The most basic respect in which the observational partial order is relevant to this problem is the following. If one determines that a distribution is not realizable by a particular mDAG $\mathfrak{G}$, then one can conclude that it is also not realizable by any of the mDAGs that are {\em equivalent to or below} $\mathfrak{G}$ in the observational partial order.
  Similarly, if one determines that a given distribution {\em is} realizable by a particular mDAG $\mathfrak{G}'$, then one can conclude that it is also realizable by any of the mDAGs that are {\em equivalent to or above} $\mathfrak{G}'$ in the observational partial order.

It is useful here to make use of the notions of {\em upward closure} and {\em downward closure} of a set of elements in a partial order. 
If $S$ denotes a set of elements of a partial order, then the upward closure of $S$, denoted ${\rm UC}(S)$, includes every $x$ such that $x\succeq y$ for some $y\in S$. 
 Similarly, the downward closure of $S$, denoted ${\rm DC}(S)$,  includes every $x$ such that $x\preceq y$ for some $y\in S$. 

Using these notions, if a distribution is known to be realizable by all of the mDAGs of a set $S$, then one can infer that it is also realizable by all of the mDAGs of ${\rm UC}(S)$. 
Similarly, if a distribution is known to {\em not} be realizable by any mDAG in a set $S'$, then one can infer that it is also not realizable by any mDAG in ${\rm DC}(S')$. 
 If the union of ${\rm UC}(S)$ and ${\rm DC}(S')$ is not the full set of mDAGs, then the remaining mDAGs are those for which one remains uncertain whether they can realize the distribution or not. Note that to specify what one knows about realizability, it suffices to specify the least elements of ${\rm UC}(S)$ and the greatest elements of ${\rm DC}(S')$.  In general, each of these will be a set of incomparable elements.

The observational partial order of 3-node mDAGs, depicted in Fig.~\ref{fig_temporally_ordered_3_nodes}, helps to illustrate the idea. For instance,  it could be known that a probability distribution over three variables is realizable by {\tt Collider C} but not by {\tt Instrumental CAB} or {\tt Instrumental ABC}.
 In this case, one can infer that it is realizable by the upward closure of {\tt Collider C}, namely, {\tt Collider C}, {\tt Triangle} and {\tt Saturated}, and one can infer that it is not realizable by the downward closure of the set consisting of {\tt Instrumental CAB} and {\tt Instrumental ABC}.  Meanwhile it  remains unknown whether it is realizable by any mDAG that is in neither of these sets, namely, {\tt Instrumental BAC} and {\tt Evans}.

In the case where we learned everything there is to learn about the causal structure from passive observations, we have that the upward closure  ${\rm UC}(S)$ of a set $S$ of mDAGs \emph{can} realize our probability distribution, and that all of the mDAGs outside of  ${\rm UC}(S)$ \emph{cannot} realize it. In general, therefore, the set of causal structures that could in principle realize a given 
probability distribution over the visible variables (in a passive observation scheme) constitutes the elements of the upward closure (within the partial order) of a set $S$ of 
 observational equivalence classes of
  mDAGs  
 

Without appealing to interventional experiments or to any model selection principle such as faithfulness (which will be discussed below), the most we can say about a given observational distribution is that it
 can be realized by all of the mDAGs in a set of the form ${\rm UC}(S)$.

This discussion assumed the limit of  infinite sample size, 
but the same ideas extend to the case of  finite sample size. 
 Using standard techniques of model selection, we can sometimes conclude that a causal structure  $\mathfrak{G}$ {\em underfits} the  finite sample of data 
 (e.g., its chi-squared error is too large).  In this case, so too will all the causal structures that are  below it in the observational partial order. The reason is as follows.  For  $\mathfrak{G}$ to {\em underfit} the  finite sample  means that the  finite sample  is improbable relative to all of the probability distributions realizable by $\mathfrak{G}$, i.e., that it is unlikely that the  finite sample  came 
 from one of the  probability distributions realizable by $\mathfrak{G}$.  
 But for all of the causal structures  below $\mathfrak{G}$ in the observational partial order, the set of probability distributions that they can realize is a {\em strict subset} of those that can be realized by $\mathfrak{G}$, and therefore the  finite sample 
  must be improbable relative to all these distributions as well. By a similar reasoning, if an mDAG \emph{does not} underfit the data, then all of the mDAGs that are above it in the observational partial order will also not underfit the data. 
 
 In summary, if a set $S'$ of mDAGs underfits  the finite sample of data, 
  we discard all of the mDAGs in  ${\rm DC}(S')$ as candidate causal structures to explain the data. If a set $S$ of mDAGs does not underfit the data, then all elements of ${\rm UC}(S)$ are candidate causal structures.

\subsubsection{Using model-selection principles to adjudicate between possible candidates}

We established that the upward closure of a set of mDAGs in the observational partial order is the most we can learn about the underlying causal structure when we only have access to passive observations. Now, we will discuss how it is possible to make finer-grained estimates of the causal structure by appealing to certain model-selection principles. We begin with the case of  an infinite sample of data. 
  

  Among all of the mDAGs that \emph{can} realize a distribution, one might wonder whether any one of them provides as good of a causal explanation as any other.  It is generally thought that this is not the case and that one should prefer causal structures that are \emph{lower} in the observational partial order.

For instance, in Pearl's textbook on causality~\cite{causality_pearl}, he argues for this on the grounds that such causal models are 
more falsifiable.\footnote{For instance, on p.~46 of Ref.~\cite{causality_pearl}, he states: 
``such theories are more constraining and thus more falsifiable; they provide the scientist with less opportunities to overfit the data `hindsightedly' and therefore command greater credibility if a fit is found.''}
Indeed, Pearl does not introduce any language for referring to observational dominance besides the language of preference.  (Pearl's order relation of preference is opposite to the order relation of observational dominance in the sense that $\mathfrak{G}$ is preferred to $\mathfrak{G}'$ just in case $\mathfrak{G}'$ observationally dominates $\mathfrak{G}$.)

According to this proposal, the causal structures providing the best explanation of the distribution are the  {\em least elements} in the upward-closed set of classes that can realize it.  
This is referred to as the principle of {\em minimality} in Ref.~\cite{causality_pearl}  (see Definition 2.3.4).

It is worth noting that preferring causal models that satisfy the principle of {\em faithfulness}~\cite{spirtes2001causation} (referred to as {\em stability} in Ref.~\cite{causality_pearl} and {\em no fine-tuning} in Ref.~\cite{Wood_and_Spekkens}) is also an instance of preferring classes lower in the observational partial order.

In the case of  a finite sample of data, 
  it is also possible to adjudicate between a set of hypotheses about the causal structure each of which \emph{do not} underfit the data. It suffices  to use a standard technique of statistical model selection: the train-and-test methodology. 
In the train-and-test methodology, the observational data is divided into two sets, the first one being the \emph{training set} and the second one being the \emph{test set}. 
For each of the candidate causal structures, we find the parameters 
 that provide the best fit to the training set. 
 One then compares the predictions of this best-fit model to the test set to compute the test error. 
One can then compare the test error of different causal structures to compare  how well the best-fit model of each causal structure generalizes to new data, that is, to compare  their predictive power. 

Consider a comparison of two causal structures, $\mathfrak{G}$ and $\mathfrak{G}'$.  If $\mathfrak{G}'$ gives a higher test error than $\mathfrak{G}$ but a lower training error, then this suggests that  the causal model associated to $\mathfrak{G}'$ is mistaking statistical fluctuations that appear in the training data for a {\em stable signal}, i.e., a relationship that would reappear in fresh samples. 

 For example, imagine that two variables $X_a$ and $X_b$ are associated with nodes $a$ and $b$ that are d-separated given the empty set in  $\mathfrak{G}$, but are connected by an arrow $a\to b$ in  $\mathfrak{G}'$. 
Suppose that $\mathfrak{G}$ is the ground truth. In this case, one would find that
 the variables $X_a$ and $X_b$ are statistically independent in the limit of  an infinite sample of data. 
  In the  finite sample constituting the 
   training data, however, statistical fluctuations may be such that these variables  exhibit some statistical dependence. While all models on $\mathfrak{G}$ are forced to account for this dependence as a  fluctuation,  models on $\mathfrak{G}'$ might identify it as  a signature of a causal influence of $a$ on $b$. This in turn might make the latter models perform worse on the test data, because the test data has different fluctuations from the training data. When this happens,  we say that $\mathfrak{G}'$ \emph{overfits} the data relative to $\mathfrak{G}$.\footnote{ This is similar to the way in which the error of a polynomial fit to training data can be made to decrease by increasing the order of the polynomial, but because this low error is achieved by treating statistical fluctuations as real features, the fit will tend to have a large error when predicting the test data.} This train-and-test methodology was applied in Ref.~\cite{daley_experimentally} to adjudicate between different causal explanations of  a finite sample of data  
    in a quantum experiment that violated Bell inequalities.

We now turn to the significance of the observational partial order for  adjudicating between different candidate causal structures that \emph{do not} underfit  the finite sample of data. 

 Suppose $\mathfrak{G}'$ is  above $\mathfrak{G}$ is the observational partial order. If $\mathfrak{G}'$ overfits the data relative to $\mathfrak{G}$, then everything {\em above} $\mathfrak{G}'$ in the order {\em also} overfits the data relative to $\mathfrak{G}$.  The reason is that the extra expressive power of $\mathfrak{G}'$ relative to $\mathfrak{G}$ (i..e., the additional probability distributions realizable by $\mathfrak{G}'$) is what must 
 lead $\mathfrak{G}'$ to mistake statistical fluctuations  in the training data for stable signals that would reappear in a fresh sample.  But this extra expressive power is also present in all of the structures  above $\mathfrak{G}'$ in the observational partial order, and so these must be led to make the same mistake.

Note that if $\mathfrak{G}$ fits  (i.e., does not underfit)  the  finite sample of data 
 and $\mathfrak{G}'$ is a causal structure higher in the observational order, then $\mathfrak{G}'$ {  does not necessarily} overfit the data relative to $\mathfrak{G}$ simply by virtue of having additional expressive power.   This is because the extra expressive power may be of a type that does not lead $\mathfrak{G}'$ to mistake statistical fluctuations for  a stable signal.  
   In this case,  $\mathfrak{G}$ and $\mathfrak{G}'$ will have roughly {\em equal} test errors. 
This point was emphasized in Ref.~\cite{daley2024reply}, in response to a contrary claim (that one model being  more expressive than another generically {\em did} cause the first to overfit the data relative to the second.)
Nonetheless, it is worth considering the question of the circumstances in which one causal structure {\em does} tend to overfit the data relative to a causal structure that is directly below it in the observational partial order.

One such circumstance is if the lower causal structure implies an equality constraint on the realizable distributions that does not hold in the higher causal structure.  Suppose that data is generated from  the lower causal structure.  Because it is  a finite sample of data, 
 the equality constraint will generically not be satisfied exactly---there will be statistical fluctuations away from its exact satisfaction.  Because the higher causal structure can realize distributions that violate the equality constraint, it can fall into the trap of mistaking such fluctuations for  stable signals.  

Consider, for example, the pair of observational equivalence classes {\tt Collider C} and {\tt Triangle}. 
The class  {\tt Collider C}  presents the d-separation relation $a\perp_{d} b$, so the set of  distributions on three variables that are realizable by causal structures in that class satisfy 
$X_a \dbot X_b$, that is, the equality constraint $P(X_a X_b) = P(X_a)P(X_b)$.  This forces the probability distributions to lie on an algebraic variety that has a smaller dimension than the full space of such distributions.  The set of distributions realizable by the class {\tt Triangle}, on the other hand, is \emph{not} confined to this variety. In fact, since  {\tt Triangle} exhibits no equality constraints, its set of realizable distributions has the same dimension as the full space of distributions on three variables.  Now, suppose  the finite sample of   
 data is generated by a causal structure in the class {\tt Collider C}. Due to statistical fluctuation, the relative frequencies will generically  describe a point slightly outside of this variety.  A causal structure in the {\tt Triangle} class can fit this  finite sample of 
  data to a probability distribution that is outside of this variety, and thereby  interpret the statistical fluctuation as a  stable signal. 	
 (Clearly, the {\tt Saturated} class is \emph{also} prone to such overfitting, given that the set of distributions realizable by it are also not confined to this variety). 
 An example of this type, but involving a pair of 4-node mDAGs,
 is discussed in Ref.~\cite{daley_experimentally} where such a train-and-test methodology is implemented on  data from a quantum experiment. 

A second circumstance in which one causal structure may overfit  a finite sample of 
 data relative to a causal structure that is directly below it in the observational partial order is when the two are distinguished by an inequality constraint.  An inequality constraint does not lower the dimension of the space of realizable distributions, but it creates a boundary below which the distribution is realizable, and above which it is not. In this case, overfitting can arise when the causal structure generating the data implies an inequality constraint, but statistical fluctuations are taking the  finite-sample 
  relative frequencies outside the boundary.

Suppose, for instance, that the {\tt Instrumental BAC} class fits the data well, while the {\tt Evans} class overfits it relative to {\tt Instrumental BAC}. Because {\tt Instrumental BAC} only has inequality constraints, it must be that statistical fluctuations away from their satisfaction in the  finite sample of 
 data are being mistaken by  {\tt Evans}  for  a stable signal,  
  and leading to a high test error. 

Of course, the data generated must be very close to the boundary described by the inequality constraint in order for statistical fluctuations to lead to a violation of this constraint in the  finite-sample 
 relative frequencies.  Consequently, we do not expect this circumstance to arise for generic parameter values (most of which lead to distributions that lie far from the boundary).
The possibility of this type of overfitting is discussed further in Appendix A.2 of Ref.~\cite{daley2024reply}.

\subsubsection{Attesting to the presence of inequality constraints}
\label{sec_attesting_algebraic}

Another reason that classifying mDAGs into observational equivalence classes is useful is to identify which ones are \nonalgebraic. In Ref.~\cite{Evans2023} it was proven that an mDAG is \algebraic \emph{if and only if} it is observationally equivalent to some latent-confounder-free mDAG. Note that this result does not require the latent-confounder-free mDAG in question to be in the set of mDAGs consistent with a fixed nodal ordering that we have been considering in our observational partial order. In Fig.~\ref{fig_temporally_ordered_3_nodes}, for example, the class {\tt Collider B} does not include any latent-confounder-free mDAG, but it is nevertheless \algebraic. This is so because its elements are observationally equivalent to the latent-confounder-free mDAG $a\to b \gets c$, that is not shown in the figure because it is not consistent with the fixed nodal ordering $(a,b,c)$. To be able to easily refer to this idea later on, we formulate it as a proposition:

\begin{proposition}[Trivial consequence of Ref.~\protect{\cite[Theorem 1.1]{Evans2023}}]
	An observational equivalence class of mDAGs consistent with a fixed nodal ordering is \algebraic if and only if its mDAGs are observationally equivalent to a latent-confounder-free mDAG, which might not be consistent with the fixed nodal ordering.

	\label{prop_temporal_algebraic}
\end{proposition}

Using Proposition~\ref{prop_temporal_algebraic} to analyze the classes of Fig.~\ref{fig_temporally_ordered_3_nodes},  we can see that among equivalence classes of 3-node mDAGs consistent with a fixed nodal ordering,  the only \nonalgebraic ones are {\tt Triangle}, {\tt Evans}, {\tt Instrumental BAC}, {\tt Instrumental CAB} and {\tt Instrumental ABC}.  (The \nonalgebraicness of all of these mDAGs was in fact already known~\cite{Fritz_2012,Pearl_Instrumental,evans_graphs_2016}).

\subsection{How to Express Partial Information about the Observational Equivalence Partition}
\label{sec_proven_partitions}

In Fig.~\ref{fig_temporally_ordered_3_nodes}, we presented the complete observational partial order of 3-node mDAGs consistent with a fixed nodal ordering. This was only possible because, as will be made explicit in Section~\ref{sec_back_to_3visible}, the set of rules to show observational dominance and nondominance that are currently known happen to be sufficient to completely determine the observational partial order in this case. However, in general there is not yet a set of rules that is proven to be necessary and sufficient. As we will see, there are sets of 4-node mDAGs consistent with a fixed nodal ordering whose observational equivalence remains uncertain. 

The first step towards obtaining an observational partial order is to find the observational equivalence classes. These classes establish a partition of the set of all n-node mDAGs consistent with a fixed nodal ordering, which we will call the \emph{observational equivalence partition}.  An element of a partition is called a \emph{block}. This section will give the means to express partial information about the observational equivalence  partition, for cases where we cannot present its full characterization.

\begin{figure*}[h!]
\centering
\includegraphics[width=0.9\textwidth]{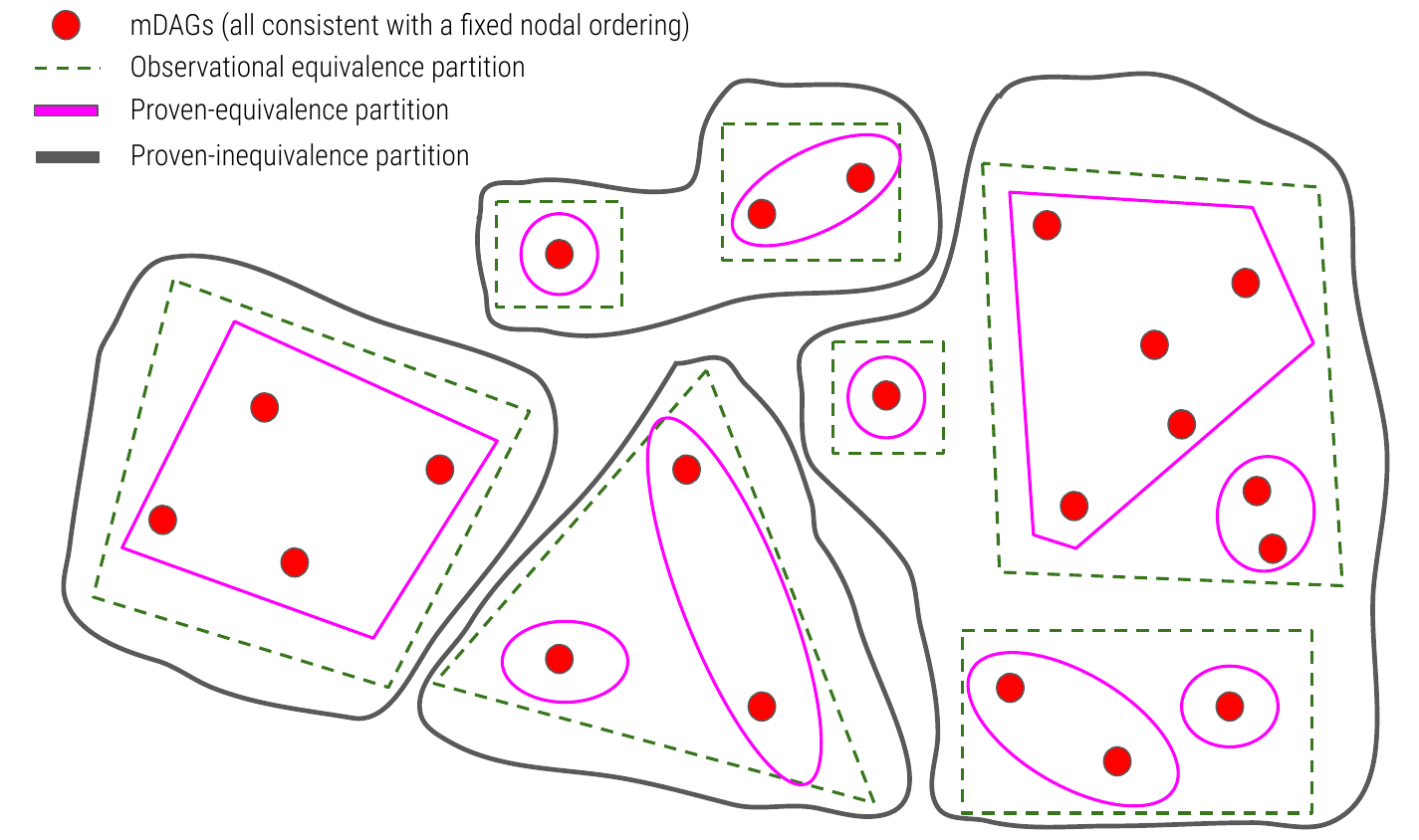}
\caption{The proven-equivalence partition (pink loops) is found by applying the equivalence rules of Section \ref{sec_show_equivalence}. The proven-inequivalence partition (gray loops) is found by the inequivalence rules of Section \ref{sec_show_inequivalence}. The final goal is to find the observational equivalence partition (green dashed loops). }
\label{fig_veinn}
\end{figure*}

Fig.~\ref{fig_veinn} illustrates how we express partial information about the observational equivalence partition. In this figure, the set of red points represent the set of n-node mDAGs consistent with a fixed nodal ordering, and the green dashed loops represent the observational equivalence partition. Partial information about the observational equivalence partition can be given in the form of  two other partitions, which we term the \emph{proven-equivalence partition}, where two mDAGs belong to \emph{the same} block of the partition when they are proven to be equivalent, and the \emph{proven-inequivalence partition}, where two mDAGs belong to \emph{different} blocks of the partition when they are proven to be inequivalent. Examples are depicted in Fig.~\ref{fig_veinn} by the pink and gray loops respectively. 

Each of the latter partitions can be built up in a step-wise fashion. For the proven-equivalence partition, one begins with a full partition of the mDAGs consistent with a fixed nodal ordering into singleton sets. When one proves that two mDAGs are observationally equivalent, one can merge the two blocks containing these mDAGs into a single block. The successive application of different rules for proving observational dominance (which will be presented in Section~\ref{sec_show_equivalence})  reveals more and more observational equivalences, thereby merging blocks of the proven-equivalence partition and lowering the total number of such blocks. This number is an upper bound on the number of observational equivalence classes. For a given set R of dominance-proving rules, the proven-equivalence partition one can obtain from this set is referred to as the proven-equivalence partition \emph{induced by R}. 

For the proven-inequivalence partition, one begins with a partition having all mDAGs in a single block. When one proves that two mDAGs are observationally inequivalent, one proves that they must be in different blocks of the proven-inequivalence partition. In fact, one thereby proves that the blocks of the proven-equivalence partition to which each of these mDAGs belong are observationally inequivalent. In other words, the blocks of the proven-inequivalence partition are always coarse-grainings of the blocks of the proven-equivalence partition. 

Successive application of different rules for proving observational nondominance (which will be presented in Section~\ref{sec_show_inequivalence}) reveals more and more observational inequivalences, thus splitting blocks of the proven-inequivalence partition and raising the total number of such blocks. This number is a lower bound on the number of observational equivalence classes. For a given set R of nondominance-proving rules, the proven-inequivalence partition one can obtain from this set is referred to as the proven-inequivalence partition \emph{induced by R}.

As we will see, the comparison of the set of d-separation relations of each mDAG is a nondominance-proving rule. The blocks of the proven-inequivalence partition induced by the comparison of d-separation relations are what are conventionally referred to as the Markov equivalence classes~\cite{Ayesha_Markov}. 

If a given set of mDAGs is a block of \emph{both} the proven-equivalence partition and the proven-inequivalence partition, then it is a block of the observational equivalence partition. The leftmost block of Fig.~\ref{fig_veinn} illustrates this circumstance. 
This is the method we use to characterize what we can prove about the observational equivalence classes of 3-node and 4-node mDAGs consistent with a fixed nodal ordering.

To obtain the proven-equivalence partition from a set of dominance-proving rules, we start by constructing a partially ordered set of mDAGs with the partial order induced by such rules. This will be called the \emph{proven-dominance partial order}. The proven-equivalence partition is then determined by taking pairs of mDAGs such that each dominates the other in the proven-dominance partial order to be part of the same block.

\section{Rules for Establishing Observational Dominance}
\label{sec_show_equivalence}

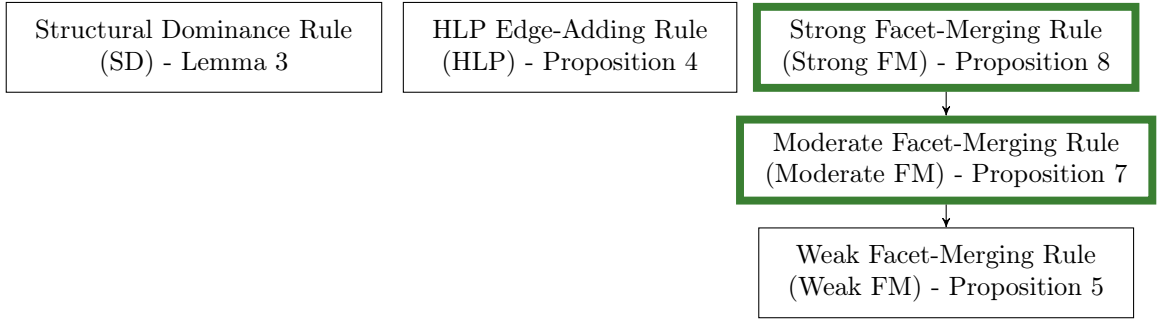
\begin{figure*}[h!]
	\centering
	\begin{tikzpicture}
		\node[c](A) at (0,2){\begin{tabular}{c} 	Structural Dominance Rule \\ (SD) - Lemma~\ref{prop_edge_dropping}\end{tabular}};
		\node[c](B) at (5,2){\begin{tabular}{c} HLP Edge-Adding  Rule  \\ (HLP) - Proposition~\ref{HLP}\end{tabular}};
		\node[draw=OliveGreen!100, line width=3pt](C) at (10,2){\begin{tabular}{c} Strong Facet-Adding Rule \\ (Strong FA) - Proposition~\ref{Simultaneous_Splitting_Prop}\end{tabular}};
		\node[draw=OliveGreen!100, line width=3pt](D) at (10,0.5){\begin{tabular}{c} Moderate Facet-Adding Rule \\ (Moderate FA) - Proposition~\ref{prop_moderateFS}\end{tabular}}edge[e] (C);
		\node[c](E) at (10,-1){\begin{tabular}{c} Weak Facet-Adding Rule \\(Weak FA) - Proposition~\ref{prop_weakFS}\end{tabular}}edge[e] (D);
	\end{tikzpicture}
	\caption{Classification of the different dominance-proving rules that we use. Classification of the different nondominance-proving rules. An arrow from one rule to another indicates that the latter is strictly less powerful than the former (i.e., the latter is subsumed in the former). The rules indicated in thick green rectangles are original results of this work.}
	\label{Dominance_Rules}
\end{figure*}

In this section, we will present the known rules that provide conditions under which one mDAG is known to observationally dominate another. 
In Fig.~\ref{Dominance_Rules} presents a diagram with all of the dominance-proving rules that we will use here. The rules that are indicated in thick green rectangles are our original theoretical results, namely, two generalizations of a rule presented in Ref.~\cite{evans_graphs_2016}.  

The first dominance-proving rule was already given in Lemma~\ref{prop_edge_dropping}: structural dominance implies observational dominance. 

\subsection{HLP Edge-Adding  Rule}

The dominance-proving rule we present in this section is Theorem 26.4 of the work of Henson, Lal and Pusey\cite{henson_theory-independent_2014}, that we are going to call the \emph{HLP Edge-Adding  Rule}:

\begin{proposition}[HLP Edge-Adding  Rule (HLP)~\cite{henson_theory-independent_2014}]
\label{HLP}
Let $\mathfrak{G}=\{\cal D, B\}$ be an mDAG, and let $x$ and $y$ be two of its nodes. Let $\mathfrak{G}'$ be the mDAG obtained from $\mathfrak{G}$ by adding a directed edge $x\rightarrow y$. 

Suppose that:
\begin{enumerate}
	\item $\pa_{\mathcal{D}}(x)\subseteq \pa_{\mathcal{D}}(y)$,
	\item Whenever $x\in B$ for a facet $B\in\mathcal{B}$, then also $y\in B$.
\end{enumerate}

In this case, $\mathfrak{G}$ observationally dominates $\mathfrak{G}'$, i.e., $\mathfrak{G}\succeq \mathfrak{G}'$.

Note that, since $\mathfrak{G}'$ structurally dominates $\mathfrak{G}$, by Lemma~\ref{prop_edge_dropping} we know  that $\mathfrak{G}'$ observationally dominates $\mathfrak{G}$, i.e., $\mathfrak{G}'\succeq \mathfrak{G}$. Therefore, $\mathfrak{G}$ and $\mathfrak{G}$ are observationally equivalent, i.e., $\mathfrak{G} \cong \mathfrak{G}'$.
\end{proposition}

\begin{figure}[h!]
\centering
\includegraphics[width=0.47\textwidth]{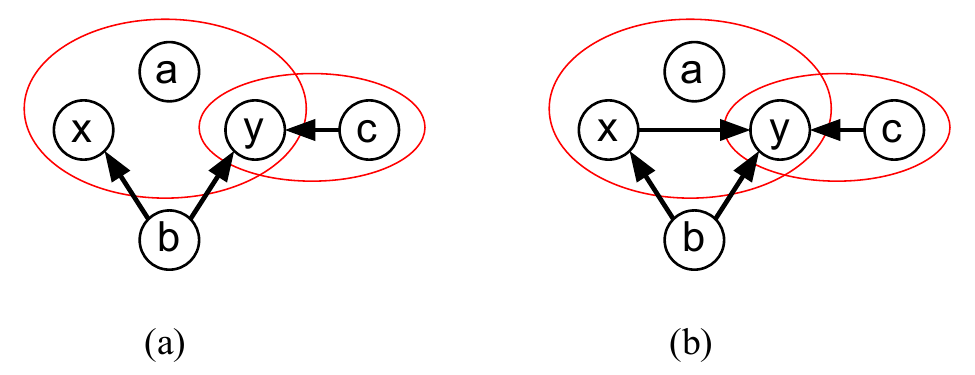}
\caption{Two mDAGs that can be proven to be observationally equivalent by the HLP Edge-Adding  Rule~\cite{henson_theory-independent_2014}.}
\label{fig_HLP}
\end{figure}

For example, from the HLP Edge-Adding  Rule it follows that the two mDAGs of Fig.~\ref{fig_HLP} are observationally equivalent. 

Note that this dominance-proving rule establishes observational \emph{equivalence}, because the conditions for its applicability already imply observational dominance in the other direction. This will be also the case for the next dominance-proving rules.

Furthermore, note that the HLP Edge-Adding rule can establish observational dominance between two mDAGs that are \emph{not} consistent with the same nodal ordering: if nodes $x$ and $y$ have the same set of parents and belong to the same facets, we can first use this rule in reverse to remove the arrow $x\to y$, and then use the rule to add the arrow $y\to x$. Because of that, it is conceivable that there exist observational dominances between mDAGs $\mathfrak{G}$ and $\mathfrak{G}'$ consistent with the same nodal ordering that can only be discovered by first showing that $\mathfrak{G}$ dominates some mDAG that is \emph{not} consistent with the same nodal ordering, and then showing that this mDAG dominates $\mathfrak{G}'$. Therefore, when constructing the observational partial order of mDAGs that follow a fixed nodal ordering, we first implement the dominance-proving rules of this section for the set of \emph{all} mDAGs, and then restrict our presentation to those that are consistent with a fixed nodal ordering.

\subsection{Facet-Adding Rules}

In this section, we will present three more dominance-proving rules. As we will explain, the first rule is obtained from a result of Ref.~\cite{evans_graphs_2016}, and the second and third rules are generalizations of the first.

\subsubsection{Weak Facet-Adding}

\begin{proposition}[Weak Facet-Adding (Weak FA) - Consequence of Ref.~\protect{\cite[Proposition 6.1]{evans_graphs_2016}}]
\label{prop_weakFS}
Let $\mathfrak{G}=\{\cal D, B\}$ be an mDAG whose simplicial complex $\cal B$ contains two disjoint faces  $C$ and $D$. If  $B=C \cup D$ is not already present in the simplicial complex of $\mathfrak{G}$,  let $\mathfrak{G}'$ be the mDAG obtained by starting from $\mathfrak{G}$ and adding  $B$  and all of the faces contained in $B$ to its simplicial complex. 

Suppose that:
\begin{enumerate}
	\item $\pa_{\mathcal{D}}(C)\cup C\subseteq \pa_{\mathcal{D}}(d)$ for each $d \in D$,
	\item 
	 Every face that contains any $c\in C$ is a subset of $C$. 
\end{enumerate}

In this case, $\mathfrak{G}$ observationally dominates $\mathfrak{G}'$, i.e., $\mathfrak{G}\succeq \mathfrak{G}'$. 

Note that, since $\mathfrak{G}'$ structurally dominates $\mathfrak{G}$, by Lemma~\ref{prop_edge_dropping} we know  that $\mathfrak{G}'$ observationally dominates $\mathfrak{G}$, i.e., $\mathfrak{G}'\succeq \mathfrak{G}$. Therefore, $\mathfrak{G}$ and $\mathfrak{G}'$ are observationally equivalent, i.e., $\mathfrak{G} \cong \mathfrak{G}'$.
\end{proposition}

Note that condition 2 implies that $C$ is not only a face, but also a facet of the simplicial complex of $\mathfrak{G}$. The face $D$, on the other hand, does not need to be a facet. 
 We call this rule ``weak'' because it will be generalized in the next two sections.

\begin{figure}[htbp]
	\centering
	\includegraphics[width=0.48\textwidth]{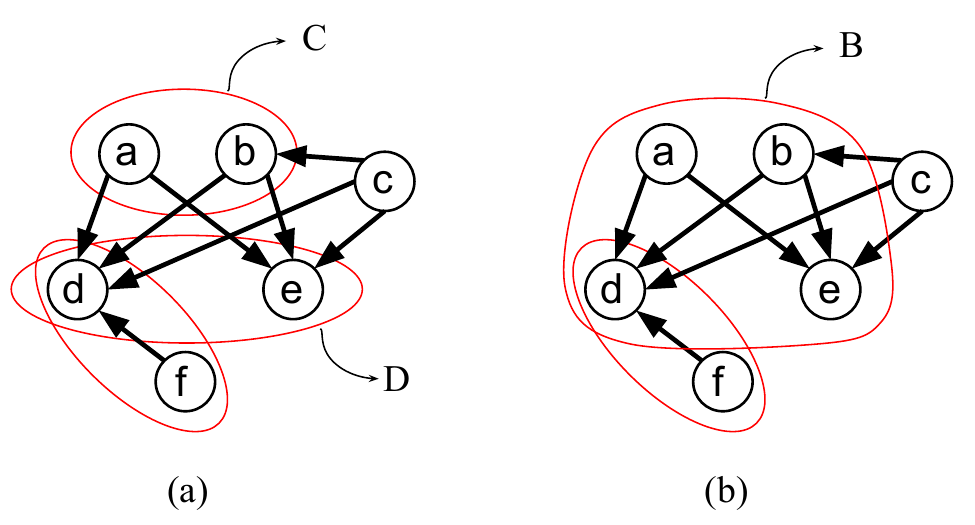}
	\caption{Two mDAGs that can be proven observationally equivalent by the Weak Facet-Adding rule. Here, the sets $B$, $C$ and $D$ from the statement of Proposition~\ref{prop_weakFS} are respectively $B=\{a,b,d,e\}$, $C=\{a,b\}$ and $D=\{d,e\}$.}
	\label{fig_WeakFM}
\end{figure}

 As an example, the Weak Facet-Adding Rule implies that the two mDAGs of Fig.~\ref{fig_WeakFM} are observationally equivalent. An intuition for the validity of this rule is discussed in Section~\ref{section_3nodes_observationalequivalences}.

The Weak Facet-Adding Rule is a special case of Proposition 6.1 of Ref.~\cite{evans_graphs_2016}, that we are going to call ``Evans' Rule''. We paraphrase it below.

\begin{proposition}[Evans' Rule~\protect{\cite[Proposition 6.1]{evans_graphs_2016}}]
	\label{Evans}
	Let $\mathfrak{G}=\{\cal D, B\}$ be an mDAG whose simplicial complex $\cal B$  contains two disjoint  faces  $C$ and $D$. Further suppose that:
	\begin{enumerate}
		\item $\pa_{\mathcal{D}}(C)\cup C \subseteq \pa_{\mathcal{D}}(d)$ for each $d \in D$.
		\item 	 Every face that contains any $c\in C$ is a subset of $C$. 
	\end{enumerate}

	  If  $B=C \cup D$ is not already present in the simplicial complex of $\mathfrak{G}$,  let $\mathfrak{G}'$ be the mDAG obtained by  starting from $\mathfrak{G}$, adding  $B$  to its simplicial complex and deleting all of the directed edges $c\rightarrow d$ for every $c\in C$, $d\in D$. In this case,   $\mathfrak{G}$ is observationally equivalent to $\mathfrak{G}'$, i.e.,  $\mathfrak{G} \cong \mathfrak{G}'$.
\end{proposition}

\begin{figure*}[htbp]
	\centering
	\includegraphics[width=0.9\textwidth]{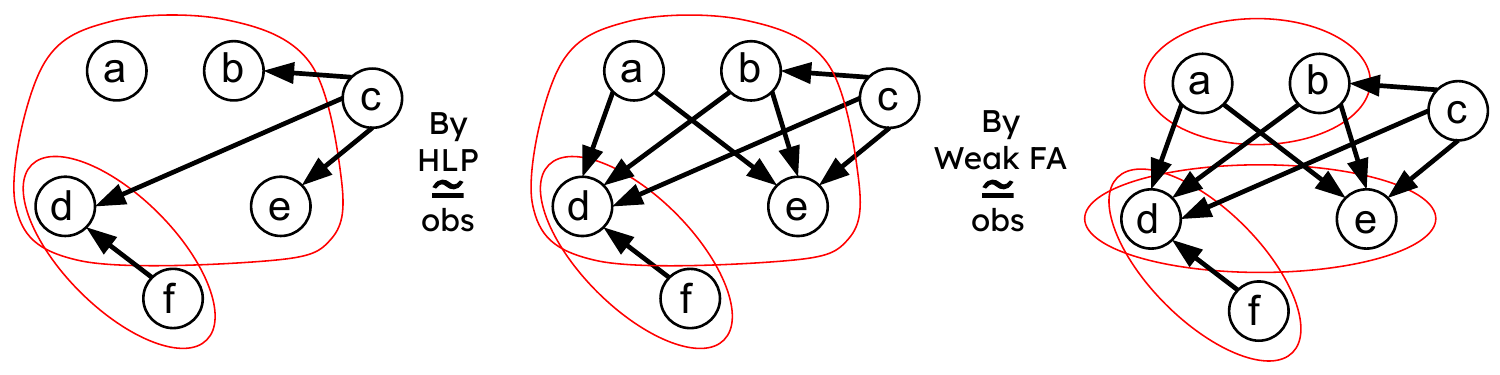}
	\caption{The first and third mDAG of this figure can be shown observationally equivalent  by Evans' rule, as well as by the HLP Edge-Adding rule together with the Weak Facet-Adding rule. This exemplifies Lemma~\ref{lemma_WeakFS_and_HLP_Evans}.}
	\label{fig_HLPandFS}
\end{figure*}

It is easy to see that the Weak Facet-Adding Rule (Proposition~\ref{prop_weakFS}) is a special case of Evans' Rule (Proposition~\ref{Evans}): in  Weak Facet-Adding we only  add a facet,  we do not delete the directed edges $c\rightarrow d$. However, as we will see below, the HLP Edge-Adding  Rule can be applied together with Weak Facet-Adding to obtain Evans' Rule. An example of this is given in Fig.~\ref{fig_HLPandFS}.

\begin{lemma}[Weak FA $\land$ HLP$\Rightarrow$Evans]
\label{lemma_WeakFS_and_HLP_Evans}
The application of Weak Facet-Adding (Proposition  \ref{prop_weakFS}) together with the HLP Edge-Adding  Rule (Proposition~\ref{HLP}) subsumes Evans's Rule (Proposition~\ref{Evans}).
\end{lemma}
\begin{proof}
	
Let  $\mathfrak{G}=\{\cal D, B\}$ be an mDAG that satisfies the conditions of Propositions~\ref{prop_weakFS} and~\ref{Evans}. By Weak Facet-Adding (Proposition~\ref{prop_weakFS}), we know that  $\mathfrak{G}$ is observationally equivalent to an mDAG $\mathfrak{G}'$ where the face $B=C\cup D$ is added.  This mDAG $\mathfrak{G}'$ is such that, for each $c\in C$ and $d\in D$:
\begin{enumerate}
	\item $\pa_{\mathcal{D}}(c)\subseteq \pa_{\mathcal{D}}(d)$.
	\item The only facet that includes $c$ is $B$. This facet also includes $d$.
\end{enumerate}

That is, the conditions of the HLP Edge-Adding  Rule are satisfied. This rule implies that the mDAG $\mathfrak{G}''$ obtained by starting with $\mathfrak{G}'$ and removing all of the arrows $c\rightarrow d$ for $c\in C$ and $d\in D$ is observationally equivalent to $\mathfrak{G}'$, i.e., $\mathfrak{G}'' \cong \mathfrak{G}'$. 

This mDAG $\mathfrak{G}''$ is the one that is obtained directly from $\mathfrak{G}$ by application of Evans's Rule. Therefore, the joint application of the Weak Facet-Adding Rule and the HLP Edge-Adding  Rule can subsume Evans's Rule.
\end{proof}

Hence, we can separate the part of Evans's Rule that adds a facet,  that is included in the Weak Facet-Adding Rule, from the part that removes the directed edges $c\rightarrow d$, that is taken care of by the HLP Edge-Adding  Rule. Here, we will use HLP and Weak FA in proving dominances. Lemma~\ref{lemma_WeakFS_and_HLP_Evans} implies  that doing so, we obtain any conclusions that could have been obtained by applying Evans' Rule.

\subsubsection{Moderate Facet-Adding}
\label{sec_face_splitting}

\begin{figure}[h!]
	\centering
	\includegraphics[width=0.48\textwidth]{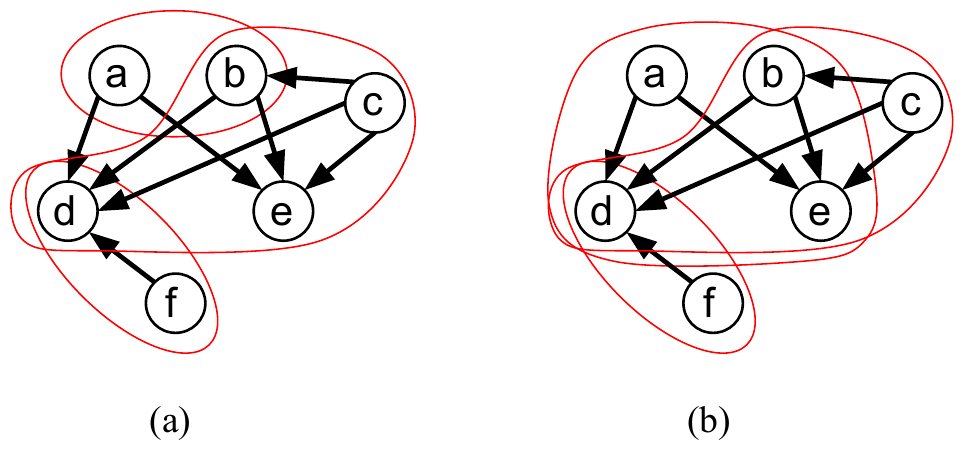}
	\caption{Two mDAGs that can be proven observationally equivalent by Moderate Facet-Adding, but not by Weak Facet-Adding. Here, the sets $B$, $C$ and $D$ from the statement of Proposition~\ref{prop_weakFS} are respectively $B=\{a,b,d,e\}$, $C=\{a,b\}$ and $D=\{d,e\}$.} 
	\label{fig_splitting} 
\end{figure}

Our first supplement to existing dominance-proving rules is an extension of Weak Facet-Adding, wherein condition 2 is relaxed. We will call it \emph{Moderate Facet-Adding} rule. Fig.~\ref{fig_splitting} presents an observational equivalence that can be shown by Moderate Facet-Adding but not by Weak Facet-Adding.

In the next subsection, the Moderate Facet-Adding rule will be further generalized, when we present the \emph{Strong Facet-Adding} (Strong FA) rule. Even though the Moderate Facet-Adding rule is an intermediate result between Weak FA and Strong FA, we opted to present it because, as can be seen from its proof, it is a very simple extension of the Weak FA rule. As we will see, the statement and proof of the Strong FA rule are less intuitive.

\begin{proposition}[Moderate Facet-Adding (Moderate FA)]
\label{prop_moderateFS}
Let $\mathfrak{G}=\{\cal D, B\}$ be an mDAG whose simplicial complex $\cal B$ contains two disjoint faces  $C$ and $D$. If  $B=C \cup D$ is not already present in the simplicial complex of $\mathfrak{G}$,  let $\mathfrak{G}'$ be the mDAG obtained by starting from $\mathfrak{G}$ and adding  $B$  and all of the faces contained in $B$ to its simplicial complex. 

Suppose that:
\begin{enumerate}
	\item $\pa_{\mathcal{D}}(C)\cup C\subseteq \pa_{\mathcal{D}}(d)$ for each $d \in D$,
	\item If for a  facet  $B'\in \cal B$ such that $B' \not \subseteq C$ we have $c\in B'$ for some $c\in C$, then $D\subset B'$.
\end{enumerate}

In this case, $\mathfrak{G}$  observationally dominates $\mathfrak{G}'$, i.e., $\mathfrak{G}\succeq \mathfrak{G}'$.

Note that, since $\mathfrak{G}'$ structurally dominates $\mathfrak{G}$, by Lemma~\ref{prop_edge_dropping} we know  that $\mathfrak{G}'$ observationally dominates $\mathfrak{G}$, i.e., $\mathfrak{G}'\succeq \mathfrak{G}$. Therefore, $\mathfrak{G}$ and $\mathfrak{G}'$ are observationally equivalent, i.e., $\mathfrak{G} \cong \mathfrak{G}'$.
\end{proposition}
\begin{proof}

	Construct the mDAG $\tilde{\mathfrak{G}}$ by starting from $\mathfrak{G}$ and i) for each \emph{facet} $\tilde{B}\in\mathcal{B}$ such that $\tilde {B}\not\subseteq C$  and $c\in \tilde{B}$ for some $c\in C$, adding a  parentless visible node $\tilde{b}$ that is a parent of every node in $\tilde{B}$, and ii) removing every \emph{face} $B'\in\mathcal{B}$ such that $B'\not\subseteq C$  and $c\in B'$ for some $c\in C$. For example, starting from the mDAG of Fig.~\ref{fig_splitting}(a), where $C=\{a,b\}$, the two steps above result in the mDAG of Fig.~\ref{fig_mdoerate_proof}, where $g$ is the visible node added in step i).
	
\begin{figure}[h!]
	\centering
	\includegraphics[width=0.25\textwidth]{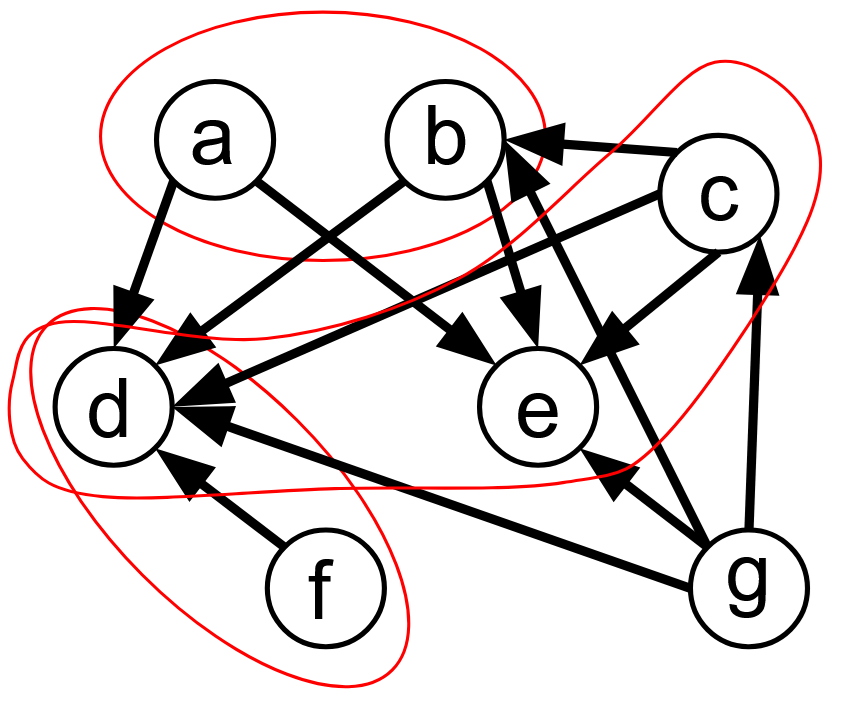}
	\caption{mDAG obtained by starting from Fig.~\ref{fig_splitting}(a) and performing the steps i) and ii) described in the proof of Proposition~\ref{prop_moderateFS}.} 
	\label{fig_mdoerate_proof} 
\end{figure}

	The mDAG $\tilde{\mathfrak{G}}$ satisfies the assumptions of Weak Facet-Adding. Therefore, we know that $\tilde{\mathfrak{G}}$ is observationally equivalent to $\tilde{\mathfrak{G'}}$, the mDAG obtained by adding the face $B=C\cup D$ to it. 
	
	If two pDAGs $\mathcal{G}_1$ and $\mathcal{G}_2$ are observationally equivalent, then the two pDAGs $\mathcal{G}_3$ and $\mathcal{G}_4$ obtained by starting from $\mathcal{G}_1$ and $\mathcal{G}_2$ respectively and transforming the same set of visible nodes into latent nodes also should be observationally equivalent. With this idea, we can show the observational equivalence between $\mathfrak{G}$ and   $\mathfrak{G'}$: by starting from $\can(\tilde{\mathfrak{G}})$ and  transforming the visible nodes added in step i) into latent nodes, we obtain a pDAG whose corresponding mDAG is $\mathfrak{G}$. By starting from  $\can(\tilde{\mathfrak{G'}})$ and transforming the same visible nodes into latent, we obtain a pDAG whose corresponding mDAG is $\mathfrak{G'}$.

\end{proof}

Note the difference in condition 2 for this proposition and for Propositions  \ref{prop_weakFS} and \ref{Evans}: here, nodes of the set $C$  are allowed to belong to faces that are not included in $C$,  as long as these faces   also include the entire set $D$.

\subsubsection{Strong Facet-Adding}
\label{sec_simultaneous_splitting}

\begin{figure}[h!]
\centering
\includegraphics[width=0.48\textwidth]{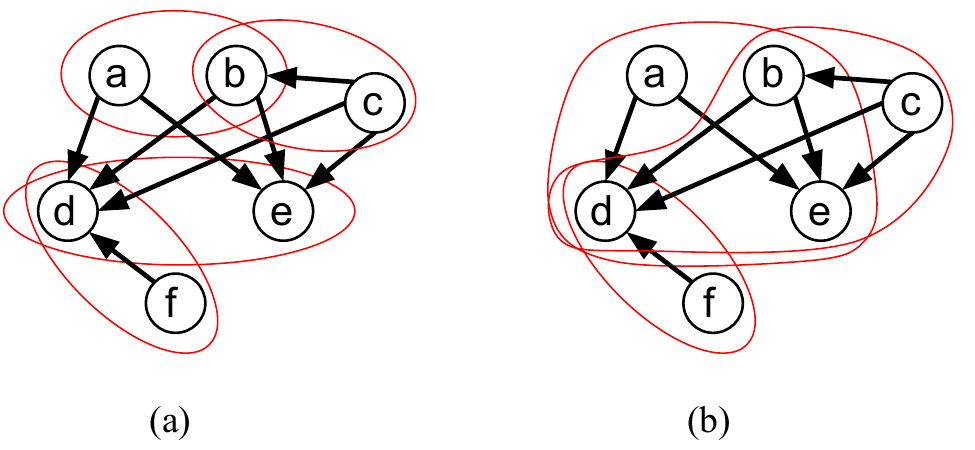}
\caption{Two mDAGs that can be shown observationally equivalent by the Strong Facet-Adding rule.} 
\label{fig_double_splitting} 
\end{figure}

In Fig.~\ref{fig_double_splitting}, we have two mDAGs that cannot be shown equivalent by the combination of the HLP Edge-Adding  Rule and the Moderate Facet-Adding Rule.

As it turns out, however, these mDAGs \emph{are} observationally equivalent. This can be shown by the following proposition, which is our second supplement to existing dominance-proving rules:

\begin{restatable}[Strong Facet-Adding (Strong FA)]{proposition}{StrongFS}
\label{Simultaneous_Splitting_Prop}
Let $\mathfrak{G}=\{\cal D, B\}$ be an mDAG whose simplicial complex contains a sequence of  faces   $C_i\in \cal B$ for $i=1,...,n$, that do not need to be disjoint. Suppose further that the mDAG $\mathfrak{G}$ also contains a   face   $D\in \cal B$ that is disjoint to every $C_i$,  $i=1,...,n$. Let $\mathfrak{G}'$ be the mDAG obtained by starting from $\mathfrak{G}$ and, if not already present,  adding the sets  $B_i=C_i \cup D$,  $i=1,...,n$, together with all of the faces they contain, to its simplicial complex.

Suppose that:
\begin{enumerate}
	\item $\pa_{\mathcal{D}}(C_i)\cup C_i\subseteq \pa_{\mathcal{D}}(d)$ for each $d \in D$, $i=1,...,n$.
	\item If for a   facet  $B'\in \cal B$  such that  $B' \not \subseteq C_i$ for all $i=1,...,n$  we have $c\in B'$ for some $c\in \cup_{i=1,...,n} C_i$, then $D\subseteq B'$.
\end{enumerate}
In this case, $\mathfrak{G}$  observationally dominates $\mathfrak{G}'$, i.e., $\mathfrak{G}\succeq \mathfrak{G}'$.

Note that, since $\mathfrak{G}'$ structurally dominates $\mathfrak{G}$, by Lemma~\ref{prop_edge_dropping} we know  that $\mathfrak{G}'$ observationally dominates $\mathfrak{G}$, i.e., $\mathfrak{G}'\succeq \mathfrak{G}$. Therefore, $\mathfrak{G}$ and $\mathfrak{G}'$ are observationally equivalent, i.e., $\mathfrak{G} \cong \mathfrak{G}'$.
\end{restatable}

The proof of this proposition is presented in Appendix \ref{appendix_facesplitting}. The general intuition is given in Section~\ref{section_3nodes_observationalequivalences}, in the discussion of the observational equivalence between the second and third mDAGs of Fig.~\ref{fig_instrumental_class}.

This result shows conditions under which multiple facets can be merged simultaneously. In the example of Fig.~\ref{fig_double_splitting}, we have $D=\{1,2\}$, $C_1=\{3,4\}$, $C_2=\{4,5\}$, $B_1=\{1,2,3,4\}$ and $B_2=\{1,2,4,5\}$.

The Structural Dominance Rule (Lemma~\ref{prop_edge_dropping}), the HLP Edge-Adding  Rule and the Strong Facet-Adding Rule are together the strongest set of dominance-proving rules currently known.

As we saw, when combined with Lemma~\ref{prop_edge_dropping}, all of the dominance-proving rules that we presented in this section establish {observational}  \emph{equivalence}.  That is, every time that one of these rules can be used to show that an mDAG $\mathfrak{G}$ observationally dominates another mDAG $\mathfrak{G}'$, the conditions for applicability of the rule imply that  $\mathfrak{G}'$ structurally dominates $\mathfrak{G}$.

Therefore, the only known dominance-proving rule that can potentially establish a strict observational dominance (i.e., observational dominance without observational equivalence) is Lemma~\ref{prop_edge_dropping}. This leads us to  the following conjecture, that says that Lemma~\ref{prop_edge_dropping} is indeed the \emph{only} way through which strict observational dominance can be established: 

\begin{conjecture}
	\label{conjecture_structural}
If an mDAG $\mathfrak{G}$ strictly observationally dominates an mDAG $\mathfrak{G}'$, i.e., $\mathfrak{G}\succ \mathfrak{G}'$, then there exists a sequence of mDAGs $\mathfrak{G}=\mathfrak{G}_1, \mathfrak{G}_2,...,\mathfrak{G}_n=\mathfrak{G}'$ such that each pair $\mathfrak{G}_i$, $\mathfrak{G}_{i+1}$ is such that either:
\begin{enumerate}[label=(\roman*)]
	\item $\mathfrak{G}_i$ is observationally equivalent to $\mathfrak{G}_{i+1}$, or
	\item  $\mathfrak{G}_i$ \emph{structurally} dominates $\mathfrak{G}_{i+1}$.
\end{enumerate}
\end{conjecture}

 While this is true for the case of 3-node mDAGs, it is still not known whether this is true in general.

\begin{figure*}[h!]
\centering
\includegraphics[width=0.9\textwidth]{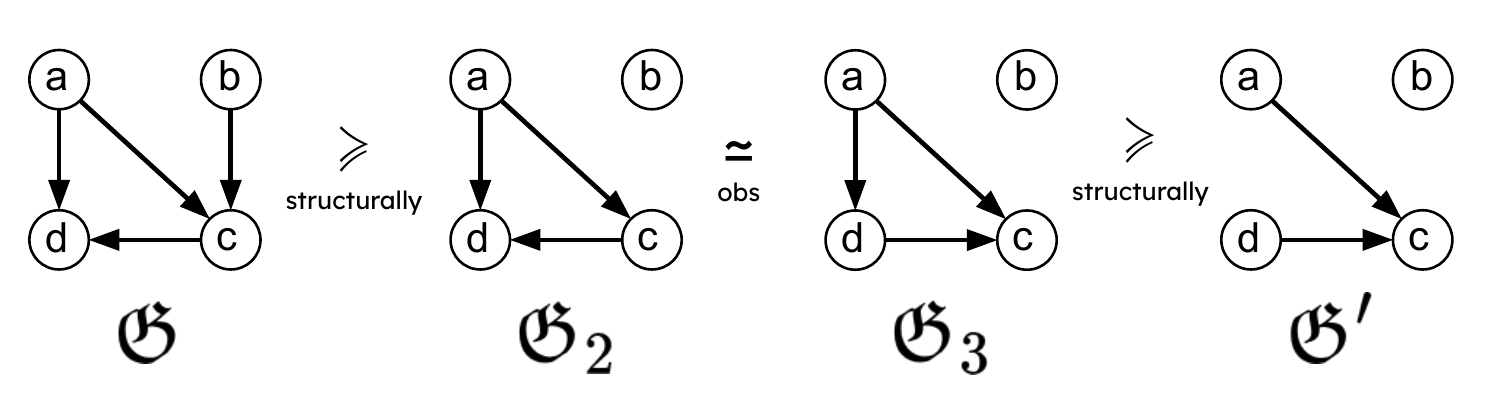}
\caption{Example of the sequence of mDAGs that appears in Conjecture~\ref{conjecture_structural}. One can see that $\mathfrak{G}_2$ and $\mathfrak{G}_3$ are observationally equivalent by applying the HLP Edge-Adding rule twice: first to remove the arrow $c\to d$ and second to add the arrow $d\to c$.  Conjecture~\ref{conjecture_structural} says that every strict observational dominance can be obtained by a sequence like this one.}
\label{fig_referee_example} 
\end{figure*}

To illustrate the need for a sequence in this conjecture, Fig.~\ref{fig_referee_example} shows an example of latent-confounder-free mDAGs $\mathfrak{G}$ and $\mathfrak{G}'$ (that are \emph{not} consistent with the same nodal ordering). Using the dominance-proving rules described in this section together with the nondominance-proving rules described in the next section, we can check that \emph{none} of the mDAGs that are observationally equivalent to $\mathfrak{G}$ structurally dominate any of the mDAGs that are observationally equivalent to $\mathfrak{G}'$. Using the sequence $\mathfrak{G}=\mathfrak{G}_1, \mathfrak{G}_2, \mathfrak{G}_3, \mathfrak{G}_4=\mathfrak{G}'$, however, we can see that observational equivalence together with {structural} dominance is sufficient to show the observational dominance of $\mathfrak{G}$ over $\mathfrak{G}'$.   In fact, the version of Conjecture~\ref{conjecture_structural} where both $\mathfrak{G}$ and $\mathfrak{G}'$ are latent-confounder-free is known to be true, as a consequence of Ref.~\cite{Chickering_greedy}.  

\section{Rules for Establishing Observational Nondominance}
\label{sec_show_inequivalence}

In this section, we are going to present all the known rules for establishing that one mDAG does \emph{not} observationally dominate another. In particular, observational nondominance implies observational inequivalence, so the successive application of a set of nondominance-proving rules will give us a proven-inequivalence partition induced by that set of rules (gray loops of Fig.~\ref{fig_veinn}).

\begin{figure*}[h!]
\centering
\begin{tikzpicture}
	\node[c](C) at (0,2){Comparison of Unrealizable Supports (Supps) - Section \ref{sec_unrealizable_supports}};
	\node[c](E) at (4.6,-1){\begin{tabular}{c} Directed-edge-free rule  \\ {(DEF) - Section \ref{sec_only_hypergraphs}}\end{tabular}};
	\draw[e] (E) to node[right,xshift=4pt,yshift=0pt]{Lemma \ref{lemma_sup_OH}} (C);
	\node[c](F) at (0,-1){\begin{tabular}{c} Comparison of \\ densely connected \\ pairs of nodes  \\ (DC) - Section \ref{sec_dense_connected} \\ \end{tabular}};
	\draw[e] (F) to node[right,xshift=-2pt,yshift=0pt]{Lemma \ref{lemma_sup_DC}} (C);
	\node[c](B) at (-4.2,-1){\begin{tabular}{c} Comparison of \\ e-separation relations  \\ (e-sep) - Section \ref{sec_esep} \end{tabular}};
	\draw[e] (B) to node[left,xshift=-5pt,yshift=0pt]{Lemma \ref{lemma_sup_esep}} (C);
	\node[c](A) at (-6.4,-4){\begin{tabular}{c} Comparison of \\skeletons  \\(skel) - Section \ref{sec_skeletons} \end{tabular}};
	\draw[e] (A) to node[left,xshift=-4pt,yshift=0pt]{Lemma \ref{lemma_esep_skeleton}} (B);
	\node[c](D) at (-2.2,-4){\begin{tabular}{c} Comparison of \\ d-separation relations \\(d-sep)  - Section \ref{section_dsep}  \end{tabular}};
	\draw[e] (D) to node[right,xshift=4pt,yshift=0pt]{Lemma \ref{lemma_esep_dsep}} (B);
\end{tikzpicture}
\caption{Classification of the different nondominance-proving rules. An arrow from one rule to another indicates that the latter is strictly less powerful than the former (i.e., the latter is subsumed in the former), and the Lemma that proves it is indicated on the arrow. We defined a shorthand name for each rule, shown inside the parentheses.
}
\label{Inequivalence_Rules}
\end{figure*}
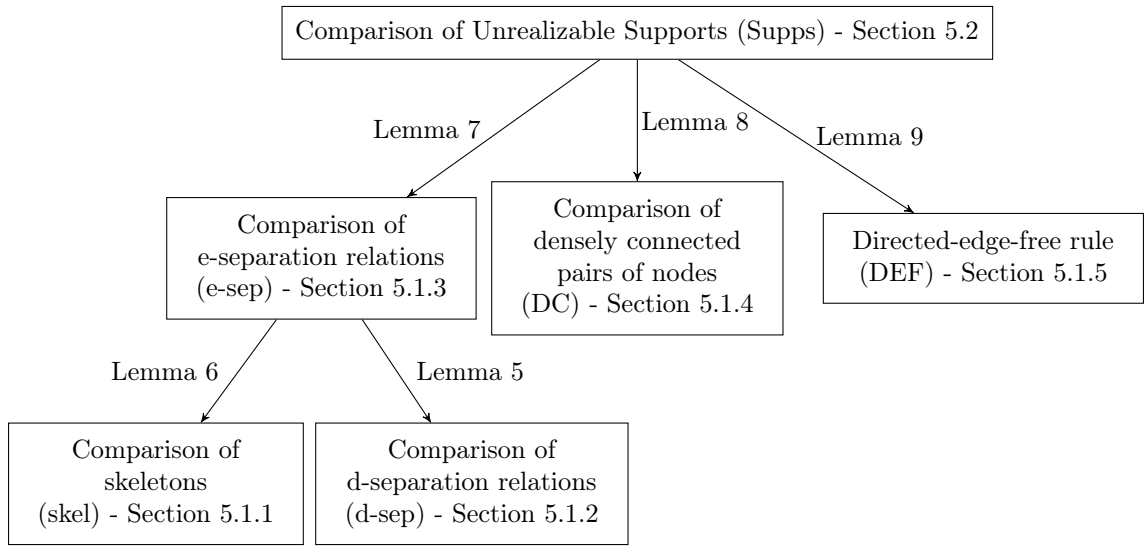

In Fig.~\ref{Inequivalence_Rules}, we present a diagram with all the nondominance-proving rules that will be presented in this section. One rule may be subsumed by another if all of the nondominances that can be shown by the first can also be shown by the second. We represent this in Fig.~\ref{Inequivalence_Rules} as arrows from the stronger to the weaker rule. However, it is still of value to discuss the weaker rules, as they are in general computationally cheaper to apply.

We believe that the comparison of the sets of nested Markov constraints~\cite{behaviormetrika} presented by each mDAG constitutes a rule for establishing observational nondominance. However, the literature still does not present an explicit construction of a distribution that is realizable by one mDAG but not by the other when their sets of nested Markov constraints differ.\footnote{According to private communication with Robin Evans, however, such a distribution exists and is to appear in future work.} In any case, we believe that it is unlikely that this comparison would show observational nondominances that our other rules do not already show. Therefore, we have opted to not include this rule here.

\subsection{Graphical Rules}
\label{sec_graphical}

We will organize our discussion of the nondominance-proving rules starting from the computationally cheaper ones (i.e., we proceed bottom-up in the diagram of Fig.~\ref{Inequivalence_Rules}). 

\subsubsection{Skeletons}
\label{sec_skeletons}

\begin{figure}[h!]
\centering
\includegraphics[width=0.47\textwidth]{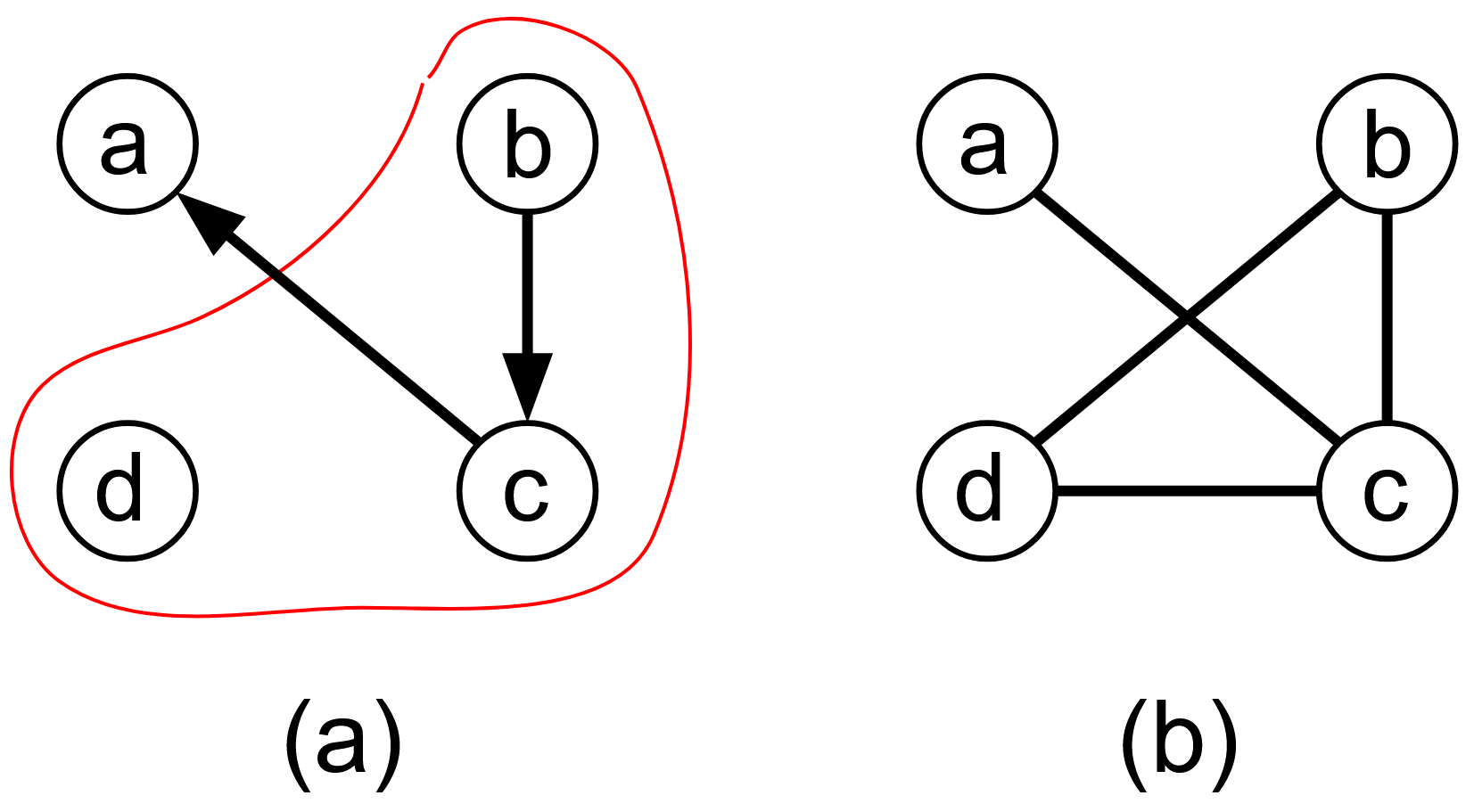}
\caption{(a) An mDAG; and (b) Its skeleton.} 
\label{fig_skeleton} 
\end{figure}

The first nondominance rule is the comparison of skeletons of mDAGs, that was introduced in Refs.~\cite{Evans2012GraphicalMF, evans_graphs_2016}. For that, we first define the \emph{skeleton} of an mDAG, which is exemplified in Fig.~\ref{fig_skeleton}.

\begin{definition}[Skeleton]
Let $\mathfrak{G}=(\cal D,B)$ be an mDAG. We define the \emph{skeleton} of $\mathfrak{G}$ by the undirected graph with the same nodes as $\cal D$ and with an edge between nodes $u$ and $w$ whenever there is a directed edge between them in $\cal D$ or when $u,w\in B$ for some $B\in \mathcal{B}$.
\end{definition}

Below, we present a corollary of Theorem 4.2 of Ref.~\cite{Evans2012GraphicalMF}, which is a generalization of Proposition 6.5 of Ref.~\cite{evans_graphs_2016}:

\begin{proposition}[Comparison of skeletons  - Consequence of Refs.~\cite{Evans2012GraphicalMF, evans_graphs_2016}]Let $\mathfrak{G}$ and $\mathfrak{G}'$  be two mDAGs such that $\nodes(\mathfrak{G})=\nodes(\mathfrak{G}')$.  
If there exist nodes $x,y\in \nodes(\mathfrak{G})$ such that the undirected edge between $x$ and $y$ is present in the skeleton of $\mathfrak{G}'$ but not in the skeleton of $\mathfrak{G}$, then $\mathfrak{G}$ \emph{does not} observationally dominate $\mathfrak{G}'$, i.e., $\mathfrak{G}\not\succeq \mathfrak{G}'$.
\end{proposition}

As a special case of this rule, we know that two mDAGs are observationally inequivalent whenever their skeletons are different.

\subsubsection{d-separation relations}
\label{section_dsep}

As discussed in Section \ref{sec_causal_compatibility}, all the conditional independence constraints that an mDAG imposes on its compatible probability distributions can be obtained from the d-separation relations of that mDAG. We can compare the sets of d-separation relations among the visible variables as a way to show observational nondominance between two causal structures:

\begin{proposition}[Comparison of d-separation relations  - Consequence of Theorem \ref{th:d-sep}]
Let $\mathfrak{G}$ and $\mathfrak{G}'$ be two mDAGs such that $\nodes(\mathfrak{G})=\nodes(\mathfrak{G}')$. If there is a d-separation relation that is presented by  $\mathfrak{G}$ but not by $\mathfrak{G}'$, then $\mathfrak{G}$ \emph{does not} observationally dominate $\mathfrak{G}'$, i.e., $\mathfrak{G}\not\succeq \mathfrak{G}'$.
\end{proposition}
\begin{proof}
We will make a proof by contradiction. Assume that $\mathfrak{G}$ observationally dominates $\mathfrak{G}'$, but there is a d-separation relation $x\dsep y|z$ for $x,y,z \in \nodes(\mathfrak{G})$ that is presented by $\mathfrak{G}$ but not by $\mathfrak{G}'$. 

By the first point of Theorem \ref{th:d-sep}, if a probability distribution $P$ is realizable by $\mathfrak{G}$, then $P$ is such that $X_x \dbot X_y |X_z$ in $P$. Because of the assumed observational dominance, \emph{every} distribution realizable by $\mathfrak{G}'$ is also realizable by $\mathfrak{G}$, and thus has to satisfy $X_x \dbot X_y |X_z$. By the second point of Theorem \ref{th:d-sep}, this would imply that $x\dsep y|z$ in $\mathfrak{G}'$, which is a contradiction.
\end{proof} 

Therefore, we can use the comparison of d-separation relations to classify the mDAGs into another proven-inequivalence partition.

\begin{figure}[!h]
\captionsetup[subfigure]{aboveskip=-2pt,belowskip=-1pt}
\centering
\begin{subfigure}[b]{0.2\textwidth}
	\centering
	\includegraphics[width=\textwidth]{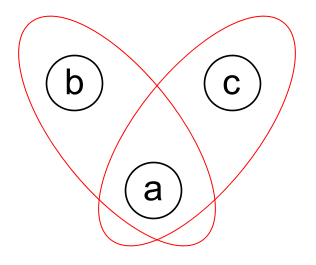}
	\caption{}
	\label{fig_Collider_skeleton}
\end{subfigure}
\hspace{1mm}
\begin{subfigure}[b]{0.2\textwidth}
	\centering
	\includegraphics[width=\textwidth]{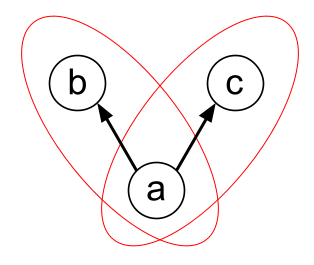}
	\caption{}
	\label{fig_Evans_skeleton1}
\end{subfigure}
\caption{Example of mDAGs that have the same skeleton, but different sets of d-separation relations. The mDAG (a) presents the d-separation relation $b\dsep c$, while the mDAG (b) does not. In Fig.~\ref{fig_temporally_ordered_3_nodes}, the mDAG (a) is in the {\tt Collider A} observational equivalence class, while (b) is in the {\tt Evans} observational equivalence class.}
\label{fig_same_skeleton_different_dsep}
\end{figure}

\begin{figure}[!h]
\captionsetup[subfigure]{aboveskip=-2pt,belowskip=-1pt}
\centering
\begin{subfigure}[b]{0.2\textwidth}
	\centering
	\includegraphics[width=\textwidth]{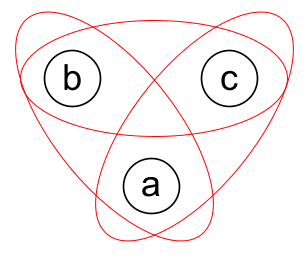}
	\caption{}
	\label{fig_Triangle_skeleton}
\end{subfigure}
\hspace{1mm}
\begin{subfigure}[b]{0.2\textwidth}
	\centering
	\includegraphics[width=\textwidth]{images/Evans_skeleton.png}
	\caption{}
	\label{fig_Evans_skeleton2}
\end{subfigure}
\caption{Example of mDAGs that have the same set of d-separation relations (i.e., no d-separation relations), but different skeletons. In Fig.~\ref{fig_temporally_ordered_3_nodes}, the mDAG (a) is in the {\tt Triangle} observational equivalence class, while (b) is in the {\tt Evans} observational equivalence class.}
\label{fig_same_dsep_different_skeleton}
\end{figure}

As it turns out, the comparison of skeletons rule and the comparison of d-separation relations rule are \emph{not} redundant to each other as rules to show observational nondominance. This can already be seen by looking at 3-node mDAGs; Figs.~\ref{fig_same_skeleton_different_dsep} and~\ref{fig_same_dsep_different_skeleton} show explicit examples.

\subsubsection{e-separation relations}
\label{sec_esep}

In Ref.~\cite{Evans2012GraphicalMF}, an extension of d-separation was introduced, called \emph{e-separation}.While the d-separation method yields all of the conditional independence constraints that a causal structure implies, the e-separation method gives \emph{some} of the inequality constraints that it implies. These are made explicit in Ref.~\cite{Finkelstein2021EntropicIC} (note that \emph{not all} inequality constraints can be obtained from e-separation).

\begin{definition}[e-separation]
\label{def_esep}
Let $\mathfrak{G}$ be an mDAG, and let $A,B,C,D\subseteq \nodes(\mathfrak{G})$ be four sets of nodes of $\mathfrak{G}$. We say that $A$ and $B$ are e-separated by $C$ after deletion of $D$, denoted $A\esep B|C\neg D$, if $A\dsep B|C$ in $\mathfrak{G}_{\nodes(\mathfrak{G})\setminus D}$, the subgraph of $\mathfrak{G}$ obtained by deleting the nodes in the set $D$.\footnote{A reader familiar with the concept of \emph{dormant independence}~\cite{dormant_independence} might wonder what is the difference between these two concepts. While a dormant independence is a type of Nested Markov constraint, which are \emph{equality} constraints, the presence of an e-separation relation leads to the existence of an \emph{inequality} constraint. Graphically, the difference is that dormant independencies need to be such that $P(X_A|X_BX_C \text{do}(X_D))$  is identifiable, while for e-separation we can check any combination of nodes being independent given the deletion of other nodes.   } 
\end{definition}

Clearly, d-separation is a special case of e-separation, where the set of nodes being deleted is the empty set, i.e., $D=\emptyset$.  As a consequence of Ref.~\cite{Finkelstein2021EntropicIC}, the comparison of e-separation relations is a nondominance-proving rule: 
\begin{proposition}[Comparison of e-separation relations  - Consequence of Ref.~\cite{Finkelstein2021EntropicIC}]
	Let $\mathfrak{G}$ and $\mathfrak{G}'$ be two mDAGs such that $\nodes(\mathfrak{G})=\nodes(\mathfrak{G}')$. If there is an e-separation relation that is presented by  $\mathfrak{G}$ but not by $\mathfrak{G}'$, then $\mathfrak{G}$ \emph{does not} observationally dominate $\mathfrak{G}'$, i.e., $\mathfrak{G}\not\succeq \mathfrak{G}'$.
\end{proposition}
\begin{proof}
	As mentioned, Ref.~\cite{Finkelstein2021EntropicIC} shows that the e-separation relation yields an inequality constraint implied by $\mathfrak{G}$. In Proposition 12 in the Appendix E of Ref.~\cite{Finkelstein2021EntropicIC}, it is shown that if $\mathfrak{G}'$ \emph{does not} present the e-separation relation in question, then \emph{there exists} some probability distribution realizable by $\mathfrak{G}'$ that violates the associated inequality (in the case where the nodes in $D$ are each associated to a discrete state space). Therefore, $\mathfrak{G}$ does not observationally dominate $\mathfrak{G}'$.
\end{proof}

The comparison of e-separation relations shows itself to be a better nondominance-proving rule than either comparison of skeletons or comparison of d-separation relations. This is made explicit by Lemmas \ref{lemma_esep_dsep} and \ref{lemma_esep_skeleton}.

\begin{lemma}[The comparison of e-separation relations subsumes The comparison of d-separation relations]
\label{lemma_esep_dsep}
Let $\mathfrak{G}$ and $\mathfrak{G}'$ be two mDAGs such that  $\mathfrak{G}$ can be shown to \emph{not} observationally dominate  $\mathfrak{G}'$ by comparison of d-separation relations. Then, $\mathfrak{G}$ can also be shown to not observationally dominate  $\mathfrak{G}'$ by comparison of e-separation relations.
\end{lemma}
\begin{proof}
This follows trivially from the fact that d-separation is a special case of e-separation (when $D=\emptyset$ in Definition \ref{def_esep}).
\end{proof}

\begin{lemma}[The comparison of e-separation relations subsumes the comparison of skeletons]
\label{lemma_esep_skeleton}
Let $\mathfrak{G}$ and $\mathfrak{G}'$ be two mDAGs such that  $\mathfrak{G}$ can be shown to \emph{not} observationally dominate  $\mathfrak{G}'$ by comparison of skeletons. Then, $\mathfrak{G}$ can also be shown to not observationally dominate  $\mathfrak{G}'$ by comparison of e-separation relations.
\end{lemma}
\begin{proof}
Suppose that the nodes $u$ and $v$ are connected by an edge in the skeleton of $\mathfrak{G}'$ but not in the skeleton of $\mathfrak{G}$. This means that, in $\mathfrak{G}$, there is neither a directed edge nor a confounder between $u$ and $v$. It is then clear that, if we delete all other nodes of $\mathfrak{G}$, $u$ and $v$ are going to be d-separated by the empty set.
In $\mathfrak{G}'$, on the other hand, $u$ and $v$ are either connected by a directed edge or are part of the same facet. Thus, if we delete all the other nodes, $u$ and $v$ are not going to be d-separated by the empty set.
Therefore, $\mathfrak{G}$ has the e-separation relation $u\esep v | \neg (\nodes(\mathfrak{G})\setminus\{u,v\})$, while $\mathfrak{G}'$ \emph{does not} have it. Thus, the comparison of e-separation relations can show that $\mathfrak{G}$ that does not observationally dominate  $\mathfrak{G}'$.
\end{proof}

\subsubsection{Pairs of Densely Connected Nodes}
\label{sec_dense_connected}

In this section, we show a rule for observational nondominance derived from the main result of Ref.~\cite{evans_dependency}. This rule is based on the concept of \emph{densely connected nodes}, first defined in Ref.~\cite{SEMs}. This concept makes use of the preliminary definitions of \emph{districts} of an mDAG and the \emph{closure} of a set of nodes. The concept of districts, first introduced in Ref.~\cite{TianandPearl}, is reproduced below.

\begin{definition}[Districts]
\label{def_districts}
Let $\mathfrak{G}=\{\cal D, B\}$ be an mDAG. Let $D$ be a set of nodes of $\mathfrak{G}$ that are connected to each other through a set of facets of $\mathcal{B}$; i.e., if $v,w\in D$, then there is a sequence of nodes $(u_0,u_1,u_2,...,u_{n-1},u_n)$ with $u_0=v$ and $u_n=w$ such that for all $i$, $\{u_i, u_{i+1}\}\subseteq B_i \in \mathcal{B}$. If $D$ is 
inclusion-maximal   (i.e., maximal in the order 
 induced by subset inclusion),
  we say that it is a \emph{district}.

If $v$ is a node of $\mathfrak{G}$, we are going to denote the district of $\mathfrak{G}$ to which $v$ belongs by $\text{dis}_{\mathfrak{G}}(v)$. Similarly, for a set of nodes $S$, we have  $\text{dis}_{\mathfrak{G}}(S)=\cup_{v\in S} \text{dis}_{\mathfrak{G}}(v)$.
\end{definition}

\begin{figure}[h!]
\centering
\includegraphics[width=0.4\textwidth]{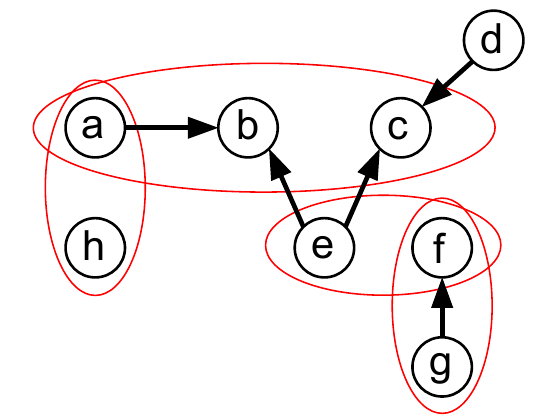}
\caption{The districts of this mDAG are $\{a,b,c,h\}$, $\{d\}$ and $\{e,f,g\}$.} 
\label{fig_def_districts_example} 
\end{figure}

In the mDAG of Fig.~\ref{fig_def_districts_example}, for example, we have three districts: $\{a,b,c,h\}$, $\{d\}$ and $\{e,f,g\}$. 

The second preliminary definition we need is the \emph{closure} of a set of nodes, first introduced in Ref.~\cite{SEMs}.  This is just a step towards defining densely connected pairs of nodes, which is the important definition for us. 

\begin{definition}[Closure of a set of nodes]

Let $\mathfrak{G}=\{\cal D,B\}$ be an mDAG, and let $A\subseteq \nodes(\mathfrak{G})$. Set $A^{(0)}\equiv \nodes(\mathfrak{G})$ and construct a sequence of sets of nodes by applying alternately the following rules:
\begin{align*}
	&A^{(i+1)}=\text{dis}_{\mathfrak{G}_{A^{(i)}}}(A) \quad \text{,} &A^{(i+2)}=\an_{\mathcal{D}_{A^{(i+1)}}}(A)
\end{align*}
Where $\text{dis}_{\mathfrak{G}}(S)$ is the district of $S$ in $\mathfrak{G}$, and $\an_{\mathcal{D}}(S)$ is the set of ancestors of the nodes of $S$ in the directed structure (which includes the nodes of $S$ themselves). As usual, $\mathfrak{G}_S$ is the subgraph of  $\mathfrak{G}$ over nodes $S$, and $\mathcal{D}_S$ is the subgraph of  $\cal D$ over nodes $S$. Each one of these operations removes nodes from the set $A^{(i)}$ or keeps it the same, the case which indicates the termination of the process. When the process terminates, we call the resulting set the closure of $A$ and denote it $\langle A \rangle_{\mathfrak{G}}$.
\end{definition}

\begin{figure}[!h]
	\captionsetup[subfigure]{aboveskip=-2pt,belowskip=-1pt}
	\centering
		\begin{subfigure}[b]{0.2\textwidth}
		\centering
		\includegraphics[width=\textwidth]{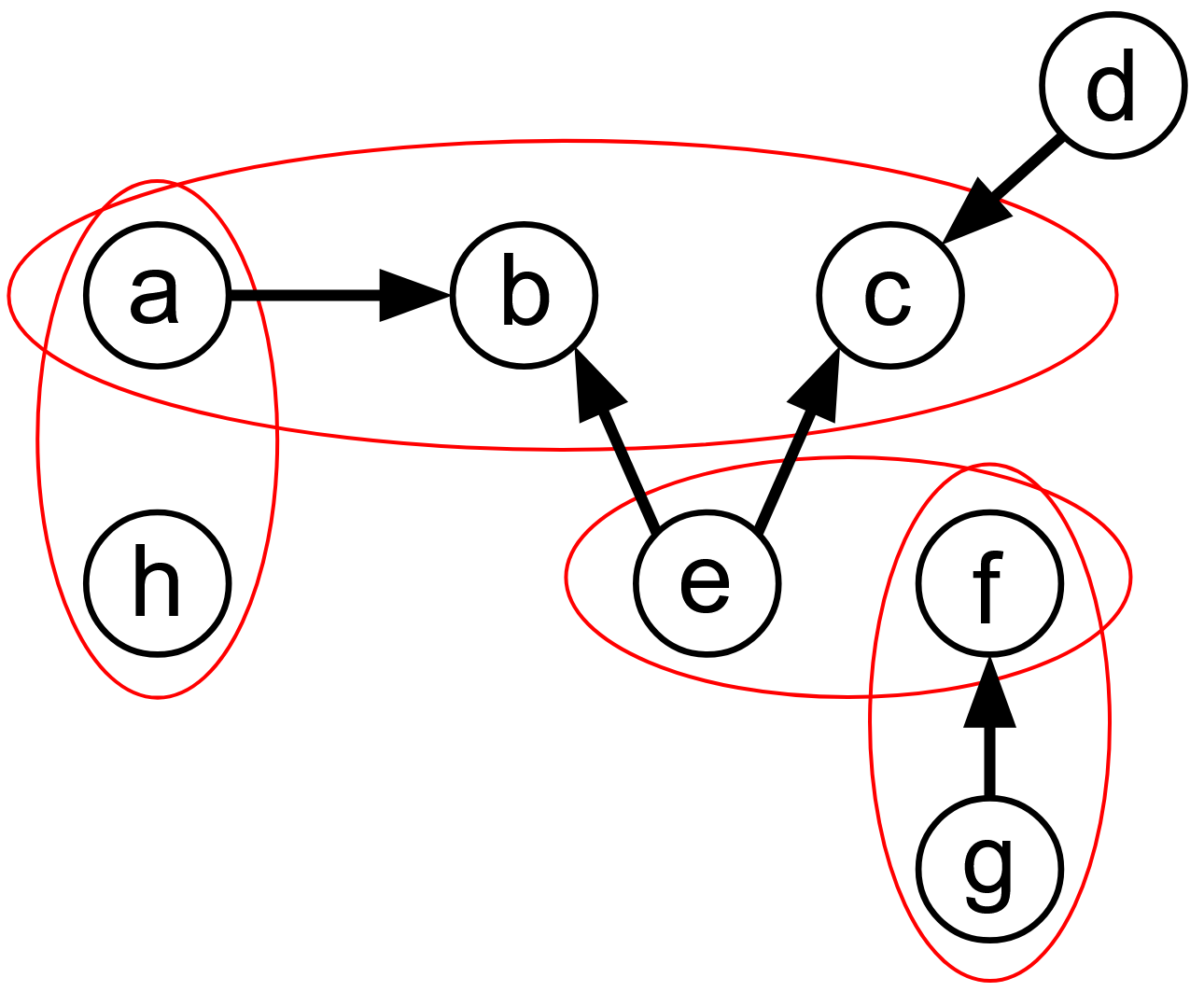}
		\caption{}
		\label{fig_G_A0}
	\end{subfigure}
	\hspace{1mm}
	\begin{subfigure}[b]{0.2\textwidth}
		\centering
		\includegraphics[width=\textwidth]{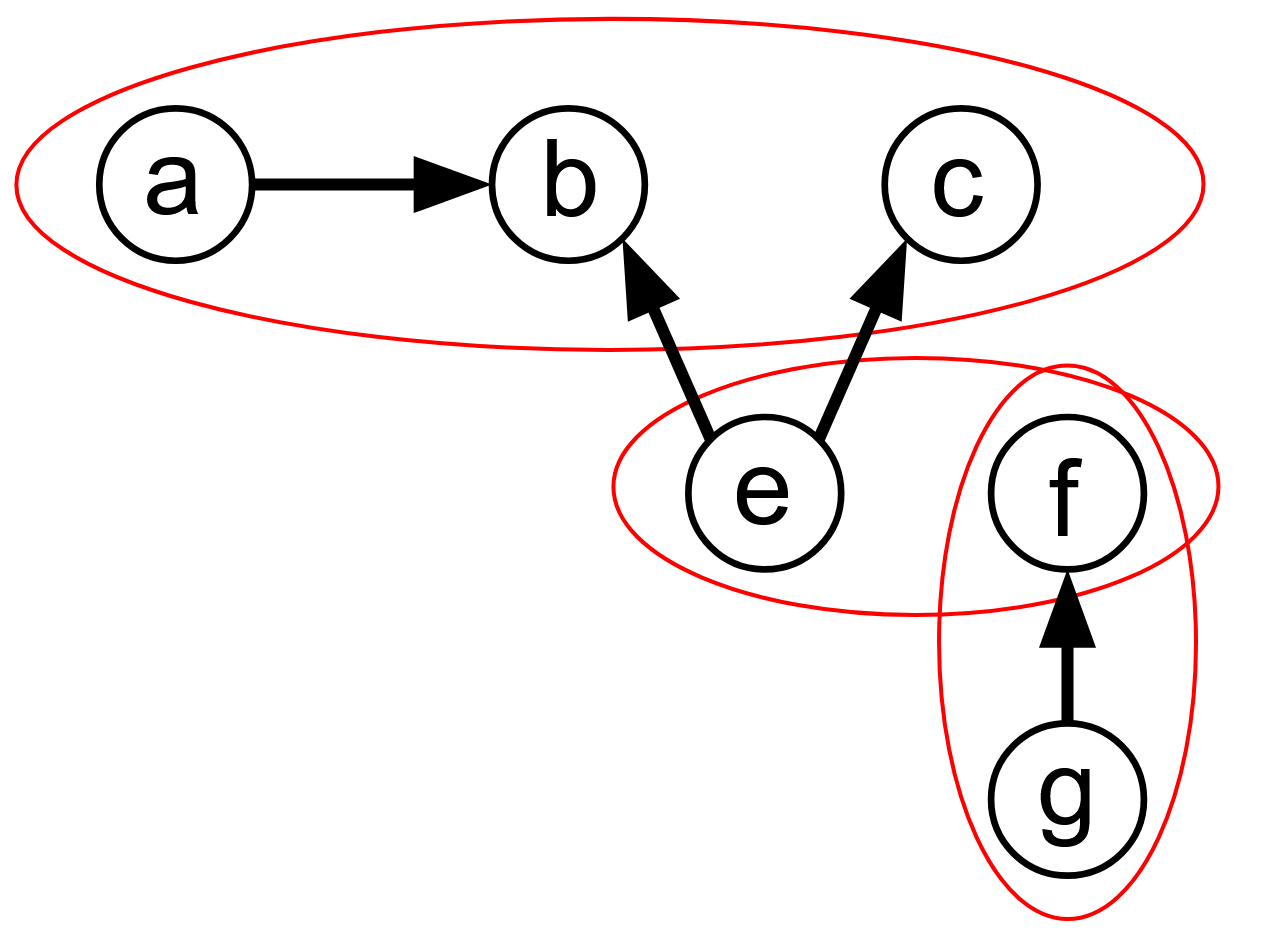}
		\caption{}
		\label{fig_G_A1}
	\end{subfigure}
	\hspace{1mm}
	\begin{subfigure}[b]{0.2\textwidth}
		\centering
		\includegraphics[width=\textwidth]{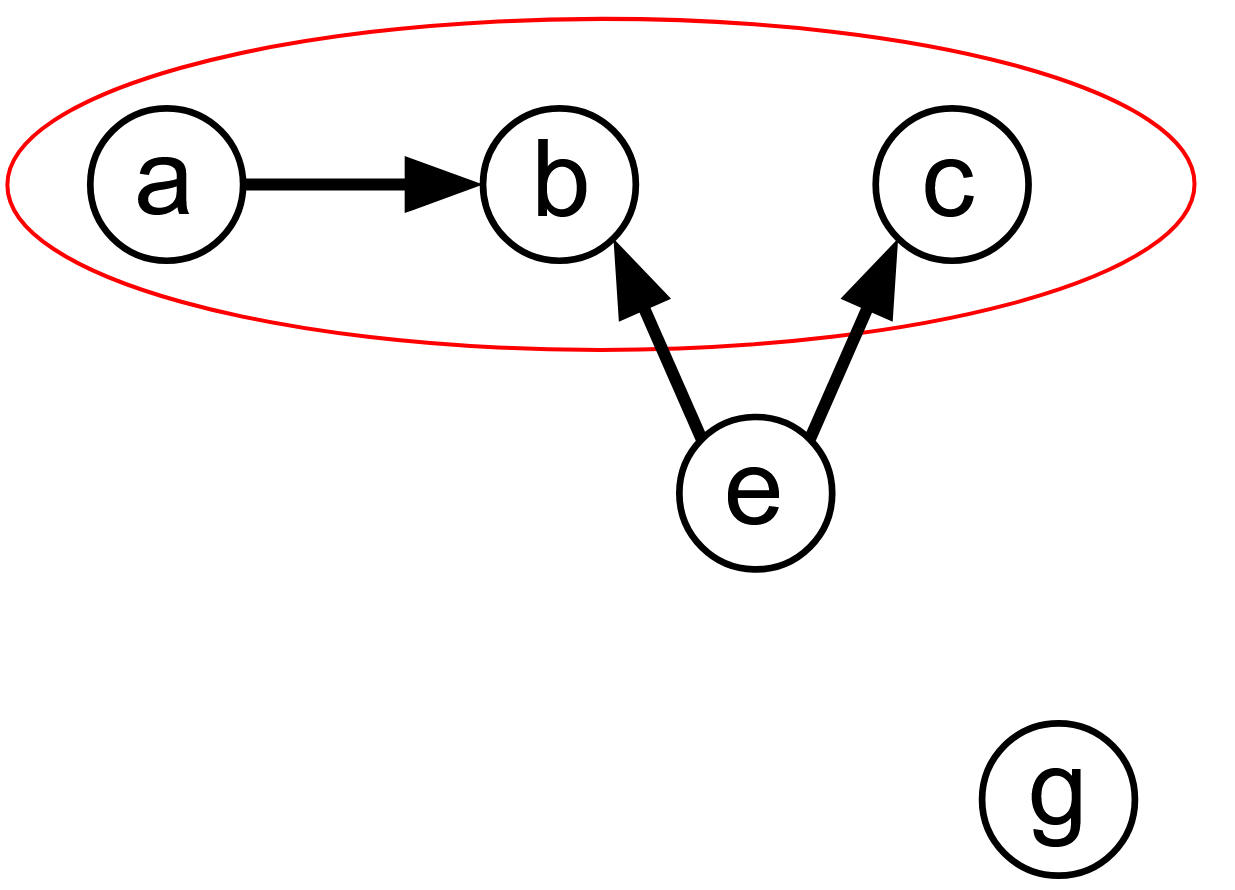}
		\caption{}
		\label{fig_G_A2}
	\end{subfigure}
	\hspace{1mm}
	\begin{subfigure}[b]{0.2\textwidth}
	\centering
	\includegraphics[width=\textwidth]{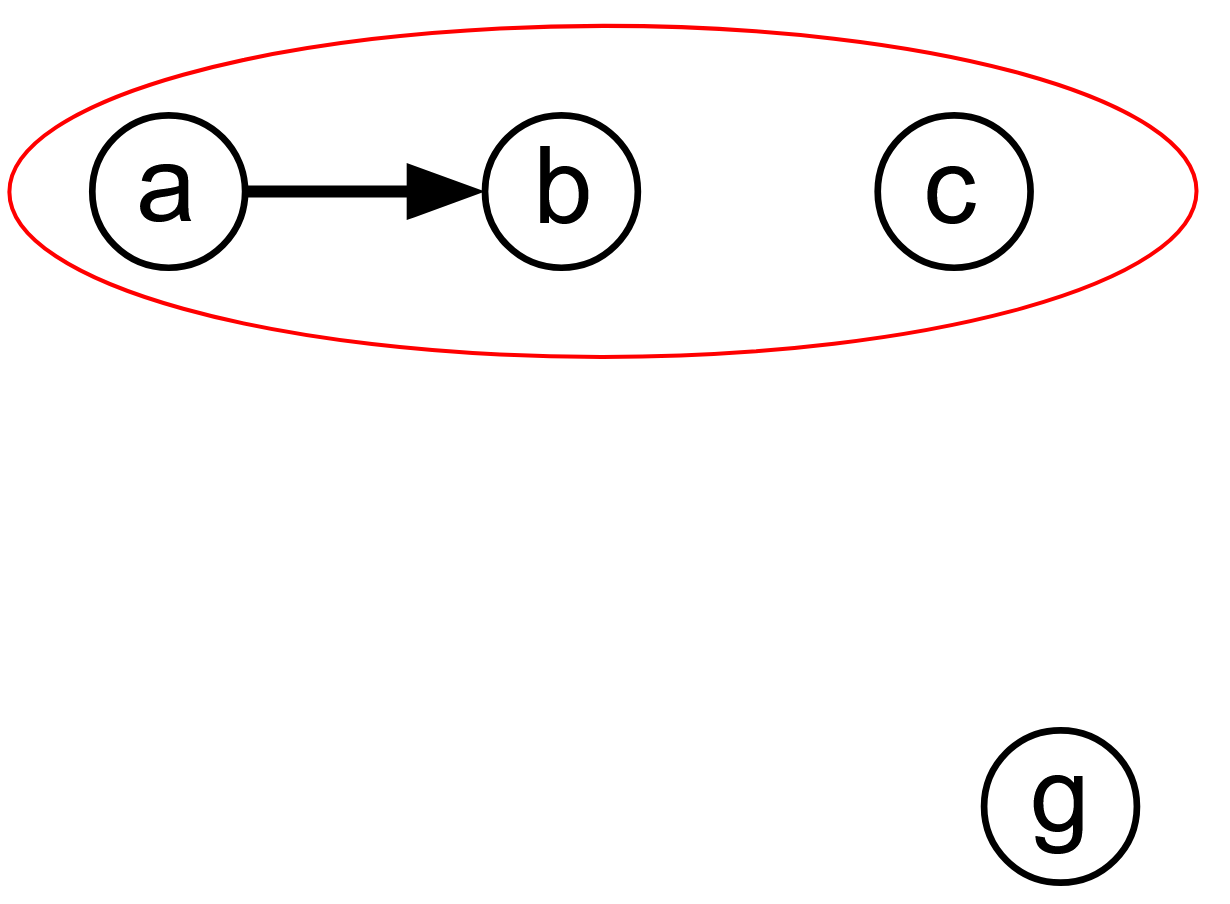}
	\caption{}
	\label{fig_G_A3}
\end{subfigure}
	\caption{Steps used to find the closure of the set $A=\{b,c,g\}$ in the mDAG of Fig.~\ref{fig_def_districts_example}.}
	\label{fig_example_closure}
\end{figure}

Let us go through an example of this definition. Say that we want to know what is the closure of the set $A=\{b,c,g\}$ in the mDAG $\mathfrak{G}$ of Fig.~\ref{fig_def_districts_example}. The set $A^{(0)}$ is the entire set of nodes of the mDAG, so we have that $\mathfrak{G}_{A^{(0)}}=\mathfrak{G}$. This mDAG is reproduced in Fig.~\ref{fig_example_closure}(a). It implies:
\begin{equation*}
	A^{(1)}=\text{dis}_{\mathfrak{G}}(A)=\{a,b,c,e,f,g\}.
\end{equation*}

Therefore, $\mathfrak{G}_{A^{(1)}}$ is the sub-mDAG of $\mathfrak{G}$ shown in Fig.~\ref{fig_example_closure}(b). This in turn gives us
\begin{equation*}
	A^{(2)}=\an_{\mathcal{D}_{A^{(1)}}}(A)=\{a,b,c,e,g\},
\end{equation*}
where $\mathcal{D}_{A^{(1)}}$ is the directed structure of $\mathfrak{G}_{A^{(1)}}$. This implies that  $\mathfrak{G}_{A^{(2)}}$ is the mDAG of Fig.~\ref{fig_example_closure}(c), which gives:
\begin{equation*}
	A^{(3)}=\text{dis}_{\mathfrak{G}_{A^{(2)}}}(A)=\{a,b,c,g\}.
\end{equation*}
Then, $\mathfrak{G}_{A^{(3)}}$ is the mDAG of Fig.~\ref{fig_example_closure}(d). Therefore:
\begin{equation*}
	A^{(4)}=\an_{\mathcal{D}_{A^{(3)}}}(A)=\{a,b,c,g\}.
\end{equation*}

The sets $A^{(3)}$ and $A^{(4)}$ are equal, which indicates the termination of the process. Therefore, in this example $\langle A \rangle_{\mathfrak{G}}=\{a,b,c,g\}$.

Now we can finally present the definition of dense connectedness between nodes, taken from Ref.~\cite{SEMs}: 

\begin{definition}[Densely connected pair of nodes]
	Let $\mathfrak{G}=\{\cal D, B\}$ be an mDAG. A pair of nodes $v\neq w \in \nodes(\mathfrak{G})$ is said to be \emph{densely connected} if any of the following conditions are satisfied:
\begin{itemize}
	\item $v\in \pa_{\mathcal{D}}(\langle w \rangle_{\mathfrak{G}})$.
	\item $w\in \pa_{\mathcal{D}}(\langle v \rangle_{\mathfrak{G}})$.
	\item $\langle \{v,w\} \rangle_{\mathfrak{G}}$ is a bidirected-connected set.
\end{itemize}
where we say that a set of nodes $S$ is bidirected-connected if it forms a district of the mDAG $\mathfrak{G}_S$, the mDAG obtained by deleting all of the nodes of $\mathfrak{G}$ that are not in $S$. In other words, the set $S$ is bidirected-connected if  every vertex of $S$ can be reached from every other using a path of nodes, all in $S$, that are connected by facets of $\cal B$.
\end{definition}

For example, one can check that the nodes $b$ and $h$ in the mDAG $\mathfrak{G}$ of Fig.~\ref{fig_def_districts_example} are densely connected in $\mathfrak{G}$, because $\langle \{b,h\} \rangle_{\mathfrak{G}}=\{a,b,h\}$, which is a bidirected-connected set in $\mathfrak{G}$. On the other hand, $b$ and $g$ are not densely connected, because $b$ is not a parent of $\langle \{g\} \rangle_{\mathfrak{G}}=\{g\}$, $g$ is not a parent of $\langle \{b\} \rangle_{\mathfrak{G}}=\{a,b\}$ and $\langle \{b,g\} \rangle_{\mathfrak{G}}=\{a,b,g\}$ is not bidirected-connected.

Due to the main result of Ref.~\cite{evans_dependency},  the comparison of pairs of densely connected nodes is a nondominance-proving rule: 

\begin{proposition}[Comparison of densely connected pairs of nodes  - Consequence of Ref.~\protect{\cite[Theorem 6.4]{evans_dependency}}]
	Let $\mathfrak{G}$ and $\mathfrak{G}'$ be two mDAGs such that $\nodes(\mathfrak{G})=\nodes(\mathfrak{G}')$. If there is an pair of nodes $v,w \in \nodes(\mathfrak{G})$ that is densely connected in $\mathfrak{G}'$ but not in $\mathfrak{G}$, then $\mathfrak{G}$ \emph{does not} observationally dominate $\mathfrak{G}'$, i.e., $\mathfrak{G}\not\succeq \mathfrak{G}'$.
\end{proposition}
\begin{proof}
	Ref.~\cite{evans_dependency} showed that the specific probability distribution where $X_v$ and $X_w$ are perfectly correlated and independent of all other visible variables (which are uniformly distributed) is realizable by a causal structure \emph{if and only if} the nodes $v$ and $w$ are densely connected in that causal structure. Therefore, this distribution is realizable by $\mathfrak{G}'$ but not by $\mathfrak{G}$, which implies that $\mathfrak{G}$ {does not} observationally dominate $\mathfrak{G}'$.
\end{proof}

\begin{figure}[h!]
	\centering
	\includegraphics[width=0.45\textwidth]{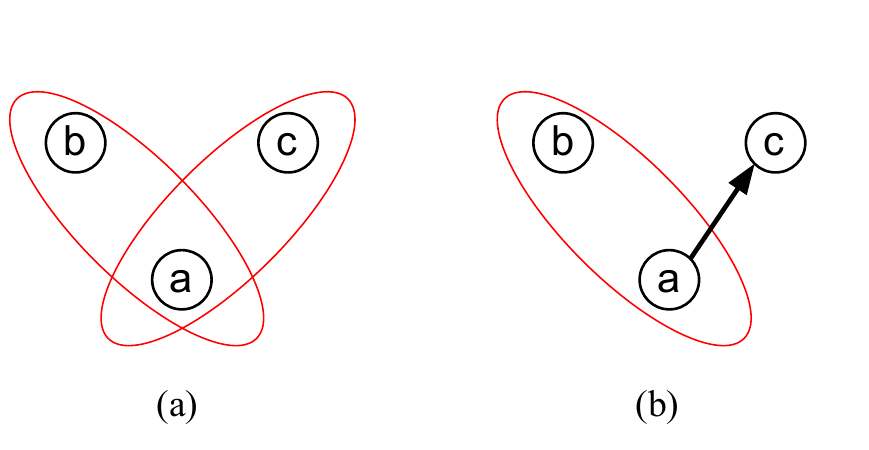}
	\caption{Example of a pair of mDAGs that have the same pairs of densely connected nodes, but different sets of e-separation relations (which in this case are simply d-separation relations). In Fig.~\ref{fig_temporally_ordered_3_nodes}, the mDAG (a) is in the {\tt Collider A} observational equivalence class, while (b) is in the {\tt Fork} observational equivalence class.} 
	\label{fig_collider_chain_comparison} 
\end{figure}

\begin{figure}[!h]
\centering
\includegraphics[width=0.45\textwidth]{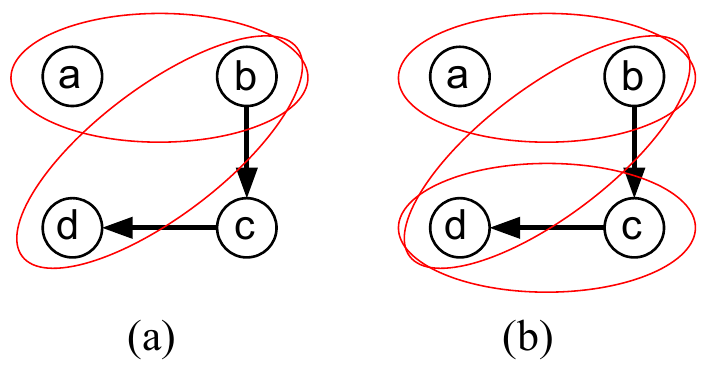}
\caption{Example of a pair of mDAGs that have the same sets of e-separation relations, but different pairs of densely connected nodes. The e-separation relations of both of these mDAGs are $a\dsep c|\neg b$, $a\dsep c|\neg bd$, $a\dsep c| b$, $a\dsep c| b \neg d$, $a\dsep c|d\neg b$, $a\dsep d|\neg b$, $a\dsep d|\neg bc$, $a\dsep d|\neg c$ and $a\dsep d|c\neg b$. Meanwhile, the pair of nodes $(a,d)$ is densely connected in (b) but not in (a).}
\label{fig_DC_comparison}
\end{figure}

The comparison of densely connected pairs of nodes is {not} redundant to the comparison of e-separation relations; both techniques can show observational inequivalences that the other one cannot. Examples of mDAGs that can be shown to be observationally inequivalent by comparison of e-separation relations but not by pairs of densely connected nodes are found already in the three-node case; an explicit example is shown in Fig.~\ref{fig_collider_chain_comparison}. On the other hand, there are no three-node mDAGs that can be shown observationally inequivalent by comparison of densely connected pairs of nodes but not by comparison of e-separation relations. Among four-node mDAGs, however, such examples appear. One such example is shown in Fig.~\ref{fig_DC_comparison}.

\subsubsection{Directed-edge-free rule}
\label{sec_only_hypergraphs}

In this section, we discuss a rule for showing observational nondominance that is based on Proposition~\ref{directed_edge_free_dominance}:

\begin{proposition}[Directed-edge-free rule  - Consequence of Ref.~\protect{\cite[Example 2]{steudel_ay_2015}}]
	Let $\mathfrak{G}$ and $\mathfrak{G}'$ be two mDAGs. Suppose $\mathfrak{G}$  is observationally equivalent to the directed-edge-free mDAG $\tilde{\mathfrak{G}}$ and  $\mathfrak{G}'$  is observationally equivalent to the directed-edge-free mDAG $\tilde{\mathfrak{G}'}$. Then, $\mathfrak{G}$ observationally dominates $\mathfrak{G}'$ if and only if $\tilde{\mathfrak{G}}$ structurally dominates $\tilde{\mathfrak{G}'}$.
	\label{corr_DEF}
\end{proposition}
\begin{proof}
	This is a straightforward corollary of Proposition~\ref{directed_edge_free_dominance}.
\end{proof}

 Note that this rule has a difference from the other graphical rules shown in this section. The other rules assign a signature to each mDAG, such as their skeleton or their set of d-separation relations, and then compare this signature. Because of that, the other rules can be applied to any pair of mDAGs, even though sometimes they give inconclusive results (for example when neither set of d-separation relations is contained in the other).    
Proposition~\ref{corr_DEF}, on the other hand, can only be applied to compare  mDAGs that are equivalent to a directed-edge-free mDAG. This makes the implications of this rule for the proven-inequivalence partition a bit more subtle than before.

\begin{figure*}[h!]
\centering
\includegraphics[width=0.8\textwidth]{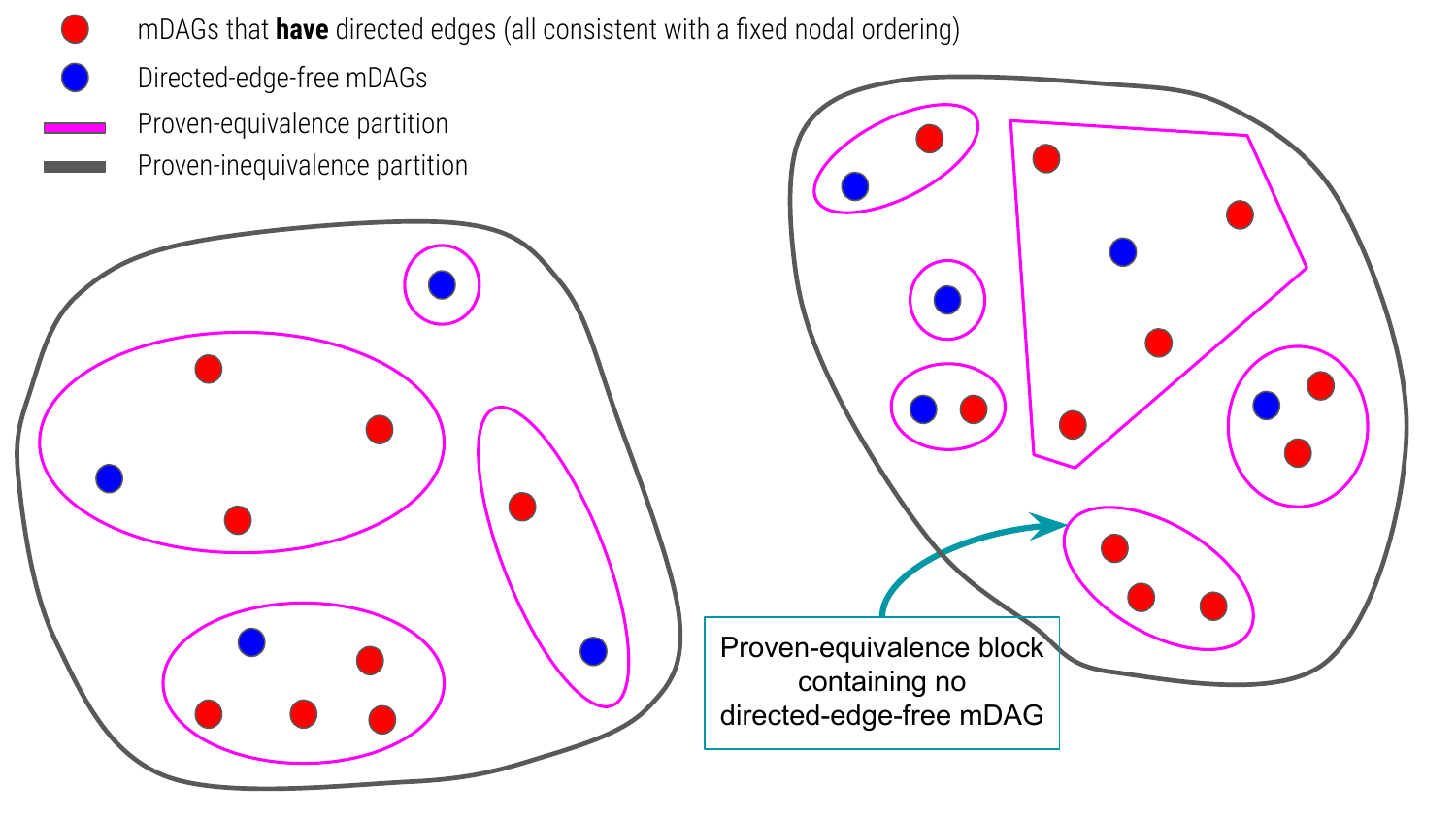}
\caption{The directed-edge-free mDAGs are represented by blue dots, while the other mDAGs are red dots. Suppose that nondominance-proving rules other than the directed-edge-free rule have been used to obtain the proven-inequivalence partition depicted. The directed-edge-free rule says that the proven-inequivalence block on the left should split into the four proven-equivalence blocks that live inside it. On the other hand, the proven-inequivalence block on the right includes one distinguished proven-equivalence block that does not contain any directed-edge-free mDAG. Therefore, the directed-edge-free rule cannot be used to split the proven-inequivalence block on the right. } 
\label{fig_hypergraph_only} 
\end{figure*}

This is explained in Fig.~\ref{fig_hypergraph_only}. There, we \emph{can} split the proven-inequivalence block (gray loop) on the left into four, because all the blocks of the proven-equivalence partition (pink loops) inside it contain a directed-edge-free mDAG, so Proposition~\ref{directed_edge_free_dominance} implies that they are all observationally inequivalent. The proven-inequivalence block on the right, on the other hand, contains a distinguished proven-equivalence block that does not contain a directed-edge-free mDAG. Therefore, the proven-inequivalence block on the right still \emph{cannot} be split, because the distinguished proven-equivalence block inside it cannot be compared to the others by Proposition~\ref{directed_edge_free_dominance}.

The directed-edge-free rule is \emph{not} redundant to the comparison of e-separation relations nor to the comparison of densely connected pairs of nodes; it can show nondominances that the other two rules cannot, and vice versa.  
The fact that the mDAG of Fig.~\ref{fig_triangle}(a) does not observationally dominate the mDAG of Fig.~\ref{fig_triangle}(b) can be shown by the directed-edge-free rule, but not by the comparison of e-separation relations or the comparison of densely connected pairs. The fact that the mDAGs of  Fig.~\ref{fig_collider_chain_comparison} do not observationally dominate each other (i.e., they are incomparable) can be shown by comparison of e-separation relations, but not by  the directed-edge-free rule (nor by the  comparison of densely connected pairs of nodes, as we have seen). The directed-edge-free rule cannot be applied here because, as we can see from Fig.~\ref{fig_temporally_ordered_3_nodes}, the mDAG of Fig.~\ref{fig_collider_chain_comparison}(b) is not observationally equivalent to any directed-edge-free mDAG. The fact that the mDAG of Fig.~\ref{fig_DC_comparison}(a) does not observationally dominate the mDAG of Fig.~\ref{fig_DC_comparison}(b) can be shown by the comparison of densely connected pairs of nodes, but not by  the directed-edge-free rule (nor by the comparison of e-separation relations, as we have seen). The directed-edge-free rule cannot be applied here because those two mDAGs are not equivalent to any directed-edge-free mDAG. To prove this we note that, as shown in Example 2 of Ref.~\cite{steudel_ay_2015}, perfect correlation between a set of nodes is realizable if and only if these nodes share a common ancestor in the causal structure (that can be visible or latent). This implies that both mDAGs of Fig.~\ref{fig_DC_comparison} are able to realize perfect correlation between the variables $X_a$, $X_b$, $X_c$ and $X_d$. Therefore, if Fig.~\ref{fig_DC_comparison}(b) were observationally equivalent to a directed-edge-free mDAG, it should be the saturated mDAG, where all four nodes are in the same facet (and thus all four nodes share a common ancestor in the corresponding canonical pDAG). This cannot be the case, however, since Fig.~\ref{fig_DC_comparison}(b) is \emph{not} saturated (this is easy to see, for example, by noting that nodes $a$ and $d$ are not connected in its skeleton).

\subsection{Comparison of unrealizable supports}
\label{sec_unrealizable_supports}

In this section, we will present a rule to establish observational nondominance based on the work of Fraser~\cite{Fraser_Combinatorial_Solution}. This rule deals with the problem of \emph{possibilistic causal realizability}, which we will define shortly.

An important \emph{caveat} on this method is that it can only be applied when the visible variables have {finite cardinalities}. As already stated, this is not a problem for our purposes: to show observational inequivalence, it suffices to show that there is a difference in the set of realizable distributions for one specific assignment of cardinalities. 

The problem that motivates the work of this paper is the problem of probabilistic causal realizability (Definition \ref{SEP_definition}), which asks whether a probability distribution is realizable by a causal structure or not. In this section, we turn our attention to a different, strictly weaker, question: the \emph{possibilistic} causal realizability problem. To define it, let us first define the support of a probability distribution:

\begin{definition}[Support] Let $P:\mathcal{X}\rightarrow [0,1]$ be a probability distribution. The elements of $\mathcal{X}$ are called ``events''. The \underline{support} of $P$ is the set of events that are deemed to be possible, in the sense of having nonzero probability:
\begin{equation*}
	\text{\em Supp}(P)=\{\omega\in\mathcal{X} | P(\omega)>0\}
\end{equation*}

\end{definition}

A set of events $S$ is said to be a realizable support of the pDAG $\cal G$ if there is \emph{some} distribution $P$ that is realizable by $\cal G$ such that $\text{Supp}(P)=S$. Let  $\mathcal{S}(\mathcal{G},\vec{c}_\text{vis} )$ denote the set of all sets of events that are realizable by the pDAG $\cal G$ for cardinalities of the visible variables given by $\vec{c}_\text{vis}$. Ref.~\cite{Fraser_Combinatorial_Solution} presented an algorithm to find the set $\mathcal{S}(\mathcal{G},\vec c_\text{vis})$. We refer to it here as \emph{Fraser's algorithm}. This algorithm is explained in Appendix \ref{appendix_supports}.

 Just as we defined the observational profile $\Mtower(\cal G)$ of a pDAG $\cal G$, we can define $\cal G$'s \emph{possibilistic observational profile} by:
\begin{equation}
\Stower(\mathcal{G})\coloneqq\left(\mathcal{S}(\mathcal{G},\vec{c}_\text{vis} ): \vec c_\text{vis} \in\mathbb{N}^{|\vis(\mathcal{G})|}\right),
\end{equation}
where again the parentheses indicate that this object behaves as a tuple: each $\mathcal{S}(\mathcal{G},\vec{c}_\text{vis})$ is indexed by $\vec{c}_\text{vis}$.

If the support of a probability distribution $P$ is \emph{not} realizable by $\cal G$, then clearly $P$ is not realizable by $\cal G$. Thus, if there is some assignment of cardinalities for which a pDAG $\cal G$ cannot realize a support that is realizable by another pDAG $\cal G'$, then $\cal G$ cannot observationally dominate $\cal G'$. In other words, we have:
\begin{gather}
	\Stower(\mathcal{G}) \not\subseteq_\text{ew} \Stower(\mathcal{G}') \nonumber \\ \Rightarrow 
	\label{eq_possibilistic_open_question} \\ \Mtower(\mathcal{G})\not\subseteq_\text{ew}\Mtower(\mathcal{G}') \nonumber
\end{gather}
where $\not\subseteq_\text{ew}$ denotes the failure of element-wise inclusion,  which is defined for the possibilistic observational profile in an analogous way to which it was defined for the observational profile in Definition~\ref{def_elementwise}.  From this implication, we know that the comparison of unrealizable supports gives us a nondominance-proving rule:
\begin{proposition}[Comparison of unrealizable supports]
	Let $\mathfrak{G}$ and $\mathfrak{G}'$ be two mDAGs such that $\nodes(\mathfrak{G})=\nodes(\mathfrak{G}')$. If there is a support that is realizable by $\mathfrak{G}'$ but not by $\mathfrak{G}$, then $\mathfrak{G}$ \emph{does not} observationally dominate $\mathfrak{G}'$, i.e., $\mathfrak{G}\not\succeq \mathfrak{G}'$.
\end{proposition}

Note that it is unclear whether the converse implication to that of \eqref{eq_possibilistic_open_question} holds.  Thus, we formalize it as a conjecture:

\begin{conjecture}
	\label{conjecture_converse_of_12}
	If it is possible to find a \emph{probability distribution} that is realizable by the pDAG $\mathcal{G}$ but not by the pDAG $\mathcal{G}'$, then there is a \emph{support} that is realizable by $\mathcal{G}$ but not by $\mathcal{G}'$. In other words, the converse of Eq.~\eqref{eq_possibilistic_open_question} holds.
\end{conjecture}

We will discuss potential prospects to tackle this open question in Section \ref{sec_every_support}.

\subsubsection{Subsumption of other nondominance-proving rules}

In Lemmas~\ref{lemma_esep_dsep} and~\ref{lemma_esep_skeleton}, we showed that the comparison of e-separation relations subsumes the comparison of skeletons and d-separation relations as rules  to show observational nondominance. In this section, we discuss how the comparison of unrealizable supports subsumes other rules.

Since Fraser's algorithm for unrealizable supports is a brute force search, one wants to apply as many of the weaker but computationally cheaper rules as possible before implementing the comparison of unrealizable supports. However, it is possible to show that all of the cheaper rules --- comparison of e-separation relations,  comparison of densely connected pairs of nodes, and the directed-edge-free rule --- are subsumed by the comparison of unrealizable supports. In fact, to subsume the cheaper rules it is sufficient to look at supports at the level of binary visible variables; that is, whenever a nondominance can be shown by one of the cheaper rules, one can find a support with $\vec c_\text{vis}=(2,2,2,...)$ that also shows said nondominance.  

\begin{restatable}[The comparison of unrealizable supports subsumes the comparison of e-separation relations]{lemma}{esep}
\label{lemma_sup_esep}
Let $\mathfrak{G}$ and $\mathfrak{G}'$ be two mDAGs that have different sets of e-separation relations. Then, it is possible to find a support which is realizable by one of these mDAGs but not by the other. In fact, it is possible to find such a support even when all of the visible variables are binary, i.e., $\vec c_\text{vis}=(2,2,2,\dots)$. 
\end{restatable}

The proof is given in Appendix~\ref{appendix_proofs_subsume}. The idea behind this proof is that the set of e-separation relations of a given mDAG might already imply that some supports are unrealizable. As an example of this, imagine a causal structure that presents the d-separation relation $a\dsep b$. Suppose that $S$ is a support over binary variables $X_a$ and $X_b$ that does not contain any events involving $\{X_a=0, X_b=1\}$ or $\{X_a=1,X_b=0\}$. This implies that in any probability distribution $P$ that has the support $S$, i.e., where $\text{Supp}(P)=S$, the variables $X_a$ and $X_b$ are correlated. This is incompatible with  $a\dsep b$, thus the support $S$ is ruled out by $a\dsep b$. To save time in the computation, we only apply Fraser's algorithm for supports that are not already ruled out by the e-separation relations of the mDAG in question.

\begin{restatable}[The comparison of unrealizable supports subsumes the comparison of densely connected pairs of nodes]{lemma}{DC}
	\label{lemma_sup_DC}
	Let $\mathfrak{G}$ and $\mathfrak{G}'$ be two mDAGs such that  $\mathfrak{G}'$ can be shown to \emph{not} observationally dominate  $\mathfrak{G}$ by comparison of densely connected pairs of nodes. Then, $\mathfrak{G}'$ can also be shown to not observationally dominate  $\mathfrak{G}$ by comparison of  unrealizable supports. In fact, it is possible to do so even when we are only comparing unrealizable supports where all of the visible variables are binary, i.e., $\vec c_\text{vis}=(2,2,2,\dots)$.
\end{restatable}

The proof is given in Appendix~\ref{appendix_proofs_subsume}.   The idea behind this proof is as follows: as  already discussed, Ref.~\cite{evans_dependency} says that the specific distribution where all of the visible variables are binary and $X_v$ and $X_w$ are perfectly correlated while the other variables are all mutually independent and uniformly distributed is realizable by an mDAG if and only if $v$ and $w$ are densely connected in this mDAG. Appendix~\ref{appendix_proofs_subsume} then shows that not only this specific probability distribution, but also its \emph{support} is realizable by an mDAG if and only if $v$ and $w$ are densely connected in this mDAG. More specifically, this support is the one where $X_v$ and $X_w$ are both $0$ in half of the events and both $1$ in the other half, while the other visible variables take all possible combinations of $0$ and $1$. The explicit construction of such a support for the case of four visible nodes is given in Eq.~\eqref{eq_appendix_denseconnected_support}.

\begin{restatable}[The comparison of unrealizable supports subsumes the directed-edge-free rule]{lemma}{DEF}
\label{lemma_sup_OH}
	Let $\mathfrak{G}$ and $\mathfrak{G}'$ be two mDAGs such that  $\mathfrak{G}'$ can be shown to \emph{not} observationally dominate  $\mathfrak{G}$ by the directed-edge-free rule. Then, $\mathfrak{G}'$ can also be shown to not observationally dominate  $\mathfrak{G}$ by comparison of unrealizable supports. In fact, it is possible to do so even when we are only comparing unrealizable supports where all of the visible variables are binary, i.e., $\vec c_\text{vis}=(2,2,2,\dots)$, and the supports are constituted by only two events.
\end{restatable}

The proof of this lemma is given in Appendix~\ref{appendix_proofs_subsume},  where an explicit construction of the support that is realizable by $\mathfrak{G}$ but not $\mathfrak{G}'$ is provided.

To conclude this section, note that unlike the case of dominance-proving rules, for the application of nondominance-proving rules we \emph{do not} need to go outside of the set of mDAGs consistent with a fixed nodal ordering. This is because, while the dominance-proving rules work by creating a sequence of mDAGs such that each one dominates the next, most of the nondominance-proving rules work by assigning a certain signature to each mDAG (such as its set of unrealizable supports or d-separation relations, for example), and comparing the signatures of two mDAGs. Therefore, such rules would not benefit by the inclusion of mDAGs that are not consistent with the fixed nodal ordering. As we will see, the only nondominance-proving rule that does not work in this way is the directed-edge-free rule, but this is not an issue because all of the directed-edge-free mDAGs are consistent with any nodal ordering.

\section{Deriving the Observational Partial Order for 3-node mDAGs}
\label{sec_back_to_3visible}

Now that we presented our rules for proving observational dominance and nondominance between mDAGs, we will go back to the results for 3-node mDAGs consistent with a fixed nodal ordering, presented in Section \ref{sec_3_observed_nodes}, and explain how they are derived.

\subsection{Observational Dominances}
\label{section_3nodes_observationalequivalences}

\begin{figure}[h!]
\centering
\includegraphics[width=0.45\textwidth]{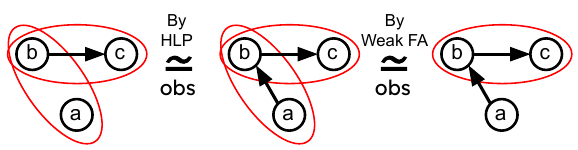}
\caption{Three observationally equivalent mDAGs, that lie in the {\tt Instrumental ABC} class.} 
\label{fig_instrumental_class} 
\end{figure}

Take, for example, the mDAGs in the {\tt Instrumental ABC} class of Fig.~\ref{fig_temporally_ordered_3_nodes}, that are reproduced in Fig.~\ref{fig_instrumental_class}. As we will see now, the rules discussed in Section \ref{sec_show_equivalence} establish that these three mDAGs are observationally equivalent to each other. Here, we will discuss the intuition behind those rules for the specific case at hand.

In the first mDAG of Fig.~\ref{fig_instrumental_class}, the node $b$ already has access to all of the parents of $a$. Since the local randomness of $a$ can also be absorbed by the latent parent that it shares with $b$, the node $b$ already has access to all information about $a$. Therefore, this mDAG is just as powerful as another one where an edge between $a$ and $b$ is added, namely the second mDAG of Fig.~\ref{fig_instrumental_class}. This is the application of the HLP Edge-Adding  Rule (Proposition~\ref{HLP}) to show the observational equivalence between the two first mDAGs of Fig.~\ref{fig_instrumental_class}.

\begin{figure}[h!]
 	\centering
 	\includegraphics[width=0.45\textwidth]{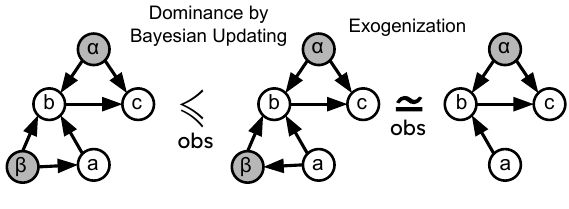}
 	\caption{Intermediary steps to show observational equivalence between the second and the third mDAGs of Fig.~\ref{fig_instrumental_class}.} 
 	\label{fig_interm_proof_sec_3nodes} 
 \end{figure}

It is clear that the second mDAG of Fig.~\ref{fig_instrumental_class} observationally dominates the third mDAG of Fig.~\ref{fig_instrumental_class}, because it structurally dominates it (Lemma~\ref{prop_edge_dropping}). Therefore, we only need to show dominance in the other way to prove equivalence; this proof will be explained using the pDAGs of Fig.~\ref{fig_interm_proof_sec_3nodes}. Note that the first pDAG of Fig.~\ref{fig_interm_proof_sec_3nodes} corresponds to the second mDAG of Fig.~\ref{fig_instrumental_class}, and the third pDAG of Fig.~\ref{fig_interm_proof_sec_3nodes} corresponds to the third mDAG of Fig.~\ref{fig_instrumental_class}. So, we need to show that the first pDAG is observationally dominated by the third pDAG in Fig.~\ref{fig_interm_proof_sec_3nodes}. 

In the first pDAG of Fig.~\ref{fig_interm_proof_sec_3nodes}, the node $a$ is influenced by the latent common cause $\beta$ that it shares with the node $b$. As such, learning the value of $X_a$ teaches us something about the latent common cause $\beta$ via Bayesian updating. As will be formalized in Lemma~\ref{lemma_dominance_bayesian_updating} of Appendix~\ref{appendix_facesplitting}, this is at most as powerful as a situation where $a$ acts as a setting for $\beta$. This implies that the first pDAG of Fig.~\ref{fig_interm_proof_sec_3nodes} is observationally dominated by the second pDAG of Fig.~\ref{fig_interm_proof_sec_3nodes}.

Via exogenization of latent nodes, the second pDAG of Fig.~\ref{fig_interm_proof_sec_3nodes} is observationally equivalent to the third pDAG of Fig.~\ref{fig_interm_proof_sec_3nodes}. Thus, we conclude that the third mDAG of Fig.~\ref{fig_instrumental_class} observationally dominates the second mDAG of Fig.~\ref{fig_instrumental_class}, which shows that they are observationally equivalent. This is the implementation of Weak Facet-Adding (Proposition~\ref{prop_weakFS}) to show the observational equivalence of these two mDAGs. 
 
In the same fashion, the Structural Dominance rule, the HLP Edge-Adding  Rule and the Weak Facet-Adding rule can be used to show that the partition of the 72 3-node mDAGs consistent with a fixed nodal ordering into the 15 blocks that are depicted in Fig.~\ref{fig_temporally_ordered_3_nodes} is a proven-equivalence partition, that is, all of the mDAGs inside the same set are observationally equivalent to each other. Moreover, these rules can show that all of the lines that connect one set to another are proven observational dominances. 

It is worth noting that even the Evans' Proposition (Proposition~\ref{Evans}), which is subsumed by HLP+Weak Facet-Adding (Lemma \ref{lemma_WeakFS_and_HLP_Evans}), is sufficient to show all of the observational equivalences and dominances of Fig.~\ref{fig_temporally_ordered_3_nodes}.

What is left to be justified is that no observational dominance between 3-node mDAGs consistent with a fixed nodal ordering is missing from Fig.~\ref{fig_temporally_ordered_3_nodes}\footnote{In Figure 13 of Ref.~\cite{evans_graphs_2016}, representatives of 8 blocks of a proven-equivalence partition of 3-node mDAGs obtained by Evans's Rule were shown. That proven-equivalence partition is different than the one we present here for two reasons: first, the organization of Ref.~\cite{evans_graphs_2016} is not focused only on mDAGs consistent with a fixed nodal ordering. Second, that organization only counts mDAGs up to permutation of nodes.}.

\subsection{Observational Nondominances}

As we will show now, for the case of 3-node mDAGs the proven-dominance partial order established in the previous section corresponds to the observational partial order. That is, every time that one 3-node mDAG $\mathfrak{G}$ cannot be shown to observationally dominate another 3-node mDAG $\mathfrak{G}'$ by the conjunction of the Structural Dominance rule, the HLP Edge-Adding  Rule and the Weak Facet-Adding rule, then it is possible to prove that  $\mathfrak{G}$ \emph{does not} observationally dominate  $\mathfrak{G}'$.

We will do so by applying the nondominance-proving rules of Fig.~\ref{Inequivalence_Rules} starting from the computationally cheaper rules and progressing to more expensive ones. Therefore, the first rule we apply is the comparison of skeletons. 

\begin{figure*}[h!]
	\centering
	\includegraphics[width=0.9\textwidth]{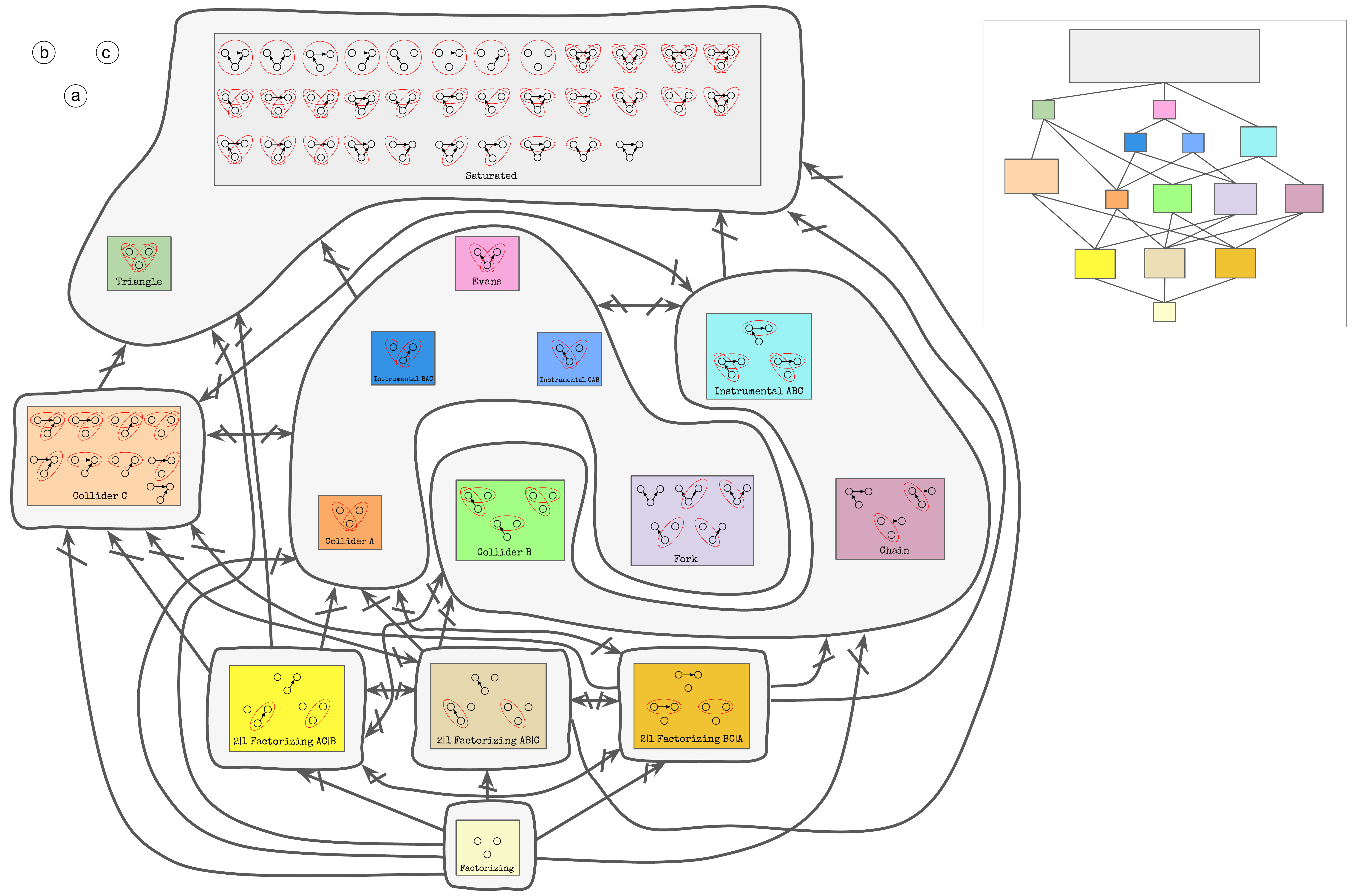}
	\caption{Representation of the information obtained by applying the comparison of skeletons to the proven-equivalence partition of 3-node mDAGs. In this figure, the loops represent blocks of the proven-inequivalence partition and the cut arrows ($\nrightarrow$) represent proven nondominances. Double-headed cut arrows mean that there is proven nondominance in both directions, i.e., proven incomparability. Compare this to the set of observational \emph{dominances} proven by Structural Dominance, the HLP Edge-Adding  rule and Weak Facet-Adding, reproduced in the miniature on the top right.}
	\label{fig_3node_skeletons}
\end{figure*}

The blocks of the proven-inequivalence partition induced by the comparison of skeletons rule are depicted by the loops of Fig.~\ref{fig_3node_skeletons}. Each cut arrow ($\nrightarrow$) between two blocks indicates a proven nondominance relation between the mDAGs of those blocks. For example, because the skeleton of {\tt Factorizing AC|B} is a subgraph of the skeleton of {\tt Collider C},the former cannot observationally dominate the latter. The skeleton of {\tt Factorizing AC|B} is also a subgraph of the skeleton that is common to {\tt Evans}, {\tt Instrumental BAC}, {\tt Instrumental CAB}, {\tt Collider A}, and {\tt Fork}. This implies observational nondominances {\tt Factorizing AC|B} over each of these. In  Fig.~\ref{fig_3node_skeletons} we represent all of these proven nondominances by one single cut arrow from {\tt Factorizing AC|B} to the block of the proven-inequivalence partition induced by comparison of skeletons that contains {\tt Evans}, {\tt Instrumental BAC}, {\tt Instrumental CAB}, {\tt Collider A}, and {\tt Fork}. Because nondominance is not a transitive property, we must explicitly represent \emph{all} of the proven nondominances in Fig.~\ref{fig_3node_skeletons}. For example, there is a cut arrow into the {\tt Saturated} block from every other block of the proven-inequivalence partition.

As we can see in Fig.~\ref{fig_3node_skeletons}, the comparison of skeletons is sufficient to show that the following blocks of the proven-equivalence partition are observational equivalence classes: {\tt Factorizing}, {\tt Factorizing AC|B}, {\tt Factorizing AB|C}, {\tt Factorizing B|CA} and {\tt Collider C}. Furthermore, it shows that in all of the instances where an observational dominance involving one of these five classes could not be proven by the conjunction of the structural dominance rule, the HLP Edge-Adding rule and the Weak Facet-Adding rule, then there is a proven observational nondominance. Therefore, the part of the observational partial order concerning these five classes is solved by the comparison of skeletons. 

Next, we proceed to applying the comparison of d-separation relations to the remaining mDAGs. The proven-equivalence blocks {\tt Saturated}, {\tt Triangle}, {\tt Evans}, {\tt Instrumental BAC}, {\tt Instrumental CAB} and {\tt Instrumental ABC} do not present any d-separation relations. On the other hand, {\tt Collider A} presents the d-separation relation $b\dsep c$, {\tt Collider B} presents the d-separation relation $a\dsep c$, {\tt Fork} presents the d-separation relation $b\dsep c|a$ and {\tt Chain} presents the d-separation relation $a\dsep c|b$.

\begin{figure*}[h!]
	\centering
	\includegraphics[width=0.9\textwidth]{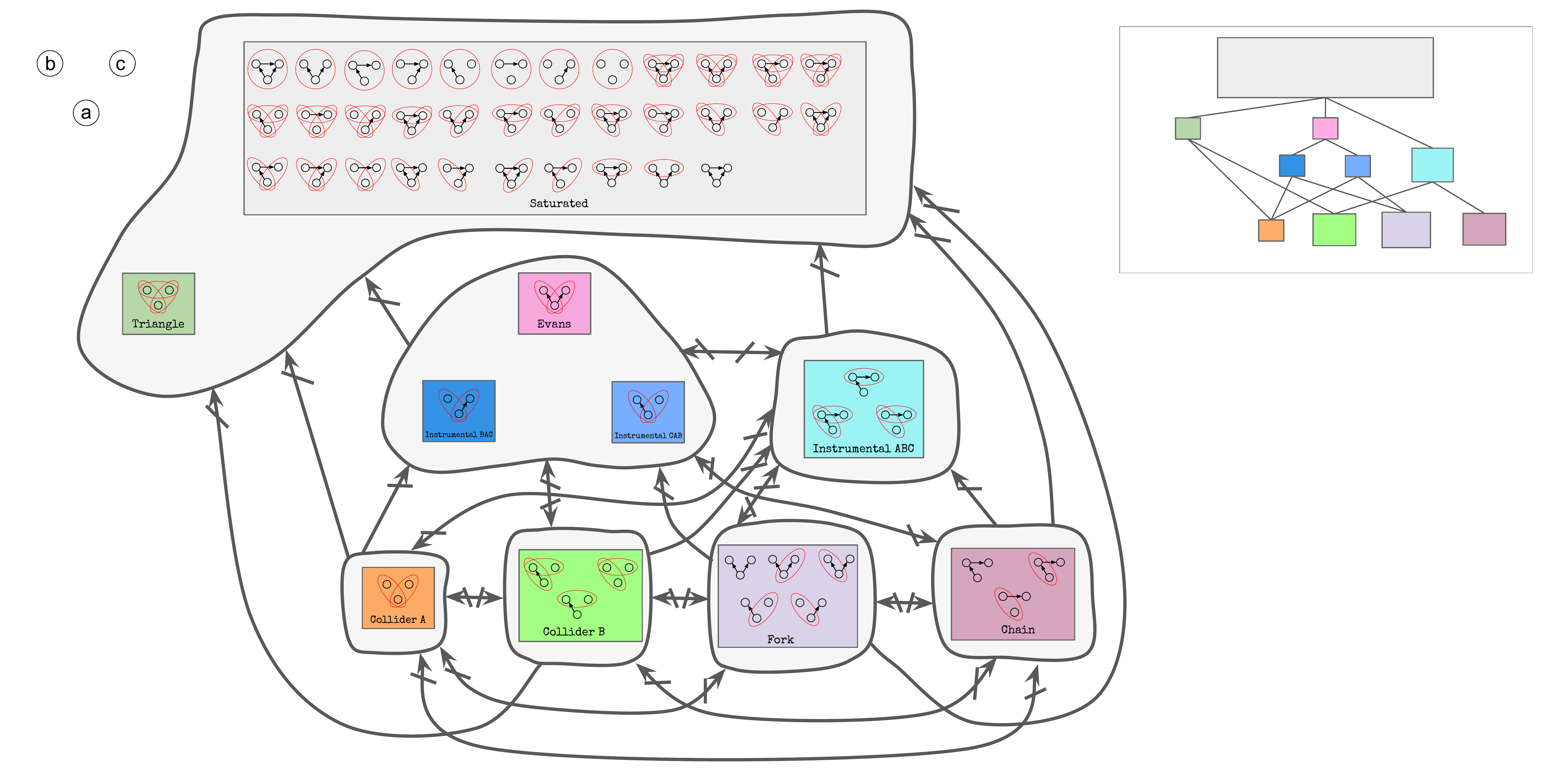}
	\caption{Representation of the information obtained by applying the comparison of skeletons and the comparison of d-separation relations to the proven-equivalence partition of 3-node mDAGs. Here, we only show the blocks of the proven-equivalence partition that were not already shown to be observational equivalence classes by the comparison of skeletons alone. The loops represent blocks of the proven-inequivalence partition and the cut arrows ($\nrightarrow$) represent proven nondominances. Double-headed cut arrows mean that there is proven nondominance on both directions, i.e., proven incomparability. Compare this to the set of observational \emph{dominances} among these mDAGs proven by Structural Dominance, the HLP Edge-Adding  rule and Weak Facet-Adding, reproduced in the miniature on the top right.}
	\label{fig_3node_dsep}
\end{figure*}

From this, we are able to show more nondominances, and in particular split some of the blocks of the proven-inequivalence partition induced by comparison of skeletons. When we present the resulting nondominances in Fig.~\ref{fig_3node_dsep}, to make it easier to visualize, we do not include the classes that were already completely classified with the comparison of skeletons alone (i.e.,  {\tt Factorizing}, {\tt Factorizing AC|B}, {\tt Factorizing AB|C}, {\tt Factorizing B|CA} and {\tt Collider C}).

As we can see in Fig.~\ref{fig_3node_dsep}, by applying the comparison of d-separation relations to the blocks of the proven-inequivalence partition induced by the comparison of skeletons, we further learn that {\tt Instrumental ABC}, {\tt Collider A}, {\tt Collider B}, {\tt Fork} and {\tt Chain} are observational equivalence classes. Furthermore, we learn that in all of the instances where an observational dominance involving one of these five classes could not be proven by the conjunction of the structural dominance rule, the HLP Edge-Adding rule and the Weak Facet-Adding rule, then there is a proven observational nondominance. Therefore, the part of the observational partial order concerning these five classes is solved by comparison of skeletons together with comparison of d-separation relations.

Next, we attempt to apply the comparison of e-separation relations and the comparison of densely connected pairs to the remaining mDAGs. This, however, does not show any nondominance that was not already known. Therefore, we move to applying the directed-edge-free rule, and this shows one extra nondominance: {\tt Triangle} does not dominate {\tt Saturated}. With this, we learn that  {\tt Triangle} is an observational equivalence class. 

At this point, we have the proven-inequivalence partition induced by all nondominance rules except for comparison of unrealizable supports. It is this last rule that establishes the remaining nondominances embodied in the observational partial order described in Fig.~\ref{fig_temporally_ordered_3_nodes}.

We start by  applying the comparison of unrealizable supports on binary variables constituted of two events.  Consider the set constituted by the events 
	\begin{align*}
	& \{X_a=0,X_b=0,X_c=0\}, \\
	&\{X_a=1,X_b=1,X_c=1\}. \\
\end{align*}

As a support, this set expresses perfect correlation between $X_a$, $X_b$ and $X_c$. This support is realizable by all of {\tt Evans}, {\tt Instrumental ABC}, {\tt Instrumental BAC}, {\tt Instrumental CAB}, {\tt Fork} and {\tt Chain}, but it is \emph{not} realizable by {\tt Triangle}. Therefore, {\tt Triangle} does not observationally dominate the aforementioned proven-equivalence blocks. With this, we learn that the observational dominances involving {\tt Triangle} that can be proven using Structural Dominance,  HLP Edge-Adding  and Weak Facet-Adding are indeed all of the observational dominance relations involving {\tt Triangle}.

When we apply  the comparison of unrealizable supports on binary variables constituted of three events, we can show more nondominances.  Let $S$ be the set constituted by the events 
	\begin{align*}
	& \{X_a=0,X_b=0,X_c=0\}, \\
	&\{X_a=0,X_b=1,X_c=1\}, \\
	& \{X_a=1,X_b=0,X_c=0\}.
\end{align*}

The set $S$ is a support realizable by {\tt Evans}, but unrealizable by  {\tt Instrumental BAC} and {\tt Instrumental CAB}. This was found by Fraser's algorithm, but here we give a direct proof. First, an explicit choice of parameters can show that there are distributions with support $S$ that are realizable by {\tt Evans}. Say that the common cause between $a$ and $c$ is $\lambda$, and the common cause between $a$ and $b$ is $\gamma$. Take each of $X_\lambda$ and $X_\gamma$ to be binary and take the distribution over them to be full support. Then, let $X_a = X_\lambda \oplus X_\gamma$, where $\oplus$ means sum modulus 2. Furthermore, let $X_b=X_\lambda \cdot(X_a\oplus 1)$ and $X_c=X_\gamma \cdot(X_a\oplus 1)$. From these functions, we have:
\begin{align*}
	&(X_\lambda, X_\gamma)=(0,0) \Rightarrow X_a=0, X_b=0, X_c=0 \\
	&(X_\lambda, X_\gamma)=(1,1) \Rightarrow X_a=0, X_b=1, X_c=1 \\
	&(X_\lambda, X_\gamma)=(0,1) \Rightarrow X_a=1, X_b=0, X_c=0 \\
	& (X_\lambda, X_\gamma)=(1,0) \Rightarrow X_a=1, X_b=0, X_c=0 .
\end{align*}

That is, choices of parameters of this type give rise to distributions that have support $S$.

Note that this construction is possible because in {\tt Evans} both $b$ and $c$ have direct access to $a$: they need to know when to copy the hidden variable and when to set their value to $0$. Therefore, this specific construction is not possible in {\tt Instrumental BAC} or {\tt Instrumental CAB}. More than that, any distribution with support $S$ violates the two versions of Pearl's instrumental inequalities \cite{Pearl_Instrumental} that are respectively associated with {\tt Instrumental BAC} and {\tt Instrumental CAB}, and thus is not realizable by these mDAGs. Therefore, this explicitly shows us that {\tt Instrumental BAC} and {\tt Instrumental CAB} do not dominate {\tt Evans}. We infer that {\tt Evans} is an observational equivalence class.

Finally, we apply  the comparison of unrealizable supports on binary variables constituted of four events. Let $S'$ be the set constituted by the events
\begin{align*}
	& \{X_a=1,X_b=0,X_c=0\}, \\
	&\{X_a=0,X_b=0,X_c=1\}, \\
	& \{X_a=0,X_b=1,X_c=1\}, \\
	& \{X_a=0,X_b=0,X_c=0\}.
\end{align*}

The set $S'$ is a support realizable by {\tt Instrumental CAB}, but unrealizable by  {\tt Instrumental BAC}. This was found by Fraser's algorithm, but here we give a direct proof. First, an explicit choice of parameters can show that there are distributions with support $S'$ that are realizable by {\tt Instrumental CAB}. Let the common cause between $a$ and $b$ be $\lambda$, and the common cause between $a$ and $c$ be $\gamma$. Take each of $X_\lambda$ and $X_\gamma$ to be binary and take the distribution over them to be full support. Then, let $X_a=X_\lambda \cdot X_\gamma$, $X_b=X_\lambda \cdot (X_a \oplus 1)$ and  $X_c= X_\gamma\oplus 1$. Choices of parameters of this type give rise to distributions that have support $S'$. On the other hand, it is not hard to see that the any distribution with support $S'$ violates the version of Pearl's instrumental inequalities~\cite{Pearl_Instrumental} that are associated with {\tt Instrumental BAC}. To find a support that is realizable by  {\tt Instrumental BAC} but unrealizable by  {\tt Instrumental CAB}, we only need to invert $b$ and $c$ in $S'$. 

Together, all of the arguments presented above close the problem of finding the observational partial order of observational equivalence classes of 3-node mDAGs consistent with a fixed nodal ordering, showing that it is indeed the one shown in Fig.~\ref{fig_temporally_ordered_3_nodes}.

Tables~\ref{table_3vis_equivs} and~\ref{table_3node_inequivalences} give a selective summary of the steps shown in this section: Table~\ref{table_3vis_equivs} presents the number of elements of the proven-equivalence partition of 3-node mDAGs consistent with a fixed nodal ordering induced by different combinations of the dominance-proving rules, and Table~\ref{table_3node_inequivalences} presents the number of elements of the proven-inequivalence partition of 3-node mDAGs consistent with a fixed nodal ordering induced by different combinations of the nondominance-proving rules.

\begin{table*}[]
	\centering
	\begin{tabular}{c c }
		Methods Applied &   Cardinality of the induced proven-equivalence partition   \\
		\hline
		None & 72  \\
		SD + HLP &  44  \\
		SD + HLP + Weak FA  &    15   
	\end{tabular}
	\captionof{table}{Number of blocks of the proven-equivalence partition for 3-node mDAGs consistent with a fixed nodal ordering obtained by applying various combinations of  the dominance-proving rules of Section~\ref{sec_show_equivalence}. The row ``None'' gives the total number of 3-node mDAGs consistent with a fixed nodal ordering. These numbers give upper bounds on the number of observational equivalence classes.}\label{table_3vis_equivs}
\end{table*}

\begin{table*}[h!]
	\centering
	\begin{tabular}{c c }
		Methods Applied  &  Cardinality of the induced proven-inequivalence partition    \\
		\hline
		skel         &    8           \\
		d-sep+skel  &         12     \\
		e-sep & 12  \\
		DC	+ e-sep & 12    \\
		DEF + DC + e-sep  &     13         \\ 
	DEF + DC + e-sep  + Supps up to 2 events & 13 \\
		DEF + DC + e-sep  + Supps up to 3 events & 14 \\
		DEF + DC + e-sep + Supps up to 4 events & 15
	\end{tabular}
	\captionof{table}{Number of blocks of the proven-inequivalence partition for 3-node mDAGs consistent with a fixed nodal ordering obtained by applying various combinations of  the nondominance-proving rules of Section~\ref{sec_show_inequivalence}. These numbers give lower bounds on the final number of equivalence classes. The fact that the last line reaches 15, which is the upper bound obtained from the proven-equivalence partition, shows that we have identified the observational equivalence partition for 3-node mDAGs consistent with a fixed nodal ordering.}
	\label{table_3node_inequivalences}
\end{table*}

\section{Partial Solution of Observational Partial Order for 4-node mDAGs}
\label{sec_4}

Based on the case of three visible nodes, one might wonder whether Structural Dominance, HLP Edge-Adding  and Weak Facet-Adding, that is, the dominance-proving rules that were known prior to this work, are sufficient to find the observational partial order for any number of visible nodes. However, the case of four visible nodes shows that this expectation is not borne out: Table \ref{table_4vis_equivs} shows that the number of elements of the proven-equivalence partition, which is an upper bound on the number of observational equivalence classes, gets smaller upon successive applications of the equivalence rules of Section~\ref{sec_show_equivalence}. Thus, our two extensions (Moderate and Strong Facet-Adding) are indeed necessary to establish certain dominances and equivalences.

\begin{table*}[]
\centering
\begin{tabular}{c c }
	Methods Applied &   Cardinality of the induced proven-equivalence partition   \\
	\hline
	None & 7296  \\
	SD + HLP &  4417  \\
	SD + HLP + Weak FA  &    1481    \\
	SD + HLP + Moderate FA & 1466  \\
	SD + HLP + Strong FA  & 1444
\end{tabular}
\captionof{table}{Cardinalities of the proven-equivalence partitions induced by various combinations of  the dominance-proving rules of Section~\ref{sec_show_equivalence} for the set of 4-node mDAGs consistent with a fixed nodal ordering. The row ``None'' gives the total number of 4-node mDAGs consistent with a fixed nodal ordering. These cardinalities give successively tighter upper bounds on the number of observational equivalence classes.}\label{table_4vis_equivs}
\end{table*}

Table~\ref{table_supports} shows the number of elements of the proven-inequivalence partition that is obtained after applying each one of the nondominance-proving rules of Section~\ref{sec_show_inequivalence} to mDAGs consistent with a fixed nodal ordering. In other words, Table~\ref{table_supports} is the analogue of Table~\ref{table_3node_inequivalences} for the case of four visible nodes. As we can see, we have not solved all ambiguity about the observational equivalence partition for 4-node mDAGs consistent with a fixed nodal ordering: the best upper bound on the number of observational equivalence classes that we have obtained is 1444, while the best lower bound is 1256.

\begin{table*}[h!]
\centering
\begin{tabular}{c c c}
	Methods Applied  & Cardinality of the induced proven-inequivalence partition   \\
	\hline
		None & 1  \\
	skel                  &       64 \\
	d-sep+skel      &   259      \\
	e-sep &  326 \\ 
	 DC + e-sep   & 334  \\
		DEF + DC + e-sep       &    350   \\ 
DEF + DC + e-sep  + Supps up to 2 events   & 447 \\
	DEF + DC + e-sep  + Supps up to 3 events  & 595 \\
	DEF + DC + e-sep  + Supps up to 4 events  & 1054 \\
	DEF + DC + e-sep  + Supps up to 5 events    & 1153  \\
	DEF + DC + e-sep  + Supps up to 6 events    &  1243  \\
	DEF + DC + e-sep  + Supps up to 7 events    &  1253 \\
	DEF + DC + e-sep  + Supps up to 8 events    &   1253
\end{tabular}
\captionof{table}{Cardinalities of the proven-inequivalence partitions induced by various combinations of  the nondominance-proving rules of Section~\ref{sec_show_inequivalence} for the set of 4-node mDAGs consistent with a fixed nodal ordering. These cardinalities give successively tighter lower bounds on the number of observational equivalence classes. All the support comparison applied assumed all the visible variables to be binary, i.e., $\vec c_\text{vis}=(2,2,2,2)$.}
\label{table_supports}
\end{table*}

It is still possible that the reason  we did not solve the problem is that we did not use the full power of the comparison of unrealizable supports: we only compared supports for binary variables and only up to 8 events. In Section \ref{sec_every_support}, we will study one specific block of the proven-inequivalence partition, and show that there are still cases where whether observational equivalence holds or not is unknown even after making all of the possible comparisons of unrealizable supports while assuming that all visible variables are binary, i.e., for $\vec c_\text{vis}=(2,2,2,2)$. We will then discuss the possibility of comparing unrealizable supports under the assumption that one or more of the visible variables have higher cardinality, even though this becomes increasingly impractical and computationally expensive.

Before doing so, we will look into the set of observational equivalence classes that were completely characterized by our analysis.

\subsection{Completely Identified Classes}
\label{sec_completely_solved}

When a block of the proven-equivalence partition is also a block of the proven-inequivalence partition (a pink loop coincides with a gray loop in Fig.~\ref{fig_veinn}), we infer that it is a block of the observational equivalence partition (green dashed line in Fig.~\ref{fig_veinn}). In that case, we say that the observational equivalence class is ``identified''. 

For the case of 3-node mDAGs consistent with a fixed nodal ordering, as discussed, \emph{all} of the classes are identified when we apply the graphical rules of Section~\ref{sec_graphical} together with the comparison of unrealizable supports assuming visible variables that are all binary and up to 4 events. For the case of four visible nodes, on the other hand, even the comparison of unrealizable supports for binary variables up to 8 events cannot identify all classes. However, some equivalence classes \emph{can} be identified. Table \ref{table_solved} presents the number of  classes that are identified by various combinations of nondominance-proving rules.

\begin{table*}[]
	\centering
	\begin{tabular}{c c }
		Methods Applied & Nº of Identified Observational Equivalence Classes  \\
		\hline
		None & 0 \\
		skel                   &       15 \\
		d-sep & 114 \\
		d-sep+skel       &   152      \\
		e-sep & 174 \\
		 DC + e-sep    &  186  \\
		DEF + DC + e-sep    &    218   \\
		DEF + DC + e-sep   + Supps up to 2 events & 298 \\
		DEF + DC + e-sep   + Supps up to 3 events &  378 \\
		DEF + DC + e-sep   + Supps up to 4 events &  859  \\
		DEF + DC + e-sep   + Supps up to 5 events & 990  \\
		DEF + DC + e-sep   + Supps up to 6 events &  1136 \\
		DEF + DC + e-sep   + Supps up to 7 events &  1156 \\
		DEF + DC + e-sep   + Supps up to 8 events &  1156
	\end{tabular}
	\captionof{table}{Number of identified classes of 4-node mDAGs consistent with a fixed nodal ordering obtained by applying various combinations of  the nondominance-proving rules of Section~\ref{sec_show_inequivalence}. These are sets of mDAGs that are both blocks of the proven-equivalence partition and of the proven-inequivalence partition. }\label{table_solved}
\end{table*}

\begin{figure}[htbp]
	\centering
	\includegraphics[width=0.47\textwidth]{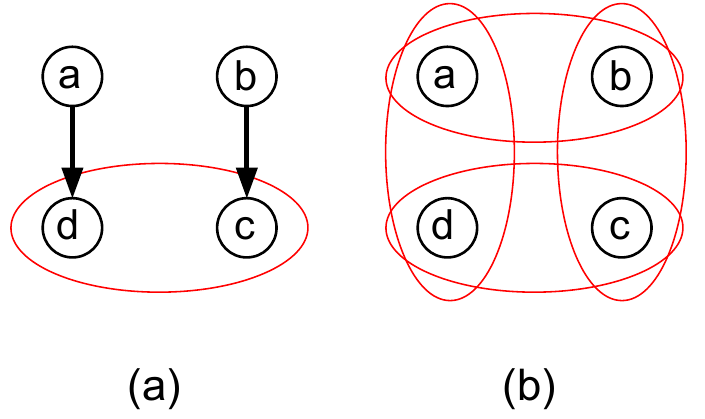}
	\caption{(a) Bell mDAG. (b) {\tt Square} mDAG. Both of these belong to equivalence classes that are completely identified by d-separation, meaning that there is no mDAG that is observationally inequivalent to them while still imposing the same conditional independence constraints on the realizable distributions.} 
	\label{fig_Bell_and_Square}
\end{figure}

In particular, with this analysis we reproduced one of the results of Ref.~\cite{Wood_and_Spekkens}: the observational equivalence class of the Bell mDAG (Fig.~\ref{fig_Bell_and_Square}(a)) consists of the 9 mDAGs whose associated pDAGs are consistent with a fixed nodal ordering depicted in Figure 24 of Ref.~\cite{Wood_and_Spekkens}. The Bell equivalence class is identified by d-separation alone, meaning that the particular pattern of d-separation relations presented by this observational equivalence class is not presented by any other class. In other words, there is \emph{no} mDAG that possesses the same pattern of d-separation relations as the Bell mDAG while being inequivalent to it.

The main point of Ref.~\cite{Wood_and_Spekkens} is that if one has a probability distribution that violates Bell inequalities (which are the inequality constraints imposed by the Bell mDAG) but satisfies the no-signalling relations (which are the conditional independence constraints imposed by the Bell mDAG), then it is only possible to find a causal explanation for this distribution in the framework of \emph{classical} causal models if one allows for \emph{fine-tuning}. Fine-tuning, which is another name for the notion of \emph{unfaithfulness}~\cite{Spirtes2000}, is the property that a probability distribution exhibits conditional independence relations that are not entailed by the d-separation relations of the causal structure that realized this distribution. Those extra conditional independence relations arise due to the specific choice of parameters that is used to realize the distribution in question, and not due to the structure of the mDAG. As it turns out, the conclusion that was drawn in  Ref.~\cite{Wood_and_Spekkens} for the Bell causal structure can be extended to any observational equivalence class of causal structures that is \nonalgebraic and is identified by d-separation alone. For such a class of causal structures, any distribution that satisfies all and only the conditional independence relations implied by the set of d-separation relations of that class, but that violates the inequality constraints of the class necessarily cannot be explained within the framework of classical causal models without fine-tuning.

Fig.~\ref{fig_Bell_and_Square}(b) shows an example of a causal structure (distinct from the Bell causal structure) whose equivalence class is identified by d-separation alone: the {\tt Square} mDAG. The argument above says that any distribution whose set of conditional independence relations is exactly $X_a \dbot X_c$ and $X_d \dbot X_b$ but that violates the inequality constraints implied by the {\tt Square} mDAG does not admit of any non-fine-tuned causal explanation in the framework of classical causal models.

Now, we go back to one of our motivations discussed in the \hyperref[introductionsection]{Introduction}: the question of the significance of analyzing inequality constraints for causal discovery.

\subsection{Importance of Inequality Constraints for Causal Discovery}

\subsubsection{Nonalgebraicness is Ubiquitous}

As discussed in Section~\ref{sec_3_observed_nodes}, in the set of 3-node mDAGs consistent with a fixed nodal ordering, there are 15 observational equivalence classes and 5 of them are \nonalgebraic. Therefore, for the case of three visible variables, the fraction of \nonalgebraic classes among all classes of mDAGs consistent with a fixed nodal ordering is $1/3$. As we will see in this section, the fraction of \nonalgebraic classes gets much bigger in the case of 4-node mDAGs consistent with a fixed nodal ordering.

Since we still have not obtained a complete characterization of the observational equivalence partition of 4-node mDAGs consistent with a fixed nodal ordering, the best we can do is to find a lower bound on the fraction of \nonalgebraic equivalence classes. We will explain two methods for obtaining such a lower bound. We then pick the best lower bound out of the two results. Method 1 consists in obtaining a \emph{lower bound on the number of \nonalgebraic classes}, and dividing it by our best \emph{upper bound} on the total number of observational equivalence classes (which is currently 1444). The resulting number $f_1$ is a lower bound on the fraction of \nonalgebraic classes. Method 2 consists in obtaining an \emph{upper bound on the number of \algebraic classes}, and dividing it by our best \emph{lower bound} on the total number of observational equivalence classes (which is 1253). The resulting number $f_2$ is such that $1-f_2$ is a lower bound on the fraction of \nonalgebraic classes.

We first implement method 1. An observational equivalence class of mDAGs consistent with a fixed nodal ordering is identified as \nonalgebraic, according to Proposition~\ref{prop_temporal_algebraic}, if its mDAGs are not observationally equivalent to \emph{any} latent-confounder-free mDAG, including ones that are not  consistent with the fixed nodal ordering.  We thus look at the proven-inequivalence partition and, using the nondominance-proving rules, find all of the blocks thereof whose mDAGs are proven to be inequivalent to every latent-confounder-free mDAG (including ones that are not consistent with the chosen nodal ordering).  The number of blocks of this type is a lower bound on the number of \nonalgebraic observational equivalence classes. Applying this method on the proven-inequivalence partition induced by all of the graphical rules of Section~\ref{sec_graphical} together with comparison of unrealizable supports up to 8 events, we obtain the lower bound of 1137 on the number of \nonalgebraic observational equivalence classes. Therefore, the lower bound on the fraction of \nonalgebraic classes obtained by method 1 is $f_1=1137/1444\approx 78.7\%$.

Now we implement method 2, where we want to find an \emph{upper bound} on the number of \algebraic classes. Again according to Proposition~\ref{prop_temporal_algebraic}, an observational equivalence class of mDAGs consistent with a fixed nodal ordering is \algebraic only if its mDAGs are observationally equivalent to \emph{some} latent-confounder-free mDAG (not necessarily one that is consistent with the fixed nodal ordering).   This implies that the \emph{total number} of observationally inequivalent latent-confounder-free mDAGs (not restricting to a set that is consistent with a fixed nodal ordering) gives us an upper bound on the number of \algebraic observational equivalence classes in the set of mDAGs that are consistent with a fixed nodal ordering. It is easy to find this number, since the observational profile of latent-confounder-free mDAGs is completely characterized by d-separation. Doing this, we find that the number of \algebraic classes of 4-node mDAGs in the set of mDAGs that are consistent with a fixed nodal ordering is \emph{at most} $185$. To obtain the maximum fraction of \algebraic mDAGs consistent with a fixed nodal ordering, we divide this number by our best lower bound on the total number of equivalence classes, thus obtaining $f_2=185/1253 \approx 14.8\%$. Therefore, the lower bound on the fraction of \nonalgebraic classes obtained by method 2 is $1-f_2\approx 85.2\%$. The lower bound obtained by method 1 is worse than the lower bound obtained by method 2, so we will use the latter.

To conclude, the rate of incidence of inequality constraints among observational equivalence classes of 4-node mDAGs consistent with a fixed nodal ordering is \emph{at least} $85.2\%$. This high rate and the fact that it increases with the number of nodes suggests that inequality constraints are not rare; on the contrary, they seem to be generic.  This is of special relevance to quantum physicists interested in causal inference, as it shows that there is much potential for finding new observational equivalence classes of causal structures that exhibit  a quantum-classical gap. 

\subsubsection{ Singling out equivalence classes under the assumption of faithfulness via equality and inequality constraints }

As discussed in Section~\ref{significanceforcausaldiscovery}, without assuming any model-selection principles, the most we can learn about the underlying causal structure starting from one observational distribution is the upward closure (in the observational partial order) of a given set of observational equivalence classes. In this section, we discuss the case where we assume faithfulness to get a finer-grained estimate of the underlying causal structure, and consider the importance of inequality constraints for causal discovery in this case.

Under the assumption of faithfulness, we only consider candidate causal structures that impose all and only the conditional independence relations that appear in our observed probability distribution. In general, this does not single out only one observational equivalence class: there could be multiple equivalence classes whose set of  conditional independence constraints correspond exactly to the relations that appear in our data, but that have different sets of nested Markov constraints or inequality constraints. In this case, we would have to test our distribution against these extra constraints (and our distribution could turn out to be realizable by more than one of these classes). 

However, there are special cases where one \emph{can} single out one observational equivalence class under the assumption of faithfulness without having to check nested Markov constraints or inequality constraints: when the exact set of d-separation relations of that observational equivalence class does not appear in any other observational equivalence class.  When this happens, using the language of Section~\ref{sec_completely_solved}, the observational equivalence class is identified by the comparison of d-separation relations. We will say that such classes are \emph{identified by conditional independence relations alone}; in Fig.~\ref{fig_Bell_and_Square}, we gave examples of two mDAGs, Bell and Square, that belong to equivalence classes of this type.  If our observed probability distribution satisfies all and only the conditional independence relations that are imposed by an equivalence class that is identified by conditional independence relations alone, then that class provides the only possible faithful causal explanation to our observed data.  If the conditional independence relations of the set of data do not correspond to those of any equivalence class that is identified by conditional independence relations alone, then there is remaining ambiguity about the observational equivalence class that explains the data, even when assuming faithfulness.

For the case of 2-node mDAGs there are no inequality constraints, so every observational equivalence class can be identified by conditional independence relations alone. For the case of 3-node mDAGs consistent with a fixed nodal ordering, there are  9  equivalence classes that can be identified by d-separation alone. The remaining classes, namely,
  {\tt Triangle}, {\tt Evans}, {\tt Instrumental ABC},  {\tt Instrumental BAC}, {\tt Instrumental CAB} and {\tt Saturated}, are all devoid of nontrivial d-separation relations.  Therefore, for this case the fraction of causal structures that can be singled out by the analysis of conditional independence constraints alone is $3/5$ . For the case of 4-node mDAGs consistent with a fixed nodal ordering, Table \ref{table_solved} shows that only 114 equivalence classes are identified by conditional independence relations alone. 
  This is less than 10\% of our best lower bound on the total number of equivalence classes. Therefore, in most cases it will \emph{not} be possible to discriminate between observationally inequivalent classes of causal structures if one only avails oneself of conditional independence relations.

Therefore, our analysis shows that as well as the fraction of \nonalgebraic classes increasing
 with the number of nodes for the cases of 2, 3 and 4 nodes, the fraction of all observational equivalence classes that can be singled out by analysis of conditional independence relations alone drastically decreases with the number of nodes for the cases of 2, 3 and 4 nodes.

When the set of conditional independence relations satisfied by our data corresponds to the set of d-separation relations of a class that is identified by conditional independence relations alone, we might still want to check the data against the inequality constraints of the class. If the data does not satisfy these inequality constraints, then, as already discussed in Section~\ref{sec_completely_solved}, there are no classical causal structures that can faithfully generate this data.

\subsection{mDAGs that can Realize Every Support on Binary Variables}
\label{sec_every_support}

\begin{figure}[htbp]
	\centering
	\includegraphics[width=0.49\textwidth]{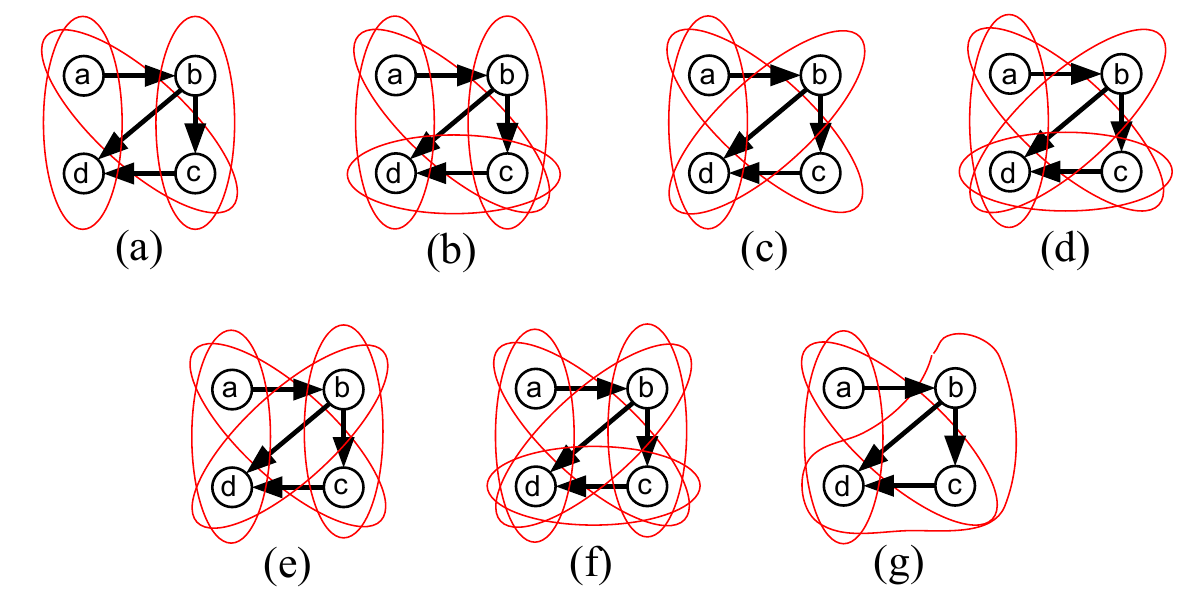}
	\caption{Seven mDAGs that do not appear in the same block as the saturated mDAGs in the proven-equivalence partition induced by the rules described in Section~\ref{sec_show_equivalence}, but that nevertheless can realize any support on binary variables. mDAGs (a)-(d) are shown to \emph{not} be saturated via analysis of supports on variables of higher cardinality. It is still unknown whether mDAGs (e)-(g) are observationally inequivalent to the saturated mDAGs.}
	\label{fig_compatible_any_support}
\end{figure}

In this section, we will discuss a particular set of 4-node mDAGs consistent with a fixed nodal ordering: those that have no unrealizable supports on binary variables, that is, no unrealizable supports when $\vec c_\text{vis}=(2,2,2,2)$. In other words, an mDAG $\mathfrak{G}$ of this set, whose canonical pDAG is $\cal G$, is such that $\mathcal{S}(\mathcal{G},\vec c_\text{vis}=(2,2,2,2))$ includes all possible sets of events over four binary variables. This set of mDAGs includes all of the mDAGs that are proved to be saturated (i.e., the saturated proven-equivalence block), which consists of 2134 mDAGs, together with the extra 7 mDAGs depicted in Fig.~\ref{fig_compatible_any_support}. Each one of these 7 mDAGs belongs to a different block of the proven-equivalence partition, and each of these blocks includes just one mDAG.

As it turns out, this set of mDAGs is a block of the proven-inequivalence partition induced by the graphical rules of Section~\ref{sec_graphical} together with the comparison of unrealizable supports on $\vec c_\text{vis}=(2,2,2,2)$ that have up to 4 events. As one can check, comparing unrealizable supports on  $\vec c_\text{vis}=(2,2,2,2)$ that have larger numbers of events does not help, as these mDAGs do not have any unrealizable supports on  $\vec c_\text{vis}=(2,2,2,2)$  even up to the maximum number of $2^{4}=16$ events.

If we go beyond $\vec c_\text{vis}=(2,2,2,2)$, we can show that the mDAGs of Fig.~\ref{fig_compatible_any_support}(a)-(d) are \emph{not} saturated. Specifically, for $\vec c_\text{vis}=(3,2,2,2)$ they cannot realize the support constituted by the following events: 
\begin{align*}
		& \{X_a=0,X_b=0,X_c=0,X_d=0\}, \\
		& \{X_a=0,X_b=0,X_c=1,X_d=0\}, \\
		& \{X_a=0,X_b=1,X_c=0,X_d=0\}, \\
		& \{X_a=1,X_b=0,X_c=0,X_d=0\}, \\
		& \{X_a=1,X_b=1,X_c=0,X_d=0\}, \\
		& \{X_a=2,X_b=0,X_c=0,X_d=1\}, \\
		& \{X_a=2,X_b=1,X_c=1,X_d=0\}. 
\end{align*}

This was already noted in Eq. (18) of Ref.~\cite{Khanna_2023}. In fact, the mDAGs of Fig.~\ref{fig_compatible_any_support}(a)-(d) correspond to Table II of Ref.~\cite{Khanna_2023}.

It follows that the mDAGs of Fig.~\ref{fig_compatible_any_support}(a)-(d) are not saturated. Furthermore, by the comparison of unrealizable supports, we can also see that the support on  $\vec c_\text{vis}=(3,2,2,2)$ constituted by the following events  cannot be realized by Fig.~\ref{fig_compatible_any_support}(c) and Fig.~\ref{fig_compatible_any_support}(d), but can be realized by Fig.~\ref{fig_compatible_any_support}(a) and Fig.~\ref{fig_compatible_any_support}(b):
\begin{align*}
		& \{X_a=0,X_b=0,X_c=0,X_d=0\}, \\
		& \{X_a=0,X_b=0,X_c=1,X_d=0\}, \\
		& \{X_a=0,X_b=1,X_c=0,X_d=0\}, \\
		& \{X_a=1,X_b=0,X_c=0,X_d=1\}, \\
		& \{X_a=1,X_b=1,X_c=1,X_d=0\}, \\
		& \{X_a=2,X_b=0,X_c=1,X_d=1\}, \\
		& \{X_a=2,X_b=1,X_c=1,X_d=1\}. 
\end{align*}

\begin{figure*}[h!]
	\centering
	\includegraphics[width=0.9\textwidth]{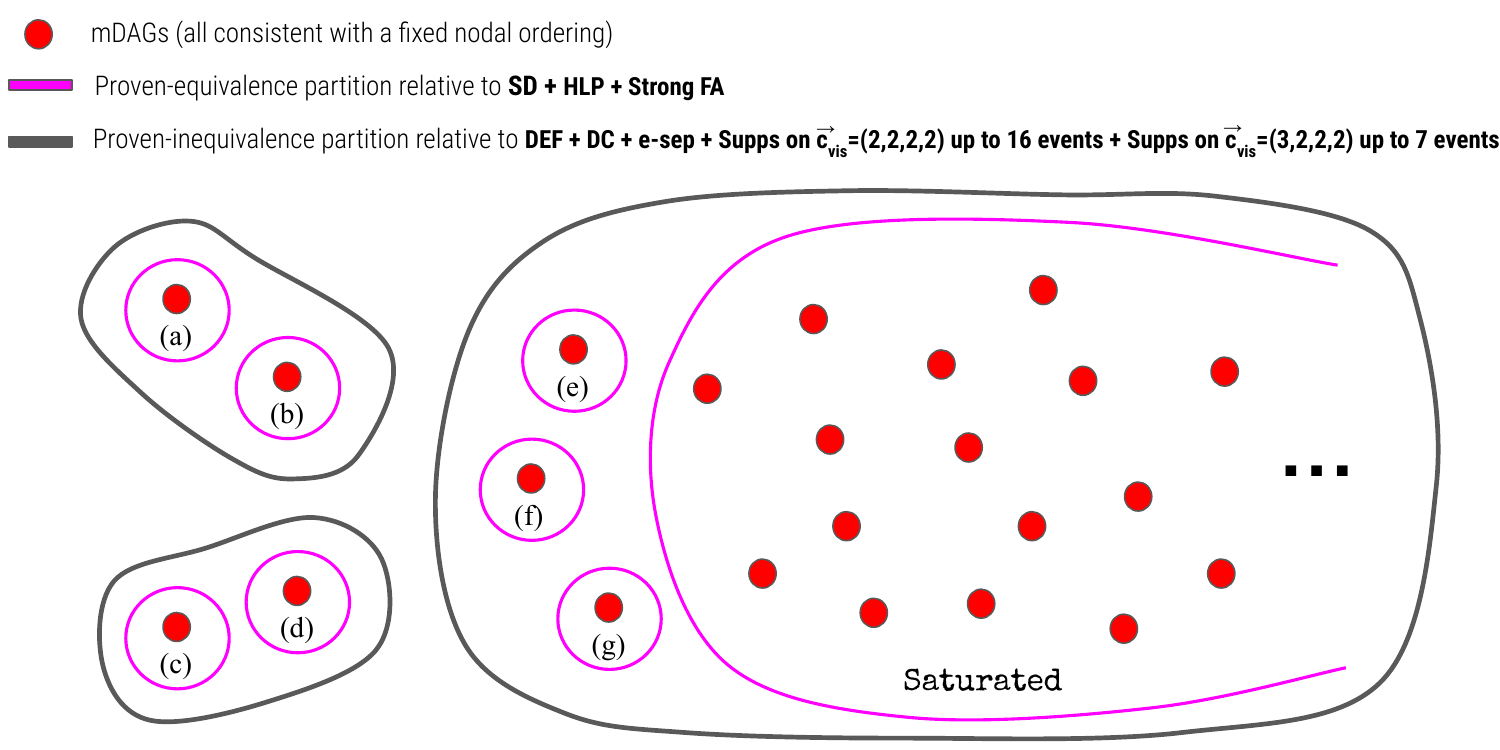}
	\caption{Diagrammatic representation of the proven-equivalence and proven-inequivalence partitions of 4-node mDAGs consistent with a fixed nodal ordering that are induced by the rules indicated in the labels of the figure. The saturated proven-equivalence block includes 2134 mDAGs consistent with a fixed nodal ordering. The letters (a)-(g) in this diagram refer to the corresponding mDAGs of Fig.~\ref{fig_compatible_any_support}.}
	\label{veinn_big_class}
\end{figure*}

Therefore, we learn that the mDAGs of Fig.~\ref{fig_compatible_any_support}(a)-(b) are inequivalent to the mDAGs of Fig.~\ref{fig_compatible_any_support}(c)-(d), which gives us the proven-inequivalence blocks depicted in Fig.~\ref{veinn_big_class}.  This is the most we can learn about inequivalences between these mDAGs by checking supports on $\vec c_\text{vis}=(3,2,2,2)$ up to 7 events.

Regarding mDAGs \ref{fig_compatible_any_support}(e)-\ref{fig_compatible_any_support}(g), there are three possibilities. The first one is that, if we go to even higher cardinality of the visible variables, we can prove inequivalence by comparison of unrealizable supports. This option would only use theoretical tools that we already have at hand, but it would require significant computational power. The second option is that these mDAGs are actually saturated for all cardinalities of the visible variables, even if the dominance-proving rules that we have at hand cannot prove this. In this case, we would potentially be guided to new dominance-proving rules. The third possibility is that these mDAGs are \emph{not} saturated, potentially even for binary cardinalities of the visible variables, but that this cannot be deduced from the comparison of unrealizable supports. That is, it could be the case that they have unrealizable \emph{distributions} over binary variables, even if there is no unrealizable \emph{support} over binary variables. If we find out that this is the case, then we would have found an example of pDAGs $\mathcal{G}_1$ and $\mathcal{G}_2$ such that $\mathcal{M}(\mathcal{G}_1,\vec c_\text{vis}=(2,2,2,2))\neq \mathcal{M}(\mathcal{G}_2,\vec c_\text{vis}=(2,2,2,2))$, while $\mathcal{S}(\mathcal{G}_1,\vec c_\text{vis}=(2,2,2,2))= \mathcal{S}(\mathcal{G}_2,\vec c_\text{vis}=(2,2,2,2))$. This would imply that the converse implication to that of Eq.~\eqref{eq_possibilistic_open_question} \emph{does not} hold,  that is, that Conjecture~\ref{conjecture_converse_of_12} is false.

Note that this third case could also happen with mDAGs \ref{fig_compatible_any_support}(a)-\ref{fig_compatible_any_support}(d): since they can realize any support on on  $\vec c_\text{vis}=(2,2,2,2)$, if we find that they \emph{are not} saturated on  $\vec c_\text{vis}=(2,2,2,2)$ we would  disprove Conjecture~\ref{conjecture_converse_of_12}. On the other hand, note that if we find that they \emph{are} saturated on  $\vec c_\text{vis}=(2,2,2,2)$, we would disprove Conjecture~\ref{conj_nonalgebraic}, since we already know that these mDAGs are \nonalgebraic.    

It is worth noting that, by Lemma~\ref{prop_edge_dropping}, the mDAG \ref{fig_compatible_any_support}(a) is dominated by \ref{fig_compatible_any_support}(b) and \ref{fig_compatible_any_support}(e)-(g), while the mDAG \ref{fig_compatible_any_support}(c) is dominated by \ref{fig_compatible_any_support}(d) and \ref{fig_compatible_any_support}(e)-(g). Therefore, if \ref{fig_compatible_any_support}(a) and \ref{fig_compatible_any_support}(c) are saturated for $\vec c_\text{vis}=(2,2,2,2)$, all of the others are as well.

\section{Conclusion}
\label{sec_conclusion}

In this work, we have gathered all the currently known techniques to prove observational dominance or nondominance between causal structures, extended the scope of these techniques, proved logical implications among them, and applied them to causal structures with three and four visible nodes.  

For the case of three visible variables, Fig.13 of Ref.~\cite{evans_graphs_2016} had already presented representatives of blocks of a proven-equivalence partition. However, in Ref.~\cite{evans_graphs_2016} it was not explicitly shown that all of these were observational equivalence classes (i.e., that they were all observationally inequivalent), and the observational partial order was not studied. Here,  we were able to deduce the observational equivalence classes and observational partial order for 3-node mDAGs.

For the case of four visible variables, we showed that the total number of observational equivalence classes for 4-node mDAGs consistent with a fixed nodal ordering is between $1253$ and $1444$. Furthermore, $1156$ observational equivalence classes are already identified.

Classifying the classical causal structures into observational equivalence classes and establishing the observational partial order among these classes is important for several reasons. For one, it drastically lowers the number of possible causal explanations to be considered for a set of data relative to the total number of mDAGs. For another, the dominance order can be used to simplify the task of model selection (as discussed in Section \ref{significanceforcausaldiscovery}).
  It also helps in assessing how many classes present non-trivial inequality constraints.

Inequality constraints seem to be generic in the sense of the number of observational equivalence classes having inequality constraints growing
with the number of visible variables: it is $1/3$ for 3-node mDAGs consistent with a fixed nodal ordering, and at least $85.2\%$ for 4-node mDAGs consistent with a fixed nodal ordering.  We also argued against relying only on conditional independence relations
 for causal discovery on the basis that these constraints have a poor distinguishing power for observational equivalence classes: for 3-node mDAGs consistent with a fixed nodal ordering,   $3/5$  of the observational equivalence classes can be identified by conditional independence relations 
  alone; for 4-node mDAGs consistent with a fixed nodal ordering, however, this fraction drops to less than $10\%$. These arguments highlight the importance 
   of inequality constraints and nested Markov constraints for the task of classical causal discovery.\footnote{ We acknowledge that in real-life causal discovery scenarios, with many different variables involved, the analysis of conditional independence constraints is already difficult, and the analysis of inequality constraints may be more difficult still. 
     However, our hope is that  by recognizing the foundational importance of nested Markov constraints and inequality constraints, the community might put more emphasis towards developing techniques to study them. }

From the point of view of quantum causality,  this result is also relevant because only the causal structures that exhibit inequality constraints  have the potential of presenting quantum-classical gaps, that is, of being such that 
 the distributions realizable by quantum semantics are a superset of the distributions realizable by classical semantics.  All of the \nonalgebraic mDAGs that we identified here were already identified in Ref.~\cite{Khanna_2023}, but our classification shows that the \emph{majority} of observational equivalence classes have the potential of presenting quantum-classical gaps. This shows that the property of exhibiting a quantum-classical gap might not be the exception, but rather the norm.

Another result from this work that is relevant to quantum physicists who are interested in causal inference is an identification of the set of all 4-node mDAGs consistent with a fixed nodal ordering that admit a fine-tuning theorem similar to the one that holds for Bell's causal structure~\cite{Wood_and_Spekkens}. That is, if the set of conditional independence relations of an observed distribution matches exactly the set of d-separation relations of the mDAG but the distribution \emph{violates} one of the inequality constraints of the mDAG, then this distribution \emph{does not} admit any non-fine-tuned causal explanation in the classical framework for causal modelling. The mDAGs for which such an argument can be made are the ones that belong to observational equivalence classes identified by d-separation alone, one example being the {\tt Square} structure of Fig.~\ref{fig_Bell_and_Square}(b). This result is of interest for quantum physicists, because many times quantum theory predicts such violations.

To finish the classification of structures with four visible nodes, there are two options: (i) we continue the comparison of unrealizable supports (Section \ref{sec_unrealizable_supports}) but with higher cardinalities of the visible variables, that might complete the classification but will be computationally expensive. (ii) we find new dominance-proving or nondominance-proving rules that help us finish the classification. Note that these two options are not mutually exclusive: we might find a new nondominance-proving rule that helps us complete the classification, at the same time that a comparison of unrealizable supports at high enough cardinality would in principle also solve it.

Another open problem that this work has generated is the question of whether possibilistic equivalence implies probabilistic equivalence,  as expressed in Conjecture~\ref{conjecture_converse_of_12}.  In other words, if two causal structures have the same set of unrealizable supports for a certain assignment of cardinalities of the visible variables, does this mean that they have the same set of realizable distributions for those cardinalities? This problem was not solved here, but a possible route of investigation was outlined: all the mDAGs of Fig.~\ref{fig_compatible_any_support} can realize every support on binary variables, even if there is currently no rule that shows that \ref{fig_compatible_any_support}(a)-(g) are observationally equivalent to the saturated class for binary variables. Therefore, if one can show that any of these mDAGs \emph{cannot} realize every probability distribution on binary variables, one has found a counter-example to the conjecture that possibilistic equivalence implies probabilistic equivalence.

\section*{Data Availability}
\begin{tcolorbox}
	The code developed for this work is available at  \url{https://github.com/eliewolfe/mDAG-analysis}
\end{tcolorbox}

\section*{Acknowledgements}
We thank an anonymous reviewer for finding  mistakes in two  of our proofs, as well as presenting the DAGs that are now in Fig.~\ref{fig_referee_example} as a counter-example to a conjecture from a previous version of our manuscript. 
We thank Robin Evans for discussions about his work during the early stages of this project.

\section*{Research Funding}
Research at Perimeter
Institute is supported in part by the Government of Canada through the Department of
Innovation, Science and Economic Development and by the Province of Ontario through
the Ministry of Colleges and Universities. MMA is supported by the Natural Sciences and Engineering Research Council of Canada (Grant No. RGPIN-2024-04419).

\section*{Author Contributions}
The initial idea for this project came from RWS and EW. Conceptual developments and technical proofs were developed by all authors. Code was developed by EW and MMA. Most of the writing was made by MMA and RWS.
All authors accept responsibility for and consent to the submission of this manuscript, have reviewed all the results, and approve the final version of the manuscript.

\section*{Conflict of Interest}
Authors state no conflict of interest.

\FloatBarrier
\bibliographystyle{unsrt} 
\bibliography{references}

\begin{appendices}
\input{./appendix_facesplitting_arXiv}

\input{./appendix_supports}
\input{./appendix_proofs_subsume_arXiv}
\end{appendices} 

\end{document}

%% file: appendix_facesplitting_arXiv.tex
\section{Proof of Strong Facet-Adding}
\label{appendix_facesplitting}

In this appendix we will prove the Strong Facet-Adding proposition, which is reproduced below.

\StrongFS*

To facilitate the proof, in this appendix we will use the language of pDAGs instead of that of mDAGs. Note that, as a special case, this is an alternative proof of the original Evans's Proposition (different from the proof presented in \cite{evans_graphs_2016}).

\begin{figure*}[h!]
    \centering
    \includegraphics[width=0.9\textwidth]{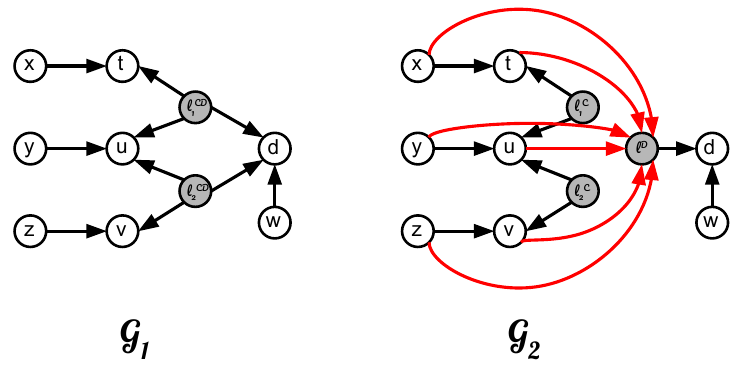}
    \caption{By Lemma \ref{lemma_dominance_bayesian_updating}, we have $\mathcal{G}_2\succeq \mathcal{G}_1$. Here, $C_1=\{t,u\}$, $C_2=\{u,v\}$, and $D=\{d\}$. Note that $\mathcal{G}_1$ does \emph{not} observationally dominate $\mathcal{G}_2$: for example, $\mathcal{G}_1$ has the d-separation relation $x\dsep d$, that is not presented by $\mathcal{G}_2$. Edges that point into a latent node are represented in red.}
    \label{fig_bayesian_update}
\end{figure*}

We start with Lemma \ref{lemma_dominance_bayesian_updating}, that generalizes the idea that a causal structure where the nodes of the set $C$ and their parents are settings for a latent variable is at least as powerful as a causal structure where the nodes of the set $C$ are causally influenced by that latent variable. An example of the application of Lemma \ref{lemma_dominance_bayesian_updating} is presented in Figure \ref{fig_bayesian_update}.

\begin{lemma}[Dominance by Bayesian Updating]
\label{lemma_dominance_bayesian_updating}
Let $\mathcal{G}_1$ be a pDAG that has a set of parentless latent variables $\{l^{CD}_i\}_{i=1,...,n}$, and let the children of $l^{CD}_i$ be $ch_{\mathcal{G}_1}(l_i)=C_i\cup D$, such that $\cup_i C_i$ and $D$ are disjoint. All the nodes in $\{l^{CD}_i\}_{i=1,...,n}$ share the children $D$, and the sets $C_i$, $i=1,...,n$ do not have to be disjoint from each other. Further suppose that none of the nodes in $D$ is an ancestor of any node $c\in \cup_i C_i$.

Let $\mathcal{G}_2$ be another pDAG that is constructed from ${\mathcal{G}_1}$ by deleting all the nodes $\{l^{CD}_i\}_{i=1,...,n}$, adding a latent node $l^D$ whose children are $ch_{\mathcal{G}_2}(l_D)=D$ and whose parents are $pa_{\mathcal{G}_2}(l_D)= \cup_i ( C_i\cup pa_{\mathcal{G}_1}(C_i) )\setminus \{l^{CD}_i\}_{i=1,...,n}$ and adding latent common causes $l^C_i$, $i=1,...,n$  that have no parents and are such that $ch_{\mathcal{G}_2}(l^{C}_i)=C_i$.

Then, $\mathcal{G}_2$ observationally dominates $\mathcal{G}_1$, i.e., $\mathcal{G}_2 \succeq \mathcal{G}_1$.
\end{lemma}
\begin{proof}
    For simplicity of notation, let $C\equiv \cup_i C_i$, $L^{CD}\equiv \{l^{CD}_i\}_{i=1,...,n}$ and $L^C\equiv \{l^{C}_i\}_{i=1,...,n}$. Also define $V\equiv \vis(\mathcal{G}_1) = \vis(\mathcal{G}_2)$.
    
    The idea of this lemma is that, due to $D$ not being part of the ancestors of $C$, one can assess the values of $X_C$ before assessing the values of $X_D$. The knowledge of $X_C$ in $\mathcal{G}_1$ will thus teach us something about the shared common causes $X_{L^{CD}}$ through Bayesian updating. As it turns out, this will be at most as powerful as $\mathcal{G}_2$, where $C$ and its parents act as settings for the latent variable $l^D$. 

    We will show this by explicitly writing down the Markov realizability conditions for each one of the pDAGs. The Markov realizability conditions are equivalent to Eq.~\eqref{eq_realizability}, but written in terms of conditional probability distributions instead of in terms of functional dependences and error variables. It might be helpful to follow the proof below while looking at the example of Figure \ref{fig_bayesian_update}.

    To further simplify the notation, let $V^*=V\setminus(C\cup D\cup (\an_{\mathcal{G}_1}(C)\cap V))$ be the set of visible nodes of $\mathcal{G}_1$ and $\mathcal{G}_2$ except for $C$, $D$ and the visible ancestors of $C$, and  let $L^*=\lat(\mathcal{G}_1)\setminus(\an_{\mathcal{G}_1}(C)\cap \lat(\mathcal{G}_1))$ be the set of latent nodes of $\mathcal{G}_1$ except for the latent ancestors of $C$. Also let $\text{pa}^*_{\mathcal{G}_1}(A)=\text{pa}_{\mathcal{G}_1}(A)\setminus L^{CD}$ be the set of parents of a set of nodes $A$ in $\mathcal{G}_1$ except for $L^{CD}$ and $\an^*_{\mathcal{G}_1}(A)=\an_{\mathcal{G}_1}(A)\setminus L^{CD}$ be the set of ancestors of a set of nodes $A$ in $\mathcal{G}_1$ except for $L^{CD}$.   
      A probability distribution over the visible variables that is realizable by $\mathcal{G}_1$ can be factorized as:
    \begin{align}
   &P_{\mathcal{G}_1}(X_V) = \nonumber \\ & \sum_{X_{ \lat(\mathcal{G}_1)}} P_{\mathcal{G}_1}(X_D|X_{L^{CD}}X_{\text{pa}^*_{\mathcal{G}_1}(D)}) \cdot \nonumber\\
    &\cdot P_{\mathcal{G}_1}(X_C X_{\an^*_{\mathcal{G}_1}(C)}X_{L^{CD}})\cdot \nonumber \\
    & \cdot P_{\mathcal{G}_1}(X_{V^*L^*}|X_C X_D X_{\an^*_{\mathcal{G}_1}(C)})
    \label{eq_decomposition_G1}
    \end{align}

    Note that Eq.~\eqref{eq_decomposition_G1} is a coarse-grained version of the equation that determines realizability by $\mathcal{G}_1$: the probability distributions realizable by $\mathcal{G}_1$ obey even more specific constraints than the ones presented by Eq.~\eqref{eq_decomposition_G1}, because for example not all elements of $D$ are necessarily children of every element of $\text{pa}_{\mathcal{G}_1}(D)$.

    Note that the term $P_{\mathcal{G}_1}(X_C X_{\an^*_{\mathcal{G}_1}(C)}X_{L^{CD}})$ can be written as: 
    \begin{align}
        &P_{\mathcal{G}_1}(X_C X_{\an^*_{\mathcal{G}_1}(C)}X_{L^{CD}})= \nonumber
        \\ &P_{\mathcal{G}_1}(X_C X_{\an^*_{\mathcal{G}_1}(C)})P_{\mathcal{G}_1}(X_{L^{CD}}|X_C X_{\an^*_{\mathcal{G}_1}(C)}).
        \label{bayes_identity}
    \end{align}

    This equality is true by Bayes's theorem, that is, it does not make use of the structure of $\mathcal{G}_1$. 
    
    Note also that $L^{CD}$ is d-separated from the ancestors of $C$ that are not parents of $C$ when we condition on $C$ and its other parents, i.e., the following conditional independence has to hold:
    \begin{equation}
        P_{\mathcal{G}_1}(X_{L^{CD}}|X_C X_{\an^*_{\mathcal{G}_1}(C)}) =  P_{\mathcal{G}_1}(X_{L^{CD}}|X_C X_{\pa^*_{\mathcal{G}_1}(C)}). 
        \label{eq_proofA_dsep}
    \end{equation}

    Substituting Eqs.~\eqref{bayes_identity} and~\eqref{eq_proofA_dsep} into Eq.~\eqref{eq_decomposition_G1}, we get: 
    \begin{align}
   &P_{\mathcal{G}_1}(X_V) = \nonumber \\ & \sum_{X_{ \lat(\mathcal{G}_1)}} P_{\mathcal{G}_1}(X_D|X_{L^{CD}}X_{\text{pa}^*_{\mathcal{G}_1}(D)}) \cdot \nonumber\\
    &\cdot P_{\mathcal{G}_1}(X_C X_{\an^*_{\mathcal{G}_1}(C)})P_{\mathcal{G}_1}(X_{L^{CD}}|X_C X_{\pa^*_{\mathcal{G}_1}(C)})\cdot \nonumber \\
    & \cdot P_{\mathcal{G}_1}(X_{V^*L^*}|X_C X_D X_{\an^*_{\mathcal{G}_1}(C)})
   \label{eq_decomposition_G1_two}
    \end{align}

 To show that $\mathcal{G}_2$ can simulate $\mathcal{G}_1$, we need to show that all the probability distributions that can be factorized as in Eq.~\eqref{eq_decomposition_G1} can also be factorized according to the structure of $\mathcal{G}_2$. To write down the Markov realizability condition for $\mathcal{G}_2$, first note that  $L^*=\lat(\mathcal{G}_2)\setminus((\an_{\mathcal{G}_2}(C)\cap \lat(\mathcal{G}_2))\cup \{l^D\})$,  and that $\text{pa}^*_{\mathcal{G}_1}(A)=\text{pa}_{\mathcal{G}_2}(A)\setminus (L^{C}\cup\{l^D\})$ and  $\text{an}^*_{\mathcal{G}_1}(A)=\text{an}_{\mathcal{G}_2}(A)\setminus (L^{C}\cup\{l^D\})$.  Furthermore, define  $L^{C}_c\subseteq L^{C}$ as the set of elements of $L^C$ that are parents of $c\in C$. For example, in Figure \ref{fig_bayesian_update}(b), $L^{C}_t=\{l_1^C\}$ and $L^{C}_u=\{l_1^C,l_2^C\}$. With this, for $\mathcal{G}_2$ we have:
      
    \begin{align}
   &P_{\mathcal{G}_2}(X_V) = \nonumber \\ & \sum_{X_{ \lat(\mathcal{G}_2)}} P_{\mathcal{G}_2}(X_D|X_{l^{D}}X_{\text{pa}^*_{\mathcal{G}_1}(D)}) \cdot \nonumber\\
    &\cdot P_{\mathcal{G}_2}(X_C X_{\an^*_{\mathcal{G}_1}(C)} X_{L^{C}})P_{\mathcal{G}_2}(X_{l^{D}}|X_C X_{\pa^*_{\mathcal{G}_1}(C)})\cdot \nonumber \\
    & \cdot  P_{\mathcal{G}_2}(X_{V^*L^*}|X_C X_D X_{\an^*_{\mathcal{G}_1}(C)}) = \nonumber
    \\ & = \sum_{X_{ \lat(\mathcal{G}_2)}\setminus{X_{L^C}}} P_{\mathcal{G}_2}(X_D|X_{l^{D}}X_{\text{pa}^*_{\mathcal{G}_1}(D)}) \cdot \nonumber\\
    &\cdot P_{\mathcal{G}_2}(X_C X_{\an^*_{\mathcal{G}_1}(C)} )P_{\mathcal{G}_2}(X_{l^{D}}|X_C X_{\pa^*_{\mathcal{G}_1}(C)})\cdot \nonumber \\
    & \cdot  P_{\mathcal{G}_2}(X_{V^*L^*}|X_C X_D X_{\an^*_{\mathcal{G}_1}(C)})
    \label{eq_decomposition_G2}
    \end{align}
    where in the second equality we used the fact that the latent variables $X_{L^C}$ only appeared in one term, so they can be easily marginalized out.

    If we lump all the variables $X_{L^{CD}}$ of $\mathcal{G}_1$ together in one variable $X_{l^D}$, we can see that  Eqs.~\eqref{eq_decomposition_G1_two} and~\eqref{eq_decomposition_G2} take exactly the same form.

Therefore, $\mathcal{G}_2$ can realize all of the distributions realizable by $\mathcal{G}_1$. Note that the inverse is not true; Fig.~\ref{fig_bayesian_update} presents a counter-example.

\end{proof}

With this Lemma at hand, we can now prove Proposition \ref{Simultaneous_Splitting_Prop}.  The following is a rephrasing of this proposition in pDAG language, where  $\mathcal{G}_1$ corresponds to $\mathfrak{G}'$ of Proposition \ref{Simultaneous_Splitting_Prop} and  $\mathcal{G}_3$ corresponds to $\mathfrak{G}$ of Proposition \ref{Simultaneous_Splitting_Prop}. That is, the proposition below describes the inverse of the transformation described in  Proposition \ref{Simultaneous_Splitting_Prop}. 

\begin{proposition}[Strong Facet-Adding - pDAG language]
 \label{prop_strong_FS_DAG}
 Let $\mathcal{G}_1$ be a pDAG containing a sequence of latent nodes $\{l^{CD}_i\}_{i=1,...,n}$ that share the children $D\subseteq \cap_{i=1,...,n} ch_{\mathcal{G}_1}(l^{CD}_i)$. Denote $C_i\equiv ch_{\mathcal{G}_1}(l^{CD}_i)\setminus D$.
 
 To simplify the notation, we will use $C\equiv \cup_i C_i$ and $L^{CD}\equiv \{l^{CD}_i\}_{i=1,...,n}$  
 
 If the following condition holds:
\begin{enumerate}
    \item $pa_{\mathcal{G}_1}(C)\cup C\subseteq pa_{\mathcal{G}_1}(d)$ for each $d \in D$. This has to hold for both visible and latent parents.
\end{enumerate}

Then,  ${\mathcal{G}_1}$ is observationally equivalent to the pDAG  ${\mathcal{G}_3}$ defined by starting from ${\mathcal{G}_1}$, removing the nodes $L^{CD}$ and adding the new latent nodes $\{l^{C}_i\}_{i=1,...,n}$ and $l_{D}$, whose children are respectively $ch_{\mathcal{G}_3}(l^{C}_i)=C^i$ for $i=1,...,n$ and  $ch_{\mathcal{G}_3}(l_D)=D$.
\end{proposition}
\begin{proof}
    It might be helpful to look at Figure \ref{fig_strong_FS_proof} while following this proof.
    \begin{figure}[h!]
    \centering
    \includegraphics[width=0.4\textwidth]{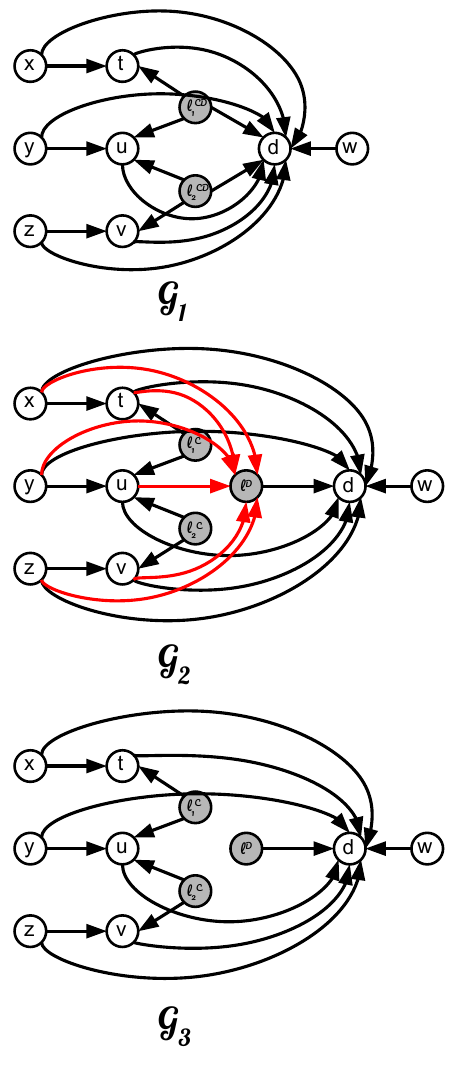}
    \caption{The pDAGs $\mathcal{G}_1$ and $\mathcal{G}_3$ can be shown observationally equivalent by Proposition \ref{prop_strong_FS_DAG}:  Lemma \ref{lemma_dominance_bayesian_updating} says that $\mathcal{G}_2\succeq\mathcal{G}_1$, and Exogenization (Lemma \ref{lemma_exogenize_latents}) says that $\mathcal{G}_2\cong\mathcal{G}_3$. We can see that $\mathcal{G}_1\succeq\mathcal{G}_3$ by structural dominance (Lemma \ref{prop_edge_dropping}). Edges that point into a latent node are represented in red.} 
    \label{fig_strong_FS_proof}
\end{figure}

    By structural dominance (Lemma \ref{prop_edge_dropping}), it is easy to see that $\mathcal{G}_1 \succeq \mathcal{G}_3$.
    
    By Lemma \ref{lemma_dominance_bayesian_updating}, the pDAG $\mathcal{G}_1$ here defined is dominated by a pDAG $\mathcal{G}_2$ where the nodes $\{l^{CD}_i\}_{i=1,...,n}$ are substituted by parentless latent nodes $\{l^{C}_i\}_{i=1,...,n}$ that point to each of the $C_i$, and a node $l^D$ that points to the set $D$ and has parents $pa_{\mathcal{G}_2}(l_D)= \cup_i ( C_i\cup pa_{\mathcal{G}_1}(C_i) )\setminus \{l^{CD}_i\}_{i=1,...,n}$. 

    The next step is to exogenize the latent node $l^D$, using Lemma \ref{lemma_exogenize_latents}. Since $pa_{\mathcal{G}_1}(C)\cup C\subseteq pa_{\mathcal{G}_1}(d)$ for each $d \in D$, when we exogenize $l^D$ there are no new edges added; only edges removed (the edges between $\text{pa}_{\mathcal{G}_2}(l^D)$ and $l^D$). With this process, we reach a pDAG $\mathcal{G}_3 \cong \mathcal{G}_2$. This is the pDAG $\mathcal{G}_3$ that appears in the statement of the proposition.

    Thus, we have that $\mathcal{G}_3\cong \mathcal{G}_2\succeq \mathcal{G}_1$, and we already knew that $\mathcal{G}_1\succeq \mathcal{G}_3$. Therefore, we conclude that $\mathcal{G}_1\cong \mathcal{G}_3$.
\end{proof}

%% file: appendix_supports.tex
\section{Fraser's Algorithm for Feasible Supports}
\label{appendix_supports}

In this section, we present the algorithm of Ref.~\cite{Fraser_Combinatorial_Solution} for finding $\mathcal{S}(\mathcal{G},\vec c_\text{vis})$, the set of supports with cardinalities $\vec c_\text{vis}$ of the visible variables that are realizable by a pDAG $\mathcal{G}$. This algorithm consists of enumerating all the possible responses that a visible variable can have to its parents, given that the cardinalities of the visible variables and the latent variables are fixed.

In Ref.~\cite{Fraser_Combinatorial_Solution}, it is shown that whenever the visible variables have finite cardinalities, we can assume the latent variables to have finite cardinalities without loss of generality. Furthermore, for a given number of events in the support, one gets an upper bound on the cardinality of the latent variables that has to be considered. This upper bound will be discussed after we describe the algorithm.

Given a pDAG $\mathcal{G}$:

\begin{enumerate}
    \item Fix the cardinalities of every latent variable to be $k$ and the cardinality of each visible variable $v$ to be $c_v$.
    \item For each visible variable, enumerate all possible response functions it can have to the valuations of its parents.
    
    For example, if a visible $v$ has only one parent and this parent is latent, it has $(c_v)^k$ possible response functions; we could have that $v$ reacts with the each one of its $c_v$ possible values for each one of the $k$ possible valuations of its parent.
     
    In general, if the product of the cardinalities of the parents of $v$ is $K(v)$, then we have $(c_v)^{K(v)}$ possibilities of response function for $v$.
    \item For each possibility of response functions of all the visible variables of $\mathcal{G}$, we compute the support: the set of visible events that occur under that response function, for some valuation of the latent variables.
\end{enumerate}
The question that is left is how to choose $k$, the cardinality of latent variables. First we note that, if a support containing $s$ events is realizable by $\mathcal{G}$, it will certainly be found with the algorithm for all $k\geq s$. Thus, $s$ is an upper bound on the values of $k$ that we need to consider.

Assuming that all the visible variables have the same cardinality $c$, and considering that the total number of visible variables is  $|\vis(\mathcal{G})|$, the maximum number of events that we can have is $c^{|\vis(\mathcal{G})|}$. So, to be certain that we found all compatible supports, we should choose $k=c^{|\vis(\mathcal{G})|}$. 

Note that if $\mathcal{S}(\mathcal{G},\vec c_\text{vis}=(2,2,2,2))\neq\mathcal{S}(\mathcal{G}',\vec c_\text{vis}=(2,2,2,2))$, i.e., if the set of realizable {\em binary} supports of $\cal G$ and $\cal G'$ are different, then the sets of realizable supports will also be different when we go to higher cardinalities. This is so because we can simply gather events together (for example joining events $x$ and $y$ in $z=x\land y$) to go back to the binary case, where the difference is proven. 

%% file: appendix_proofs_subsume_arXiv.tex
\section{Proofs that comparison of unrealizable supports subsumes graphical methods as rules to show observational nondominance}
\label{appendix_proofs_subsume}

\esep*
\begin{proof}
	The proof of  Theorem 48 of Ref.~\cite{Khanna_2023} starts from the assumption that $\mathfrak{G}$ presents an e-separation relation that is not presented by $\mathfrak{G}'$ and shows that there must be a set of events over binary visible variables that is a realizable support of $\mathfrak{G}'$ but not of $\mathfrak{G}$. This implies that the fact that  $\mathfrak{G}$ does not observationally dominates $\mathfrak{G}'$ can be shown by the comparison of unrealizable supports. 
\end{proof}

\DC*
\begin{proof}
	Assume that the nodes $v$ and $w$ are densely connected in  $\mathfrak{G}=\{\cal D, B\}$ but not in $\mathfrak{G}'=\{\cal D', B'\}$. 
	
	Let $S$ be a set of $2^{|\vis({\mathfrak{G}})|-1}$ events such that in half of the events both $X_v$ and $X_w$ are $0$ and in the other half both are $1$, while every possible combination of $0$'s and $1$'s occurs for all of the other visible variables. For example, for the case when there are four visible nodes $x$, $y$, $v$, $w$, $S$ contains the events
	\begin{equation}
		\label{eq_appendix_denseconnected_support}
	\begin{aligned}
		& \{X_x=0,X_y=0,X_v=0,X_w=0\}, \\
		& \{X_x=1,X_y=0,X_v=0,X_w=0\}, \\
		& \{X_x=0,X_y=1,X_v=0,X_w=0\}, \\
		& \{X_x=1,X_y=1,X_v=0,X_w=0\}, \\
		& \{X_x=0,X_y=0,X_v=1,X_w=1\}, \\
		& \{X_x=0,X_y=1,X_v=1,X_w=1\}, \\
		& \{X_x=1,X_y=0,X_v=1,X_w=1\}, \\
		& \{X_x=1,X_y=1,X_v=1,X_w=1\}.
	\end{aligned}
\end{equation}
	
	The result of Ref.~\cite{evans_dependency} says that a distribution $P$ where $X_v$ and $X_w$ are perfectly correlated while all of the other visible variables are mutually independent and uniformly distributed is realizable by $\mathfrak{G}$ if $v$ and $w$ are densely connected in $\mathfrak{G}$. Therefore, $S$ is a realizable support of  $\mathfrak{G}$.
	
	On the other hand, in $\mathfrak{G}'$, the nodes $v$ and $w$ are not densely connected. As pointed out in the proof of Theorem 6.4 of Ref.~\cite{evans_dependency}, this implies that $\mathfrak{G}'$ presents a nested Markov constraint between $X_v$ and $X_w$. In the binary case, such a constraint takes one of two forms~\cite{SEMs}:	
	\begin{align}
		&P(X_v=x_v|X_A,X_w=0,\text{do}(X_B))= \nonumber \\ &P(X_v=x_v|X_A, X_w=1,\text{do}(X_B)); \text{ or}
		\label{eq_1_do}
	\end{align}
	\begin{align}
		&P(X_v=x_v|X_A,\text{do}(X_B, X_w=0))= \nonumber  \\ &P(X_v=x_v|X_A,\text{do}(X_B, X_w=1)),
		\label{eq_2_do}
	\end{align}
	for sets $A\subseteq\nodes(\mathfrak{G})$ and $B\subseteq\nodes(\mathfrak{G})$ that do not include $v$ nor $w$, and where the do-conditionals that appear in these constraints are identifiable from observational data. 
	
	Regardless of whether it is identifiable or not, every component of a do-conditional is always larger than or equal to the corresponding component of the respective joint observational distribution. More specifically, $P(X_a|\text{do}(X_b))\geq P(X_a,X_b)$. Therefore, since the support $\cal S$ includes all events where $X_v= X_w$, then if $P$ has support $\cal S$ we must have:
	\begin{align}
		& P(X_v=0|X_A,X_w=0,\text{do}(X_B)) >0 \label{eq_aaa} \\
		& P(X_v=0|X_A,\text{do}(X_B, X_w=0)) >0.  \label{eq_bbb}
	\end{align}
	
	On the other hand, we know that the do-conditionals of the forms that appear in Eqs.\eqref{eq_1_do} and \eqref{eq_2_do}, when identifiable, are obtained by the joint probability distribution over all visible variables divided by marginal conditional probability distributions, \emph{none of which} include both $X_v$ and $X_w$ at the same time~\cite{causality_pearl}. Since $P$ has support $\cal S$, these marginal conditional probability distributions will always take non-zero values. The joint distribution over all visible variables, on the other hand, is zero when $X_v\neq X_w$. When the nested constraint between $v$ and $w$ is of the form of Eq.~\eqref{eq_1_do}, the do-conditionals that appear there are identifiable, and thus
	\begin{equation}
		P(X_v=0|X_A,X_w=1,\text{do}(X_B)) =0.
	\end{equation}
	
	Together with Eq.~\eqref{eq_aaa}, this shows that Eq.~\eqref{eq_1_do} cannot be satisfied by a distribution with support $\cal S$.  When the nested constraint between $v$ and $w$ is of the form of Eq.~\eqref{eq_2_do}, the do-conditionals that appear there are identifiable, and thus
	\begin{equation}
		P(X_v=0|X_A,\text{do}(X_B, X_w=1))  =0.
	\end{equation}
	
	Together with Eq.~\eqref{eq_bbb}, this shows that Eq.~\eqref{eq_2_do} cannot be satisfied by a distribution with support $\cal S$.

	Thus, we find a binary support that is realizable by   $\mathfrak{G}$ but not by  $\mathfrak{G}'$, showing that the comparison of supports subsumes the comparison of densely connected pairs as a rule to show observational inequivalence.
	
\end{proof}

\DEF*
\begin{proof}
	By the assumption that  $\mathfrak{G}$ and $\mathfrak{G}'$  can be shown inequivalent by the directed-edge-free rule, it follows that  $\mathfrak{G}$ is observationally equivalent to a directed-edge-free mDAG $\mathfrak{G}_\text{DEF}$, and  $\mathfrak{G}'$ is observationally equivalent to a directed-edge-free mDAG $\mathfrak{G}'_\text{DEF}$. Both $\mathfrak{G}_\text{DEF}$ and $\mathfrak{G}'_\text{DEF}$ have a trivial directed structure. Since they are different, their simplicial complexes must be different. Suppose that the simplicial complex of  $\mathfrak{G}$  has a facet $B$ that is not a face of  the simplicial complex of $\mathfrak{G}'$. This means that the nodes in $B$ share a joint common cause in $\can(\mathfrak{G})$, but not in $\can(\mathfrak{G}')$. 
	
	In Example 2 of Ref.~\cite{steudel_ay_2015}, it was shown that a distribution where all the variables $X_1,...,X_n$ are perfectly correlated (i.e., where all the weight is on elements of the event space where $X_1=X_2=...=X_n$) is only realizable by causal structures where all the nodes corresponding to such variables share a common ancestor. Therefore, in our case, $\mathfrak{G}$ can realize perfect correlation between the node variables in $B$ while $\mathfrak{G}'$ cannot. Taking the case of distributions over binary variables, this implies that there are supports consisting of two events that are realizable by $\mathfrak{G}$  but not by $\mathfrak{G}'$; namely, supports where in one of the events all of the variables associated with nodes of $B$ take the value $0$ (and the remaining variables can take any value), and in the other event all of the variables associated with nodes of $B$ take the value $1$ (and the remaining variables can take any value).
 
\end{proof}